%% file: arxiv.tex
\documentclass[article]{elsarticle}

\usepackage[utf8x]{inputenc}
\usepackage{hyperref}
\usepackage{graphicx}
\usepackage{svg}
\usepackage{subfig}
\graphicspath{{./img/}}
\usepackage{amsmath}
\usepackage{dsfont}
\usepackage{textcomp}
\usepackage{amsmath,amssymb,amsfonts}
\usepackage{booktabs}
\usepackage{multirow}

\begin{document}

\begin{frontmatter}

\title{Description of Structural Biases and Associated Data\\in Sensor-Rich Environments}

\author{Massinissa Hamidi and Aomar Osmani}
\address{LIPN-UMR CNRS 7030, Univ. Sorbonne Paris Nord\\
{\tt \{hamidi,ao\}@lipn.univ-paris13.fr}}




\begin{abstract}
In this article, we study activity recognition in the context of sensor-rich environments. We address, in particular, the problem of inductive biases and their impact on the data collection process. To be effective and robust, activity recognition systems must take these biases into account at all levels and model them as hyperparameters by which they can be controlled.
Whether it is a bias related to sensor measurement, transmission protocol, sensor deployment topology, heterogeneity, dynamicity, or stochastic effects, it is important to understand their substantial impact on the quality of activity recognition models.
This study highlights the need to separate the different types of biases arising in real situations so that machine learning models, e.g., adapt to the dynamicity of these environments, resist to sensor failures, and follow the evolution of the sensors topology. We propose a metamodeling process in which the sensor data is structured in layers. The lower layers encode the various biases linked to transformations, transmissions, and topology of data. The upper layers encode biases related to the data itself. This way, it becomes easier to model hyperparameters and follow changes in the data acquisition infrastructure. We illustrate our approach on the SHL dataset which provides motion sensor data for a list of human activities collected under real conditions. The trade-offs exposed and the broader implications of our approach are discussed with alternative techniques to encode and incorporate knowledge into activity recognition models. 
\end{abstract}

\end{frontmatter}

%\linenumbers

%
\section{Introduction}
\label{sec:introduction}
Activity recognition aims to provide accurate and opportune information based on people's activities and behaviors~\cite{abdallah2018activity}.
It is of utmost importance in many applications ranging from patient monitoring systems~\cite{aliverti2017wearable}, ambient assisted living~\cite{storf2009event}, etc.
Tracking daily activities and providing, for example, real-time feedback to patients with obesity, diabetes, or cardiovascular diseases as well as up-to-date reports to clinicians has the potential to enhance the health system~\cite{dang2020sensor,andreu2015wearable,spiegel2014validation,gjoreski2011accelerometer}.
Similarly, energy consumption in large infrastructures and housings could be monitored and regulated based on the real-time tracking of human activities~\cite{lima2015user}.
{\bf Sensor rich environments.}
A growing number of domains is witnessing the development of sensor-rich environments powered by the ever-increasing pervasiveness of sensing devices.
%
%
%In this paper, we investigate integration of data sources in sensor-rich environments.
%__
%This topic is the subject of much research with regards to its importance, with the increase, on the one hand, of so-called traditional sources of information as we know them, e.g. web resources, databases, etc., but also, on the other hand, sensors that generate huge amounts of data.
%The information generated by these data sources is, above all, characterized by heterogeneity, conflict, and redundancy.
%__
%Enabling machine learning models to exploit the profusion of data sources and diversity of perspectives provided by sensor-rich environments, like wearables in our case, has the potential to yield more effective results.
%__
%
%
%Une notion importante caractérise ces environnements est celle de la couverture fournie par les déploiements d'objets connectés. Cette notion est primordiale pour pouvoir capturer les mouvements des utilisateurs de manière non-ambigüe.
An important notion that characterizes these environments is that of \textit{coverage} provided by the sensing nodes encompassing the deployment. This notion is essential to allow capturing user movements in an unambiguous way.
In this sense, integration of the diverse sensing modalities plays a key role via three broad goals:  increasing the completeness, conciseness, and correctness of data~\cite{dong2018data}.
%
%cette notion de couverture se traduit au travers à la fois des déploiements corporels et non-corporels d'objets connectés.
%Des exemples de déploiements concréts sont présentés.
%Cette notion de couverture est liée (i) aux capacités intrinsèques de chaque capteur à couvrir une surface et à la manière dont cela est fait.
%Elle est ensuite liée (ii) à la manière dont les différents capteurs sont disposés sur le corps (dans le cas des déploiements corporels) et dans l'environnement dans lequel l'utilisateur est appelé à évoluer (dans le cas des déploiements non-corporels).
This notion of coverage is linked to (i) the intrinsic capabilities of each sensor to cover a surface and the way this is done.
It is then linked to (ii) the way the different sensing nodes are placed on the body (in the case of body deployments) and in the environment in which the user has to evolve (in the case of non-corporeal deployments).
This poses roughly two challenges: the first one is related to the sensors placement (and displacement from an initial configuration)~\cite{attal2015physical} and the second is related to the heterogeneity of the sensors deployments~\cite{mainetti2011evolution}, i.e. the proprietary and non-proprietary solutions (both in on-body and non-corporeal deployments) which hinder the process of integration.
Both issues have an important impact on the notion of coverage which determine the quality and robustness of the final activity recognition models.
%{\bf Sensors placement and displacement.}
%Authors in~\cite{kunze2008dealing} proposed a set of heuristics that significantly increase the robustness of motion sensor-based activity recognition with respect to sensor displacement. In particular, they show how, within certain limits and with modest quality degradation, motion sensor based activity recognition can be implemented in a displacement tolerant way.

%{\bf Heterogeneity of deployments}
%"WSNs are characterized by high heterogeneity because there are many different proprietary and non-proprietary solutions. This wide range of technologies has delayed new deployments and integration with existing sensor networks."~\cite{mainetti2011evolution}, \dots

{\bf Constraints related to the sensing devices.}
In addition to coverage issues, many kind of constraints related to the sensing devices have to be taken into account when designing activity recognition models.
These constraints fall generally into 3 major parts which can be organized in a bottom-up fashion.
First, the sensing constraints related to the intrinsic characteristics of the sensing devices (precision, sensitivity, dynamic range, thermal drift, etc.)~\cite{ida2014sensors}.
Second, energy and computational constraints which are related to the individual sensing nodes and the way they modulate the measurement (sampling frequency, wake-up/sleep modes, etc.) in order to comply to these constraints~\cite{wang2013energy}.
Third, we consider the collective dimension of the sensor deployments where many different challenges related to transmissions arise including power, computation, security and interference, material constraints, robustness, continuous operation, and regulatory requirements~\cite{krishnamachari2005networking,pervez2009medical}.
%Third, constraints related to the sensing nodes taken collectively along with the transmission issues that arise 
%We start by describing sensing constraints which impact the sensed measurements.
%We then move to the energy and computational constraints related to the individual sensing nodes and how they impact the measurement (sampling frequency, etc.) as well as the transmission aspects.
%Next, we present the constraints related to the sensing nodes taken collectively and in particular the transmission issues that arise in body sensor networks as well as in device-free settings.
%
Along with the coverage issues, these constraints have an important impact on the final activity recognition models as the quality of the data, its availability, its reliability, among other things, are not ensured.

{\bf Dynamic selection of inductive biases.}
%\paragraph{Evolution of the phenomenon and sensors deployments}
Current approaches for activity recognition are based on the \textit{activity recognition chain}~\cite{bulling2014tutorial} which defines several steps through which the sensed signals pass.
This is an \textbf{inductive} process which involves searching for a hypothesis (or theory), among a hypothesis space, able to explain the observations.
Often the hardest problem in this process is the initial choice of a hypothesis space; it has to be large enough to contain a solution to the problem at hand, yet small enough to ensure good generalization from a small number of examples~\cite{mitchell1980need}.  The choice of inductive biases (eg: preprocessing filter, segment size, feature set) has a significant impact on this problem.%This is determined by the choice or \textbf{selection} of inductive \textbf{biases}, e.g. preprocessing filter, segment size, set of features, etc.
In addition to the challenges of coverage and the constraints related to detection devices, environments rich in sensors and the phenomena to be detected are subject to change during the actual deployment of activity recognition models in real situations. 
%Beyond the coverage challenges and the constraints related to the sensing devices, sensor-rich environments as well as the sensed phenomenon (users movements) are subject to evolution during the actual deployment of the activity recognition models in real-world situations.
%As we will have the occasion to see in the next sections (Sections~\ref{sec:sensor-rich-environments} and~\ref{sec:contraintes-liees-aux-objets-connectes}), sensor-rich environments are characterized by dynamicity and evolution.
%We saw in particular that sensors deployments evolve often and are subject to packets loss among many other issues.
While correcting for inductive biases applying to specific problems may be of benefit in controlled environments, doing so during the early stages of the activity recognition chain in such environments inevitably leads to a spatial exploration of inefficient hypotheses. Worse yet, the final hypothesis that would be chosen may not explain the learning examples.
A natural solution is to delay the selection of the inductive biases as late as possible and to maintain competing hypotheses able to quickly dealing with new situations by making a dynamic selection of the inductive biases.
This leads to different implications from an operational point of view, namely, maintaining a set of inductive bias alternative candidates (the domain) and rapidly exploring the space in order to elect the appropriate hypothesis (amount of supervision with learning examples).
In other words, the exploration of the hypotheses space must be structured by exploiting a priori knowledge on the deployments of sensors and the phenomenon itself. 

%Note that this operational \dots exhibit a trade-off between richness of domain models and the quantities of examples discussed in more details in Sect.~\ref{sec:trade-off-domain-models-quantities-examples}.
%
%Learning human activities \dots 
%\textbf{inductive processes} \dots
%\textbf{bias} \dots
%\textbf{dynamic selection} \dots
%
%Here we provide \dots 
%Unlike the traditional activity recognition chain~\cite{bulling2014tutorial} which is widely used in the literature around activity recognition, delaying the incorporation of inductive biases has two main advantages.
%The first is \dots
%The second is \dots
%

{\bf Use-case and evaluation.}
To illustrate the advantages of the dynamic selection of inductive biases, we present a use-case pertaining to the SHL dataset~\cite{gjoreski2018university}, one of the most recent and featured datasets in human activity recognition literature.
%small description
This dataset is a highly versatile and precisely annotated dataset dedicated to mobility-related human activity recognition (750 hours of labeled locomotion data).
%why shl dataset
In contrast to related representative datasets like~\cite{zheng2010geolife,zhang2012usc,yu2014big,carpineti2018custom},
the SHL dataset provides, simultaneously, multimodal and multilocation locomotion data recorded in real-life settings.
%how evaluated
We evaluate a first model based on the traditional activity recognition chain instantiated using neural networks-based architectures.
We then illustrate the dynamic inductive bias selection using the proposed approach based on the optimization of architecture's hyperparameters~\cite{hamidi2020data}.
%Both aspects are evaluated empirically.
%We concentrate on two important aspects (1) adaptive segmentation and (2) adaptive sampling.

{\bf Organization of the paper.}
The article is organized as follows.
The ~\ref{sec:human-activity-recognition} section will introduce the context of the human activity recognition and clarify the scope of our work.
The ~\ref{sec:sensor-rich-environments} section will insist on the notion of coverage characterizing the deployments of sensors and their topologies.
The section ~\ref{sec:contraintes-liees-aux-objets-connectes} will review the constraints related to the sensing devices and their impact on the actual sensor observations used in the subsequent recognition steps.
Section ~ \ref{sec:dynamic-inductive-bias-selection} will describe the perspective of dynamic selection of inductive biases to deal with some of encountered problems.
The ~\ref{sec:case-study} section will present the used dataset. Evaluation results on this data will be presented in section ~\ref{sec:evaluations}.
A discussion of the proposed perspectives and future directions concludes this document. 

\section{Human Activity Recognition}\label{sec:human-activity-recognition}
There are various types of human activities.  Depending on their complexity, authors in~\cite{aggarwal2011human} categorized human activities into four different levels: \textit{gestures}, \textit{actions}, \textit{interactions}, and \textit{group} activities.
\textit{Gestures} are elementary movements of a person’s body part, and are the atomic components describing the meaningful motion of a person. ‘Stretching  an  arm’  and  ‘raising  a  leg’  are  good  examples  of  gestures. \textit{Actions}  are  single  person  activities  that  may  be  composed  of  multiple  gestures organized temporally, such as ‘walking’, and ‘waving’.  \textit{Interactions} are human activities that involve two or more persons and/or objects.  For example, ‘two persons discussing’ is an interaction between two humans and ‘a person handing an object to another’ is a human-object interaction involving two humans and one object.  Finally, \textit{group} activities are the activities performed by conceptual groups composed of multiple persons and/or objects.  ‘A group of persons marching’ and  ‘a group having a meeting’ are typical examples of them.

%how these activities are monitored?
Many different approaches have been introduced in the literature to tackle human activity recognition.
These approaches differ in terms of the type of sensing strategies that are used to capture the body movements.
These approaches can be categorized into: (i) radio-frequency/device-free, (ii) vision and depth images, and (iii) inertial sensor.
%\begin{itemize}
%    \item inertial sensors;
%    \item vision and depth images;
%    \item channel state information;
%\end{itemize}
%
%\textcolor{blue}{Rajouter ici les stretchable strain sensors, etc. (tout ce qui touche à la reconnaissance d'activités humaines et non pas aux context-aware applications où les capteurs d'activités ne forment qu'une partie du setup.}
%
Device-free activity recognition refer to the use of the signals generated by ordinary wireless equipment (such as WLAN) to capture users movements in a non-invasive way~\cite{jiang2018towards}.
%One of the issues in these kind of approaches is "The wireless signals arriving at the receiving devices usually carry substantial information that is specific to the environment where the activities are recorded and the human subject who conducts the activities. Due to this reason, an activity recognition model that is trained on a specific subject in a specific environment typically does not work well when being applied to predict another subject’s activities that are recorded in a different environment."~\cite{jiang2018towards}.
%
The vision and depth images-based methods utilize spatio-temporal characteristics extracted from video sequence and the 3D motion feature to describe the action~\cite{ramanan2003automatic}.
In the case of inertial sensor-based approaches, on-body sensors placed in different parts of the body generate streams of observations, like acceleration, which describe similarly the performed actions.
In this paper, we are mainly interested in the latter approach but the proposed perspective can apply similarly for the two other ones.
\begin{figure}[h!]
    \centering
\subfloat[]{
    \includegraphics[width=5cm]{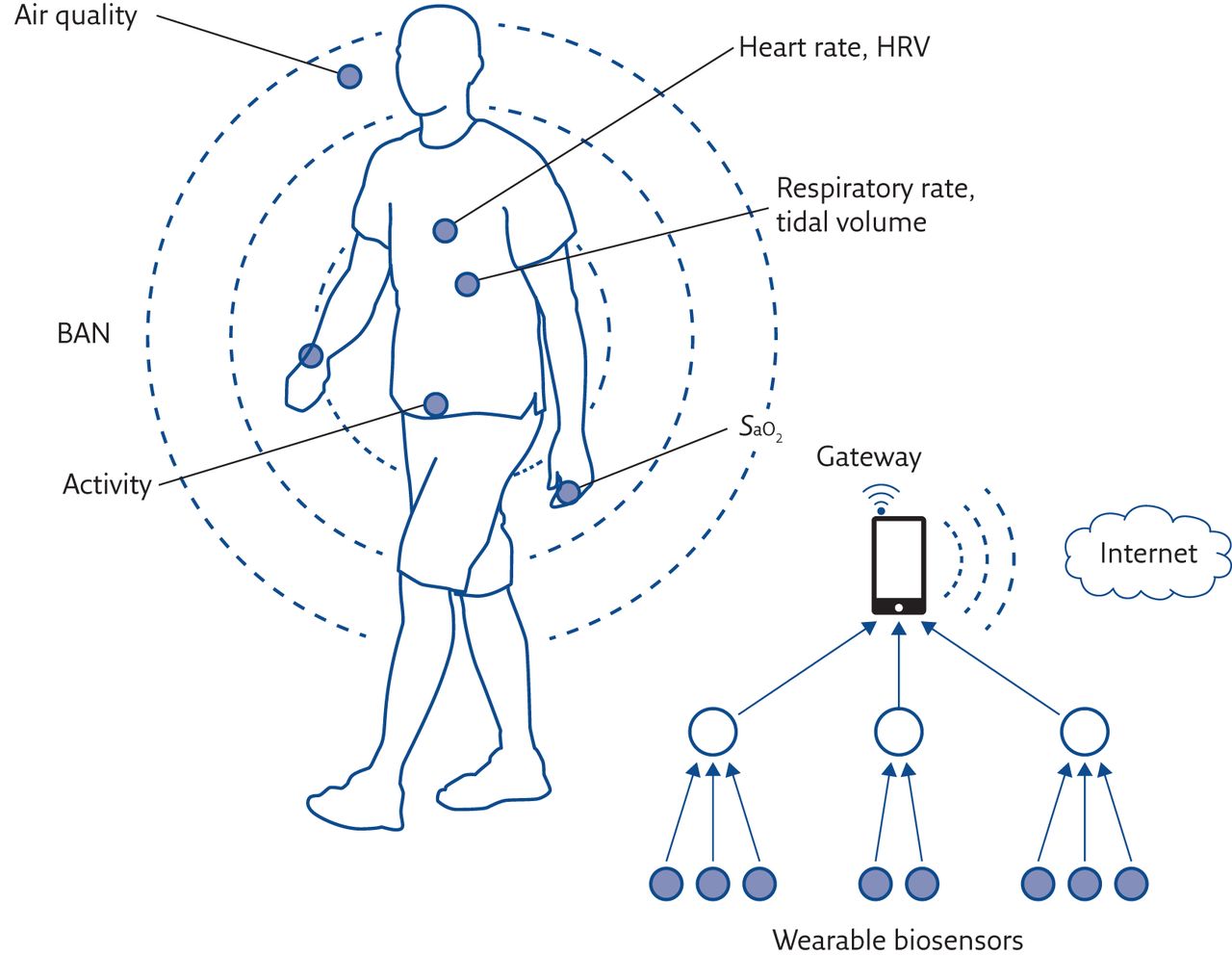}
    \label{fig:patient_monitoring_system}
}
\subfloat[]{
    \includegraphics[width=6cm]{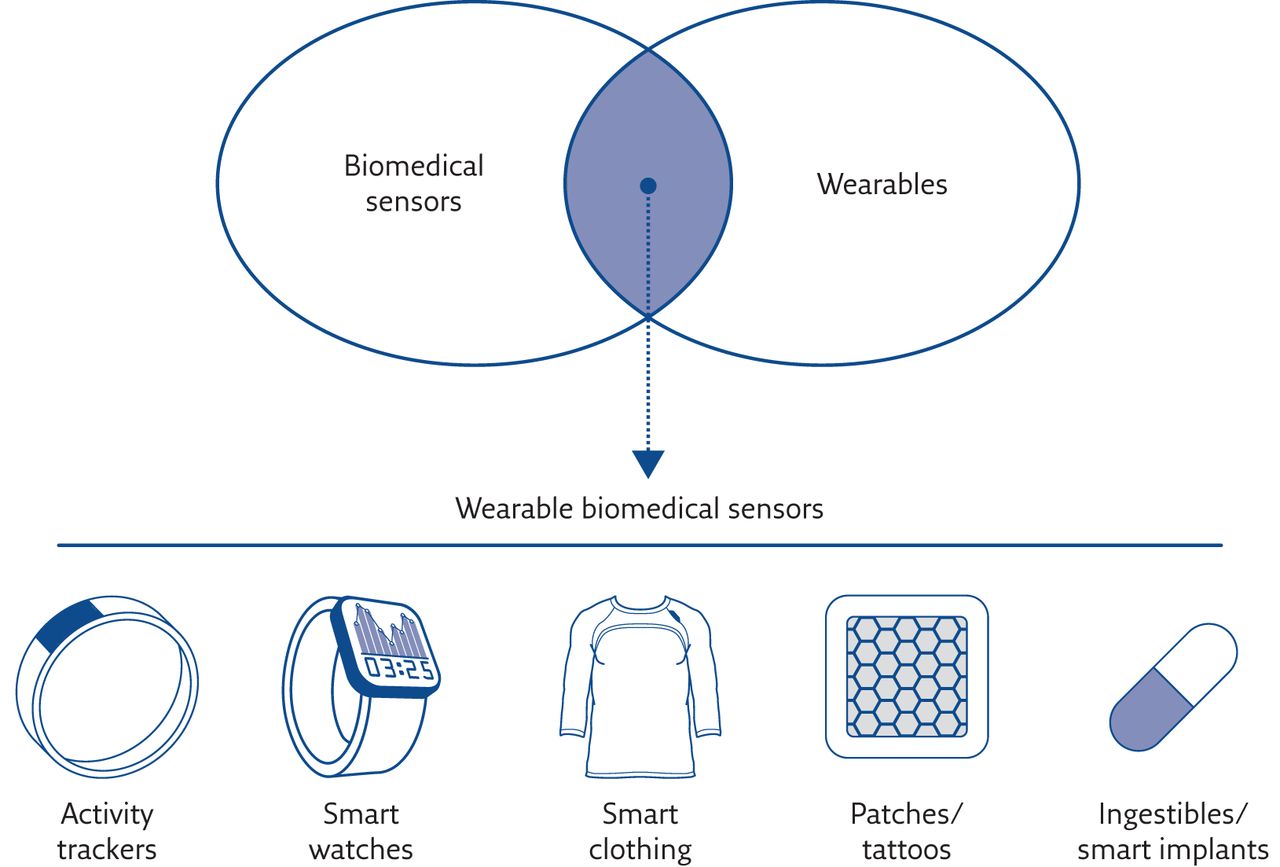}
    \label{fig:wearable_biomedical_sensors}
}
    \caption{(a) Patient monitoring system. (b) Wearables for biomedical sensing. From~\cite{aliverti2017wearable}.}
\end{figure}
For example, Figure~\ref{fig:patient_monitoring_system} illustrates a body area network dedicated to patient monitoring and encompassing various sensing nodes responsible of capturing the vital signs as well as the patient's activities.
Figure~\ref{fig:wearable_biomedical_sensors} illustrates a set of wearables for biomedical sensing. These include activity trackers, smart watches, smart clothing, patches/tattoos, ingestibles/smart implants.
%
%These are referred to as wearables and Figure~\ref{fig:wearable_biomedical_sensors} shows some wearable sensing devices used in biomedical applications.

%\subsection{Applications}
%Cette structuration est tirée de~\cite{avci2010activity}.
%\textcolor{red}{todo:} il faut refaçonner cette structuration autour de la problématique principale traitée dans ce papier qui est liée aux déploiements d'objets connectés ! concrétement, il faut donner des exemples d'applications tout en se focalisant sur les types de déploiements utilisés (BANs, earables, etc.).

%
In addition to \textit{medical applications}, many different applications see the opportunity of leveraging the context provided by human activity recognition models including \textit{assisted living \& home monitoring}, and \textit{sports \& leisure applications}~\cite{avci2010activity}.

\subsection{Medical applications}
Medical applications are categorized in~\cite{avci2010activity} into \textit{monitoring \& diagnosis}, \textit{rehabilitation}, \textit{correlation between movement \& emotions}, and \textit{child \& elderly care}.
%monitoring and diagnosis
Rather than relying on a single short medical appointment, patients continuous monitoring has many advantages as some diseases like Parkinson's disorder need longer periods of examination and in certain situations a careful assessment to detect hidden symptoms~\cite{albert2012using,cheng2017human}.
%Monitoring and diagnosis of patients using continuously generated data, rather than data from a single short medical appointment. Certain patients, such as those facing Parkinson’s Disorder, need longer periods of examination to accurately identify the symptoms and monitor the course of  the  condition~\cite{albert2012using,cheng2017human}.
%Moreover,  some  symptoms  may  not present  themselves  during  a  medical  examination.
Activity recognition  can  be  a  tool  to  help  doctors  diagnose  such conditions as they monitor daily activities in order to detect deviations  from  a  typical  routine  or  deterioration  of  a patient’s  current  physical  status~\cite{lockhart2012applications}.

%rehabilitation, postoperative care
Similarly, post surgical care is an important part of the surgical recovery process after e.g., joint replacements, cardiac surgery, stroke, breast cancer, and those in the intensive care unit~\cite{holick2008physical,cumming2008effect,appelboom2015mobile}.
With the introduction of minimally invasive surgery, the recovery time of patients has been shortened significantly. This has led to a shift of postoperative care from hospital to home environment~\cite{lo2007real}.

%movement and emotion
Physical activities provide also contextual information which can be used to disambiguate certain situations in the case of emotion recognition systems which are mainly based on the monitoring of biophysiological signs like heart rate, blood pressure, etc.~\cite{al2013context}.
%: It represents the patient’s current physical activities such as walking, running, or sleeping.
%It was adopted in several studies [3, 4, 8].
In fact, these physical activities have direct effects on the normal vital signs. For example, normal heart rate while running or climbing up stair is higher than while walking or lying down~\cite{mohomed2008context}. Similarly, normal blood pressure during sitting or sleeping is less than during eating or doing physical exercise such as running~\cite{copetti2009intelligent}.

%child & elderly care
%"In [30], an event-based activity classifier is proposed to monitor and recognize daily living activities in mobility-impaired stroke patients using a trunk-fixed sensor that integrates barometric pressure (BP) and inertial sensors. The authors proposed a double-stage hierarchical fuzzy logic inference system. The first stage processed the events such as the start/end of lying or walking periods, and detected postural transitions while the second stage improved the activity recognition by providing a simple way to integrate the typical behavior of the subject and biomechanical constraints. Sensor attachment to the human body involves fixing sensors directly to skin [13] as well as clothing."~\cite{attal2015physical}
%
In the case of child care, authors in~\cite{osmani2017platform,osmani2017machine}, for example, proposed to leverage activities, among various contextual, information in order to recognize infants' emotional state, like hungry, comfort, etc., and subsequently provide adequate soothing solutions such as lightning effects and lullabies.
In the other hand, elderly care has attracted a lot of research~\cite{gjoreski2011accelerometer,jalal2014depth,de2017mobile}.
For example, Gjoreski et al~\cite{gjoreski2011accelerometer} investigated posture recognition and fall detection using an on-body deployment of accelerometers, while many different settings and evaluation environments were introduced to further develop these kinds of applications~\cite{alvarez2013evaluation,de2017mobile}.

\subsection{Assisted living \& home monitoring}
Similarly, in the context of aging population, providing assistance, such as support for daily activities, fall detection, etc., is of utmost importance. The need for such applications is constantly increasing as it is the case for what are called smart homes and smart buildings which are becoming more of a standard nowadays.
Increasing demand for independent living lifestyle has motivated the research and development of smart home's monitoring technologies~\cite{bakar2016activity}.

Beyond the prevalent industrialization of such solutions, a long line of research has flourished around the exploitation of human activities as contextual information in order to improve assisted living.
For example, in~\cite{rafferty2017activity}, authors proposed a system that recognizes intentions in an assisted living setting.
%Intentions are recognized using \dots and human activity recognition.
%
Authors in~\cite{zhu2008wearable,zhu2009human} developed a smart assisted living system to help and provide support to elderly people when there is an emergency situation. This system consists of a body sensor network, a companion robot, a Smartphone, and a remote health provider. In order to enable natural human-robot interaction, the robot needs to infer the human intentions and situations from the motion data and vital signs of the human subject. For example, when an elderly person falls down accidentally, the algorithm will be able to detect this situation and communicate with a companion robot to help the patient.
In~\cite{zhu2012realtime}, authors proposed a method to recognize complex human daily activities which consist of simultaneous body activities and hand gestures in an indoor environment.
%Concretely, a wireless power-aware motion sensor node is developed which consists of a commercial orientation sensor, a wireless communication module, and a power management unit. To recognize complex daily activities, three motion sensor nodes are attached to the right thigh, the waist, and the right hand of a human subject, while an optical motion capture system is used to obtain his/her location information."
%Figure~\ref{} shows an overview of the hardware platform used in~\cite{zhu2012realtime} for complex daily activity recognition.

%energy
Another important application of activity recognition is for home monitoring especially energy consumption in buildings~\cite{hagras2004creating,hoque2012aalo,thomas2016activity,garcia2017framework}.
Authors in~\cite{thomas2016activity}, for example, introduced the notion of an activity-aware building automation system. The system uses activity recognition to identify current activities and activity prediction to anticipate upcoming activities. Both sources of information are used to educe energy consumption by making decisions regarding devices to turn off.
Similarly, in~\cite{hagras2004creating}, authors identified activities including sleeping, working, and leisure in order to adjust temperature and lighting conditions depending on the needs of each activity.

%\begin{figure}[h!]
%    \centering
%    \includegraphics[width=8cm]{img/overview-of-the-hardware-platform.png}
%    \caption{Overview of the hardware platform used in~\cite{zhu2012realtime} for recognizing complex daily activities in a smart home setting.}
%    \label{fig:overview-of-the-hardware-platform}
%\end{figure}

%In~\cite{shoaib2016complex}, authors \dots
%Authors in~\cite{shoaib2016complex} studied also the effect of window size on the recognition performances (See Figure~\ref{fig:effect-of-window-size-on-recognition}.

\subsection{Sports \& leisure applications}\label{sec:applications:sports-leasure-applications}
Sports and leisure applications are also part of the surge that stem from the development and widespread of sensor-rich environments.
Along with energy expenditure estimation to assess the activity level of a subject, activity recognition was used to monitor sport activities for different purposes like smart coaches or fault detection in competitions.
In~\cite{hsu2018human},  for example, authors proposed to recognize various sport activities such as tennis, badminton, dribbling basketball, etc.
Whereas in~\cite{taborri2019automatic}, authors presented an inertial sensors-based setting aimed to detect faults in race walking.
In~\cite{heinz2006using}, authors conducted a study on martial arts movements recognition using a set of body-worn gyroscopes and acceleration sensors.

%\subsection{HAR as contextual models}
In the case of leisure applications, one can evoke the use of activity recognition to provide contextual information for enhancing recommender systems.
Most existing research in the domain of personalized recommendations focuses on suggesting users the most interesting items based on the users' preferences, but without taking into account any additional contextual information, such as the user's location, the weather, the time of day, the day of the week, the user’s physical activity and mobility, etc. However, the context is an important aspect in the decision process of the user, particularly for mobile applications.~\cite{de2014context}.
In~\cite{unger2015latent,de2018heart,de2014context} for example, authors proposed frameworks to detect the current context and activity of the user by analyzing data retrieved from different sensors available on mobile devices.
On top of this framework, a recommender system is built to provide users a personalized content offer, consisting of relevant information such as sport activities, points-of-interest, train schedules, and touristic info, based on the user’s current context.

%\textcolor{blue}{finir ici par le fait que le déploiement est très important, qu'il diffère, etc. --- trouver éventuellement une référence qui parle de cela plus précisément et mieux.}

%this leads us to the sensors deployments used in practice to perform activity recognition in real-world scenarios \dots
%_
%_
\section{Sensor-Rich Environments}\label{sec:sensor-rich-environments}

The coverage problem in wireless sensor networks can be broadly defined as a measure of how efficiently it is monitored by its sensor nodes. This issue has generated a lot of interest over the years and as a result many coverage protocols have been proposed ~\cite{elhabyan2019coverage}.
The notion of coverage is linked (i) to the intrinsic capacities of each sensor to cover a surface and to the way in which this is done.
It is then linked (ii) to the way in which the various sensors are placed on the body (in the case of bodily deployments) and to the environment in which the user is supposed to operate (in the case of non-bodily deployments). We will explore it through deployment examples.

This section will illustrate various examples of sensor-rich environments, including on-body and non-corporal deployments used in the context of human activity recognition. We will then explore an important notion that defines the coverage score of a given sensor deployment, namely the sensing capabilities of individual sensory nodes. We will then proceed to the collective dimension of the sensors which is the topology or the placement of the detection nodes in Cartesian space. We will focus on the placement of sensors in the case of on-body deployments and the long line of research that has studied this aspect. 

\subsection{Examples of sensor-rich environments}
%- décrire ici différents types de déploiements (corporels et non-corporel) implémentés réellement (en labo ou pour des solutions commerciales);

%- mettre des images de ces déploiements ;

%- décrire les types de capteurs, leurs capacités de mesures, etc. ;

%\subsubsection{Non-corporal sensor deployments}

\subsubsection{On-body sensor placement and deployments}
\begin{figure}[h!]
    \centering
\subfloat[]{
    \includegraphics[height=5cm]{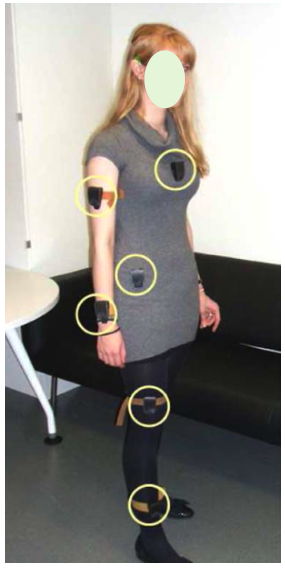}
}
\subfloat[]{
    \includegraphics[height=5cm]{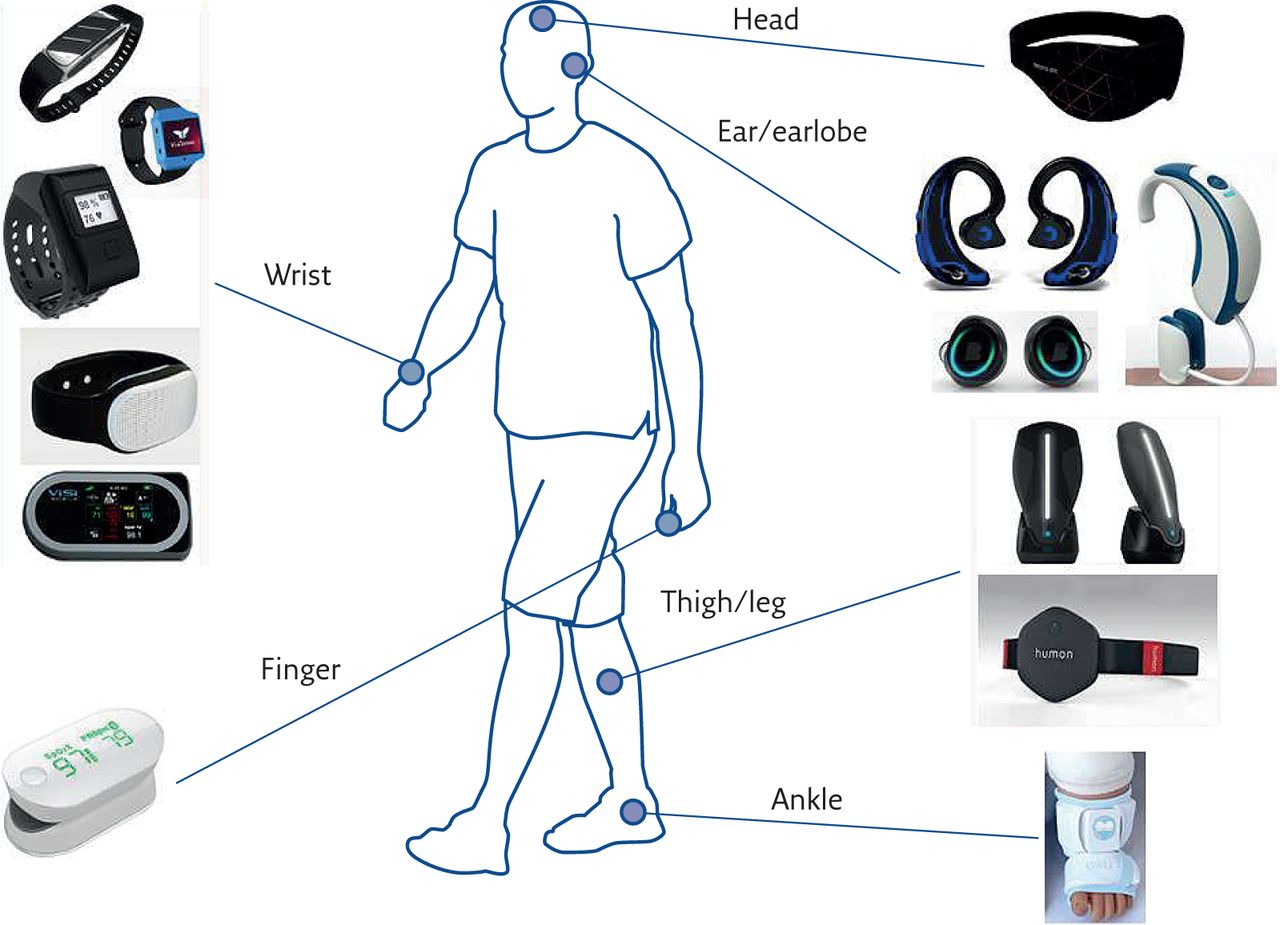}
}
    \caption{Various locations of wearable devices on the body. (a) illustration of a subject wearing the wearable sensors on the ear, chest, arm, wrist, waist, knee, and ankle (from~\cite{atallah2011sensor}). (b) example of concrete wearable devices along with their on-body location (from~\cite{aliverti2017wearable}).}
    \label{fig:wearable_devices_location}
\end{figure}

Within the framework of the recognition of human activities, sensors are generally placed on the following body positions : waist, thigh, necklace, wrist, chest, hip, lower back, trunk, shanks, ankle, pocket, hand, back pack, torso, ear, etc (see  ~\ref{fig:wearable_devices_location}). A long line of research work has focused on the problem of optimal placement and combination of sensors on the body in order to achieve satisfactory levels of recognition, and many reviews report on this such as~\cite{atallah2011sensor,attal2015physical}. As an example, Gjoreski et al.~\cite{gjoreski2011accelerometer} studied the optimal location of accelerometers for posture recognition and fall detection.  Nine  placements  of  up  to  four sensors were considered: on the waist, chest, thigh and ankle. They highlighted that three accelerometers proved  sufficient  to  correctly  recognize  all  the  events  except  one (a slow fall). Additionnally, one  accelerometer  was  able  to  recognize only  the  most  clear-cut  fall while two  accelerometers  achieved  over 90\%  accuracy  of  posture  recognition,  which  was  better  than  a detection system based on location. More generally as reviewed in \cite{attal2015physical}, several works (e.g.~\cite{karantonis2006implementation,mathie2004classification,parkka2006activity,yang2008using}) provided empirical evidence on the substantial improvements obtained using accelerometer placed on the waist for the recognition of many activities such as sitting, standing, walking, lying in various positions, running, stairs ascent and descent, vacuuming and scrubbing.

In many empirical evaluations comparing multi-sensor versus single-sensor deployments for activity recognition, e.g.~\cite{bao2004activity,gao2014evaluation}, the settings leveraging multiple sensors tend to perform far better than their counterparts. However, different on-body locations and their various combinations for activity recognition lead to varying performances and no consensus tends to emerge.

\subsection{Sensing capabilities (coverage)}
\begin{figure}[h!]
    \centering
    \includegraphics[width=8cm]{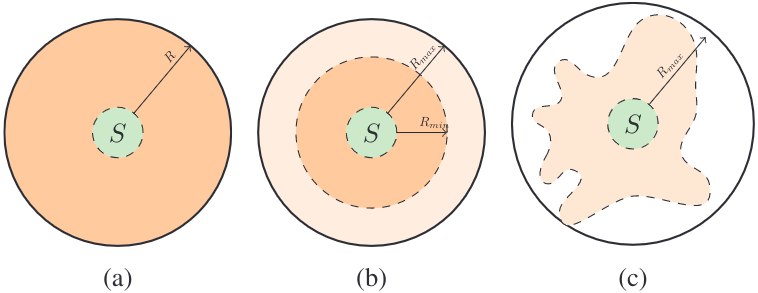}
    \caption{
    The shape of the sensing area for different sensing models:  (a) Deterministic sensing model, (b) Elfes sensing model, and (c) shadow fading sensing model (from~\cite{elhabyan2019coverage}).
    }
    \label{fig:sensing_models}
\end{figure}

In WSNs, each sensor node has a limited sensing range, and hence can only cover a limited physical area of the network field. Sensing models are abstraction models that are used to reflect the sensors' sensing ability and quality [25].  The sensing models can be classified, based on the direction of the sensing range, into either directional (or omnidirectional) sensing models.  Moreover, based on the sensing ability, sensing models are broadly classified into two types: Deterministic and probabilistic sensing models~\cite{elhabyan2019coverage} (see 
figure~\ref{fig:sensing_models}). 

\subsection{Sensor deployment topology (The collective dimension of the sensors)}
The sensor deployment topology is very important for activity recognition models. It defines the coverage model for optimal data acquisition. It ensures redundancy, robustness and data security. It is also important in WSNs by its impact on node energy, communication bandwidth and quality of service (QoS) \cite{han2019survey}.
\subsection{Variety of sensing modalities}
In addition to the on-body sensor placement which we saw above impacting substantially the performances of activity recognition models, the sensing modalities, such as acceleration, gravity, ambient pressure, etc. are also impactful and thus of utmost importance for the design of sensor-rich environments.
In a similar manner with the on-body sensor placement, sensing modalities are found to be beneficial when sensor-rich environments provide a multitude of them simultaneously. %Combination of these sensing modalities are also found to provide varying contribution to the activity recognition models.
One of the predominant sensing modalities used in the literature is obviously the acceleration which gained consensus among the empirical studies conducted around activity recognition~\cite{banos2014dealing,bao2004activity,kern2003wearable}.
In the other hand, various research works investigated the impact of combining different other modalities~\cite{ward2005gesture,stiefmeier2006combining,pentney2007common}.
Authors in~\cite{pentney2007common} for example studied activity recognition using a setting that includes eight sensors: a six-degree-of-freedom accelerometer, microphones sampling 8-bit audio at 16kHz, IR/visible light, high-frequency light, barometric pressure, humidity, temperature and compass.
In~\cite{stiefmeier2006combining}, motion sensors (accelerometers,  gyroscopes  and  magnetic field  sensors) have been combined with ultrasonic transmitters to track hands for activity recognition in a maintenance scenario. This combination have 

\subsection{Challenges in sensor-rich environments}\label{sec:challenges-in-sensor-rich-environments}
{\bf Reconciling various views/perspectives.} The placement of sensors makes it possible to give various perspectives and the use of several modalities makes it possible to give several points of view. Here the problem is to define for each modality the appropriate locations \cite{hamidi2020data}.
The problem is more complex than in the case of non-body deployments because the positions of the sensors between them change according to the movements. This can generate ambiguity and misinterpretations if the relative movements of the sensors between them are not taken into account.

{\bf Sensors placement and displacement.}
Even if the wearable sensors should be correctly attached to the body, vibration or displacement of those sensors cause signal interference and thus deterioration of the measurement accuracy~\cite{attal2015physical}. Various studies were conducted in the literature and different approaches were proposed to cope with these issues~\cite{forster2009evolving,kunze2008dealing,banos2014dealing,attal2015physical,shi2020sensor,barshan2020classifying}.
For example, authors in~\cite{kunze2008dealing} proposed a set of heuristics that significantly increase the robustness of motion sensor-based activity recognition with respect to sensor displacement. In particular, they show how, within certain limits and with modest quality degradation, motion sensor based activity recognition can be implemented in a displacement tolerant way.
In~\cite{banos2014dealing}, authors explored the effects of sensor displacement induced by both the intentional misplacement of sensors and self-placement by the user. The effects of sensor displacement are analyzed for standard activity recognition techniques, as well as for an alternate robust sensor fusion method proposed at the same occasion.

{\bf Heterogeneity of deployments.}
Another problem is related to the lack of interoperability among different sensor deployments. This problem is, in particular, due to the existence of different incompatible solutions (owners and non-owners). This makes it difficult both to integrate new deployments and their constant evolution ~\cite{mainetti2011evolution, stisen2015smart, baldominos2018evolutionary}. Another source of heterogeneity is related to the incompatibility of detection solutions. In~\cite{stisen2015smart}, authors investigated in a systematic manner sensor-, device- and workload-specific heterogeneities using 36 smartphones and smartwatches, consisting of 13 different device models from four manufacturers. Their results indicate that on-device sensor and sensor handling heterogeneities impair significantly the performances of activity recognition models.
\section{Constraints Related to the Sensing Nodes}\label{sec:contraintes-liees-aux-objets-connectes}
In addition to topological and positioning constraints, other constraints have a significant impact on the concepts to be learned. Among them we can cite: sensing constraints which impact the sensed measurements (Sect.~\ref{sec:sensing-constraints}), the energy and computational constraints related to the individual sensing nodes and how they impact the measurement (sampling frequency, etc.) as well as the transmission aspects (Sect.~\ref{sec:energy-computational-constraints}) and constraints linked to the sensing nodes taken collectively and in particular the transmission issues that arise in body sensor networks as well as in device-free settings (Sect.~\ref{sec:transmission-constraints}).
%_____
\subsection{Sensing constraints}\label{sec:sensing-constraints}
The performance characteristics of a sensor is equally (or more) important as its basic function which is to sense the phenomenon of interest~\cite{ida2014sensors}.
The choice of an appropriate sensing device and it performance characteristics for a given application is one of the most important issues sensor-rich environment designers are faced with. These aspects are discussed in~\cite{ida2014sensors}. As our focus is on activity recognition which involves very often the acceleration modality, after recalling the performance characteristics, we illustrate some of its components on concrete accelerometers which have been previously investigated in~\cite{albarbar2008suitability}.
\subsubsection{Performance characteristics of sensors}
The characteristics of a device start with its transfer function, that is, the relation between its input and output. This includes many other properties, such as \textit{span (or range)}, \textit{frequency response}, \textit{accuracy}, \textit{repeatability}, \textit{sensitivity}, \textit{linearity}, \textit{reliability}, and \textit{resolution}, among others.
Of course, not all are equally important in all sensors and actuators, and often the choice of properties will depend on the application~\cite{ida2014sensors}. Depending on many different factors, sensing characteristics may vary substantially. In mobile computing, for example, in order to keep the overall cost low, mobile devices are often equipped with low cost sensors, which are often poorly calibrated \cite{trusov2013silicon}, inaccurate, and of limited granularity and range, compared to dedicated sensors used in activity recognition models, e.g., a dedicated standalone inertial measurement unit~\cite{stisen2015smart,stisen2015smart,de2015uncertainty}.
\subsubsection{Accelerometers case}
The accelerometer factors are listed in~\cite{albarbar2008suitability}. The main of them are
the \textit{sensitivity} defines the ratio of its electrical output to its mechanical input, the \textit{amplitude limit} specifies the  maximum range of acceleration that can be measured, the
\textit{shock limit}, the \textit{natural frequency}, the \textit{resolution}, the \textit{frequency range}, and the \textit{phase shift} defining the time delay between the mechanical input and the corresponding electrical output signal of the instrumentation system.

\begin{figure}[h!]
    \centering
    \includegraphics[width=10cm]{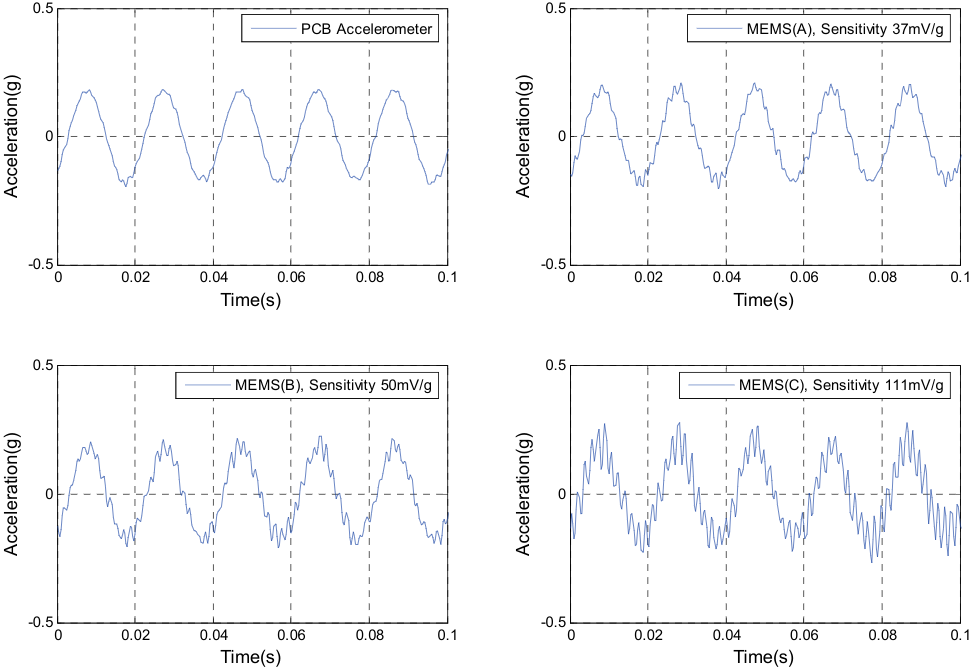}
    \caption{Measured acceleration responses by the MEMS accelerometers (A, B, and C) and the reference (PCB) accelerometer at 53Hz for the excitation amplitude 0.15g (from~\cite{albarbar2008suitability}).}
    \label{fig:measured-acceleration-responses-by-MEMS-accelerometers}
\end{figure}
According to the aforementioned factors, authors in~\cite{albarbar2008suitability} investigated the response generated by four MEMS accelerometers with various characteristics.
Figure~\ref{fig:measured-acceleration-responses-by-MEMS-accelerometers} summarizes the measured acceleration responses with various characteristics compared to the reference accelerometer.
Obtained results indicate that, in some configurations, there are lots of noise including extra un-interpretable peaks when compared against the reference accelerometer and against the remaining ones. A substantial  noise  and  shift in phase are also observed.

\subsection{Energy and Computational Constraints}\label{sec:energy-computational-constraints}
In addition to the sensory elements, sensory nodes are part of platforms which provide, in particular, computational as well as energy resources. These aspects are discussed from the perspective of the constraints they impose on the measurements acquisition process as a whole.

%__
\subsubsection{Energy constraints and solutions}
Among many considerations, energy efficiency is a key issue in enabling long-term monitoring~\cite{wang2013energy}.
A major limitation of untethered nodes is finite battery capacity—nodes will operate for a finite duration, only as long as the battery lasts. Finite node lifetime implies finite lifetime of the applications or additional cost and complexity to regularly change batteries. Nodes could possibly use large batteries for longer lifetimes, but will have to deal with increased size, weight and cost. Nodes may also opt to use low-power hardware like a low-power processor and radio, at the cost of lesser computation ability and lower transmission ranges~\cite{sudevalayam2010energy}
The wireless communication is likely to be the most power  consuming.  The  power  available  in  the  nodes  is often  restricted.  The  size  of  the  battery  used  to  store  the needed energy is in most cases the largest contributor to the sensor  device  in  terms  of  both  dimensions  and  weight.Batteries  are,  as  a  consequence,  kept  small  and  energy consumption of the devices needs to be reduced~\cite{latre2011survey}.
In addition, during communication the devices produce heat which is  absorbed  by  the  surrounding  tissue  and  increases  the temperature of the body. In order to limit this temperature rise  and  in  addition  to  save  the  battery  resources,  the energy  consumption  should  be  restricted  to  a  minimum~\cite{latre2011survey}.

\textbf{Energy-aware architecture design.}
In order to implement long term monitoring functions, energy control (low-power architecture  design, low-power processor design, low-power transceiver design, etc.) is one of the hot topics in  the  field  of BSN sensors. 

\textbf{Energy-aware routing protocols.}
Authors in~\cite{movassaghi2012energy} presented an energy efficient, thermal and power aware routing algorithm for BANs named Energy Efficient Thermal and Power Aware routing (ETPA). ETPA considers a node’s temperature, energy level and received power from adjacent nodes in the cost function calculation. An optimization problem is also defined in order to minimize average temperature rise in the network. ETPA can significantly decrease temperature rise and power consumption as well as providing a more efficient usage of the available resources.~\cite{movassaghi2012energy}.\\
In~\cite{oey2013survey}, the authors presented an investigation of temperature sensitive routing protocols in wireless body sensor networks for which temperature and heat production are fundamental. These routing protocols take the temperature of the node as a metric in the decision of the routing path. The purpose is to keep the temperature of the node below the safe level and slow down the rate of temperature rise, so that it does not harm the human body~\cite{oey2013survey}. 

\begin{figure}[h!]
    \centering
    \includegraphics[width=6cm]{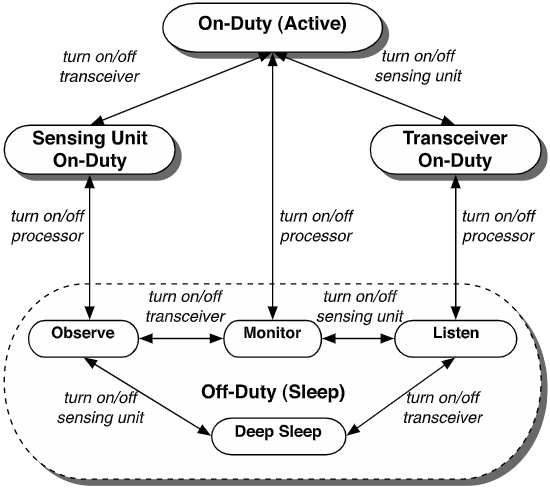}
    \caption{Transitions between different sensor modes (from~\cite{wang2006survey}).}
    \label{fig:transitions-between-different-sensor-modes}
\end{figure}

Authors in~\cite{rault2017survey} reviewed the literature around energy-efficient solutions in wearable sensor networks for activity recognition. The reviewed solutions have been categorized by authors into four main axes namely, \textit{power-on time reduction}, \textit{communication reduction}, \textit{computation reduction} and \textit{battery charging}.
Power-on time reduction schemes switch the nodes to the sleep mode. In~\cite{wang2006survey}, authors organize the sensors' energy saving modes into four major modes ( {\it on-duty}, {\it sensing unit on-duty}, {\it transceiver on-duty}, and {\it off-duty}. Figure~\ref{fig:transitions-between-different-sensor-modes} shows the transition between these modes. 
Selection of network interfaces was also investigated, where the radio used to transmit is selected depending on the environment opportunities(bandwidth, link quality, energy).
Some studies consider sensors equipped with two radios: a low-power, low throughput radio, and a high-power, high throughput radio. Indeed, higher throughput radios have a lower energy-per-bit cost, but they also have higher start up time and cost. Therefore, high bandwidth radios become more energy efficient only when a large number of byte have to be transmitted, which compensate the wakeup energy overhead~\cite{rault2017survey}.
\subsubsection{Computational constraints}
Computational constraints are often related to the \textbf{energy constraints} described above.
Indeed, the less energy a given sensory platform has, the less time it can spend to perform computations. An appropriate trade-offs between energy consumption and computational quality must be found for each application. Authors in~\cite{wang2013energy}, investigated  the  trade-offs between classification  accuracy and energy efficiency by comparing on-and off-node schemes. An empirical energy model is presented  and used to evaluate the energy efficiency of both systems, and a practical case  study (monitoring  the physical activities of office workers) is developed to evaluate the effect  on classification  accuracy.  The  results  show  a  40\%  energy  saving can be obtained with a 13\% reduction in classification accuracy.
Similarly, with the goal of analyzing the trade-off between recognition accuracy and computational complexity, authors in~\cite{maurer2006activity} investigated the impact of different sampling rates and other parameters on the performance of activity recognition models.
%on which axes the impact is being more perceptible?

%Reduction of impact of these constraints on the sampling frequency, near real-time availability of measurements, etc. \dots
%
\subsection{Transmission Constraints}\label{sec:transmission-constraints}
Although BSN are considered as a special case of wireless sensor networks (WSN), the mechanisms and WSN communication protocols are rarely applicable. After a brief comparison between the two families of networks, we will focus on some problems related to BSNs such as the organization of protocol layers. We did not find any work highlighting the definition of BSN protocols and their impact on activity recognition models. We have therefore chosen to deal with this problem. 
\subsubsection{Body sensor networks versus wireless sensor networks}
\begin{figure}[h!]
    \centering
%\subfloat[]{
    \includegraphics[height=6cm]{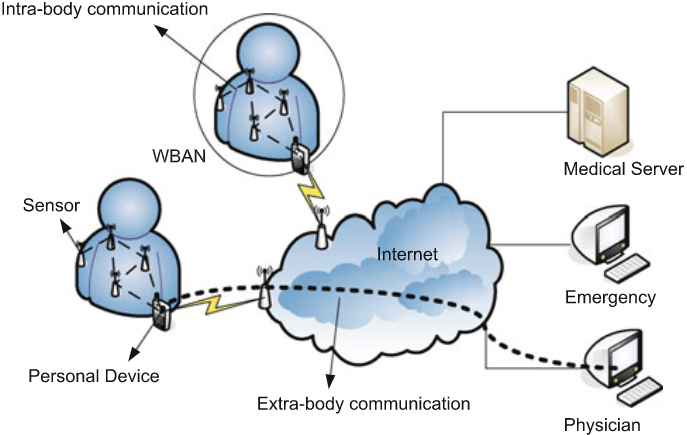}
%}

%\subfloat[]{
    \includegraphics[height=6cm]{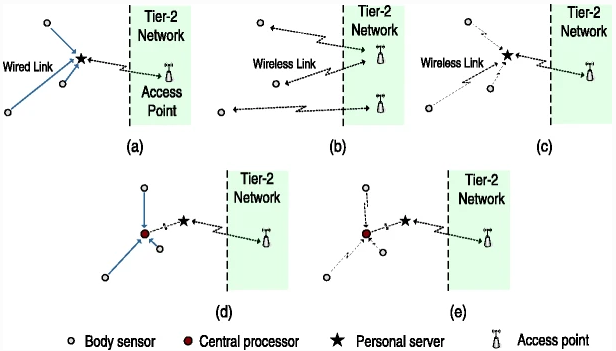}
%}
    \caption{(top) Illustration of intra-body and extra-body communication in a wireless body area network in the context of medical applications (from~\cite{latre2011survey}). (bottom) Architecture of intra-body communication: (a) wired; (b) directly connected to access point; (c) wireless; (d) hybrid; (e) cluster and wireless (from~\cite{chen2011body}).}
    \label{fig:example-of-intra-body-and-extra-body-communication-in-a-WBAN}
\end{figure}

The Deployment that we have seen until here are studied in the literature which is referred to as (wireless) body area/sensor network.
It is defined in~\cite{ullah2012comprehensive} as a technology allowing the integration of intelligent, miniaturized, low-power sensor nodes in, on, or around a human body to monitor body functions and the surrounding.
This definition allows us to highlight some intrinsic characteristics related to size, energy, and deployment topology.  Additionally, As  the  devices  get  smaller  and  more  ubiquitous,  a direct connection to the personal device will no longer be possible  and  more  complex  network  topologies  will  be needed~\cite{latre2011survey}.
In  literature,  protocols  developed for WBANs can span from communication between the  sensors  on  the  body (intra-body)  to  communication  from  a  body node to a data center connected to the Internet (extra-body). The  former controls the information handling on the body between the sensors  or  actuators  and  the  personal  device,  the latter ensures communication between the personal device and an external network~\cite{latre2011survey}.
Figure~\ref{fig:example-of-intra-body-and-extra-body-communication-in-a-WBAN} illustrates an example of a wireless body area network in the context of medical monitoring applications and different configurations of intra-body communication.

Although wireless BSNs are considered as a special type of WSNs and the challenges faced by both types of networks are similar in many respects, the transmission mechanisms and the protocols developed for WSNs are not well suited to support communication in BSNs~\cite{latre2011survey}. Indeed, the BSNs have a smaller scale compared to Wireless Sensor Networks (WSNs) and the human body consists of a complicated internal environment that responds to and interacts with its external  surroundings. 
Sensor  nodes  can  move  with  regard  to each other. For example a sensor node placed on the wrist moves in relation to yhe one attached to the hip. This requires  mobility  support.

Beyond that, BSNs and WSNs contrasts also in terms of the density of the sensing devices (in particular, BSNs do not employ redundant nodes to cope with diverse types of failures), in terms of data rates or 
latency.
\subsubsection{Challenges in body sensor networks}
Challenges related to BSNs span the whole networking stack including from Network, Medium Access Control (MAC), and Physical (PHY) layers. These challenges and the main constraints that stem from them are related to the quality of service as well as to the low-power requirements. The later largely constrain every aspect of the protocols that are being constantly developed for BSNs.

\textbf{Network layers.}
Network layers consists of the routing protocols responsible of addressing and structuring multi-node networks as well as routing packets.
Specific characteristics as enumerated in~\cite{ullah2012comprehensive}.
(1) The available bandwidth is limited, shared and can vary due to fading, noise and interference. As a result, the network control generated by the protocol should be limited; (2) The nodes that form the network can be very heterogeneous in terms of available energy or computing power. As a result, node energy should also be taken into account; (3) An extremely low transmit power per node is needed to minimize interference to cope with health concerns and to avoid tissue heating; (4) The devices are located on the human body that can be in motion. WBANs should therefore be robust against frequent changes in the network topology.
A lot of research is being done towards energy efficient routing in ad hoc networks and WSNs but the proposed solutions are inadequate for WBANs. For example, in WSNs maximal throughput and minimal routing overhead are considered to be more important than minimal energy consumption.
Network topology is defined as the logical organization or arrangement of communication devices in the system. The selection of a proper network topology in WBAN is important as it significantly affects the overall system performance and protocol design. It influences the system in many ways, e.g., in power consumption, the ability to handle heterogeneity, the robustness against failures and the routing of data, etc.

%\bigskip
\textbf{Medium access control layers.}
Medium access control layers control how devices gain access to a medium in order to transmit data.
In WBAN, the RF part of the sensor consumes most of the energy and hence becomes one of most important entities  to  be  considered.  The  MAC  protocol  plays  a significant role in controlling/duty cycling the RF module  and  in  reducing  the  average  energy  consumption of the sensor node. In other words, the MAC protocol is required to achieve maximum throughput, minimum delay, and to maximize the network lifetime by control-ling the main sources of energy waste, i.e., collision, idle listening, overhearing, and control packet overhead.
Many different low-power mechanisms exist in the literature such as Low Power Listening  (LPL), Contention and scheduled-contention, and Time Division Multiple Access (TDMA)  mechanisms for WBAN. Besides, many different protocols based on these mechanisms were proposed in the literature such as IEEE 802.15.4 MAC protocol, battery-aware TDMA protocol, energy-efficient TDMA-based MAC protocol,  heart-beat  driven  MAC,  reservation-based Dynamic TDMA (DTDMA), and BodyMAC. These protocols have been summarized in~\cite{ullah2012comprehensive}.

\textbf{Physical layers.}
Physical layers consists of the in-body RF communication and the propagation pattern in or around a human body. This layer is responsible for the transmission and reception of unstructured raw data between a device and a physical transmission medium.
The wireless radio channel poses a severe challenge as a medium for reliable high-speed communication. Not only is it susceptible to noise, interference, and other channel impediments, but these impediments change over time in unpredictable ways as a result of user movement and environment dynamics~\cite{goldsmith2005path}.
The problem of path loss is prominent in BANs making them unstable and temporary subject to high packet error rate~\cite{gorce2009opportunistic}.
We focus here on the issues related to signal propagation, noticeably path loss and its relation to (i) the disposition of sensing nodes, (ii) body movements, and (iii) the surrounding environment.

\begin{figure}[h!]
    \centering
\subfloat[]{
    \includegraphics[height=3cm]{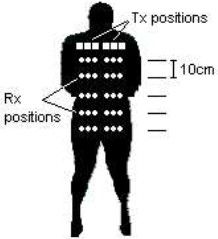}
}
\subfloat[]{
    \includegraphics[height=3cm]{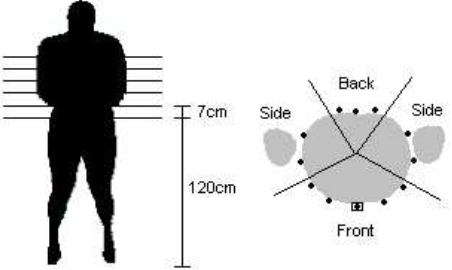}
}
\subfloat[]{
    \includegraphics[height=3cm]{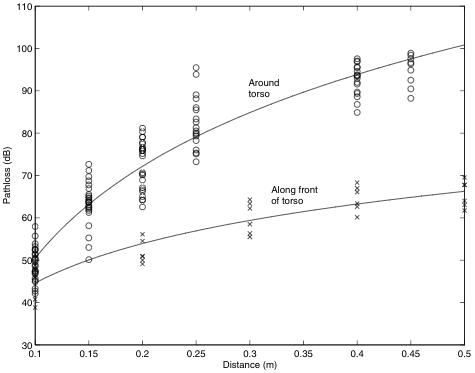}
}
    \caption{(a) Measurement locations on body (along the torso). (b) Measurement locations on body (around the torso). (c) Measure pathloss around the body. (From~\cite{fort2005characterization})}
    \label{fig:Characterization-of-the-Ultra-Wide-band-Body-Area-Propagation-Channel}
\end{figure}

\underline{(i) impact of the disposition of sensing nodes on the path loss:}
The characteristics of the physical layer are different for a WBAN compared to a regular sensor network or an ad-hoc network  due  to  the  proximity  of  the  human  body.
radio propagations from devices that are close to or inside the human body are complex and distinctive comparing to the other environments since the human body has a complex shape consisting of different tissues. Therefore, the channel models are different from the ones in the other environments~\cite{taparugssanagorn2008review}
Various studies have been carried using different models of transceivers and showed a lock of communications among nodes depending on their on-body locations~\cite{latre2011survey}.
For examples, in~\cite{ruzzelli2007energy,shah2006characteristics}, experimented with 802.15.4-based CC2420 transceivers placed in different parts of the body including chest, ankle, and back of patients, etc. Results showed lots of variations in terms of communication among the nodes.
%Tests with TelosB motes (using the CC2420 transceiver) showed \textbf{lack  of  communications}  between  nodes  located  on  the chest  and  nodes  located  on  the  back  of  the  patient  [46]. This was accentuated when the transmit power was set to a minimum for energy savings reasons. Similar conclusions where  drawn  with  a  CC2420  transceiver  in  [47]:  when  a person was sitting on a sofa, no communication was possible between the chest and the ankle. Better results were obtained  when  the  antenna  was  placed  1 cm  above  the body.  In this section, we will discuss the characteristics of  the  propagation  of  radio  waves  in  a  WBAN  and  other types of communication"~\cite{latre2011survey}
Figure~\ref{fig:Characterization-of-the-Ultra-Wide-band-Body-Area-Propagation-Channel} illustrates the impact of the transceivers' on-body locations on the path loss.
Additionally, in~\cite{gorce2009opportunistic}, authors discussed the problem of path loss with respect to the underlying network topology, noticeably star vs. multi-hope mesh, where a reduction of the emitter-receiver distance could counteract this problem. %A multi-hop mesh topology  cannot  counteract  this  problem  efficiently,  since  the path loss attenuation in a BAN environment is almost independent with the emitter-receiver distance."~\cite{gorce2009opportunistic}

\begin{figure}[h!]
    \centering
%\subfloat[]{
%    \includegraphics[width=5cm]{img/screenshot_384.png}
%    \label{fig:slow-component}
%}
%\subfloat[]{
    \includegraphics[width=9cm]{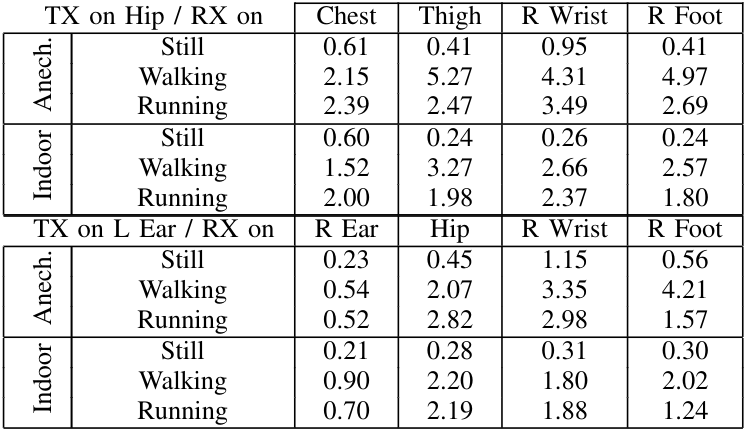}
%}
    \caption{Shadowing standard deviation (from~\cite{d2009time}).}
    \label{fig:shadowing-standard-deviation}
\end{figure}

(ii) \underline{Impact of the body movements on the path loss:}
In addition to lack of communication due to the deployment topology, authors in~\cite{fort2005characterization} studied the \textbf{influence of arm motions} on the pathloss.
Similarly, authors in~\cite{d2009time} studied the \textbf{impact of various types of activities} (still, walking, and running) on the path-loss depending on the location of the transceivers. 
In~\cite{d2009time}, three different human body movements for seven human subjects were considered to assess the influence of human activity on the channel behavior. To this aim, an analysis on mean channel gain, slow fading and shadowing correlation was presented with emphasis on the differences given by the human body variability and the movement condition.
On this matter, Table~\ref{fig:shadowing-standard-deviation} illustrates the shadowing standard deviation depending on the respective position of transmitters and receivers.

(iii) \underline{Impact of the surrounding environment:} Authors in~\cite{fort2005characterization} also studied signal propagation by taking into account factors related to the environment in which the user operates.
These include, for example, the influence of ground reflections, considered more reliable to be exploited during transmission, as well as reflections from surrounding environments on received signals. 
\section{Dynamic inductive bias selection}\label{sec:dynamic-inductive-bias-selection}
The measurement of a phenomenon as simple as temperature through a sensor is in itself an inductive process involving many biases.
Indeed, the action of the physico-electrical process of the sensor generates an electrical signal proportional to the physical phenomenon being measured.
We, actually, do not have access to the physical phenomenon itself but to a representation provided through a transfer function deduced mathematically and which is specific to the physico-electrical process of the sensor.
The choice of this process constitutes a bias similarly to the elaboration of the transfer function.

Also called the transfer characteristic function, the input/output characteristic function or response of a device is a relationship between the output and input of the device, usually defined by some kind of mathematical equation and a descriptive curve or graphical representation in a given range of inputs and outputs. The function may be linear or nonlinear, single valued or multivalued. It defines the response of a sensor or actuator to a given input or set of inputs and is one of the main parameters used in design. With the exception of linear transfer functions, it is usually difficult to describe the transfer function mathematically~\cite{ida2014sensors}.

\begin{figure}[h!]
    \centering
    \includegraphics[width=7cm]{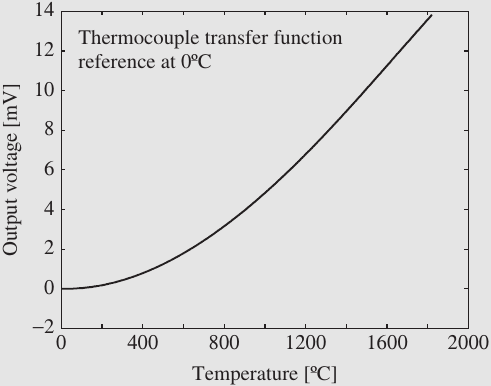}
    \caption{Transfer function of the thermocouple in the range 0\textcelsius{}–1820\textcelsius{}. The output is shown in volts (V), although the polynomial gives it in millivolts (mV) (from~\cite{ida2014sensors}).
    }
    \label{fig:transfer-function-of-the-thermocouple}
\end{figure}

The following example derived from~\cite{ida2014sensors} illustrates the notion of transfer function on a temperature sensor.
The output (voltage) of a thermocouple (temperature sensor) for a given temperature is given by a polynomial that can range from a 3rd order to a 12th order polynomial depending on the type of thermocouple. The output of a particular type of thermocouple is given by the following relation in the range 0\textcelsius{}–1820\textcelsius{}:
\begin{align*}
V = & (-2.4674601620 \times 10^{-1} \times T + 5.9102111169 \times 10^{-3} \times T^{2} \\
    & - 1.4307123430 \times 10^{-6} \times T^{3} + 2.1509149750 \times 10^{-9} \times T^{4} \\
    & - 3.1757800720 \times 10^{-12} \times T^{5} + 2.4010367459 \times 10^{-15} \times T^{6} \\
    & - 9.0928148159 \times 10^{-19} \times T^{7} + 1.3299505137 \times 10^{-22} \times T^{8}) \times 10^{-3} mV
\end{align*}
This is a rather involved transfer function (most sensors will have a much simpler response) and it is nonlinear. The main purpose of the elaborate function is to provide very accurate representation over the range of the sensor (in this case 0\textcelsius{}–1820\textcelsius{}). as shown in figure \ref{fig:transfer-function-of-the-thermocouple}.

On the other hand, much more complex phenomena, such as activity recognition, require much more sophisticated inductive processes, which can be handled by machine learning approaches in which the definition of adequate inductive biases (such as the pre-processing filter or the segmentation step) is necessary in order to make the learning of these phenomena possible.
However, as we have just seen through the previous sections, the sensor deployments are characterized at every level (from the measurement to the type of transmission protocol) by biases, which substantially influence the final representation of the phenomena of interest.\\
In the following, we will first recall the components of the known activity recognition chain~\cite{bulling2014tutorial}  into the perspective of inductive bias learning and the need for their dynamic selection (Sect.~\ref{sec:background-supervised-learning}).
We, then, present a background on dynamic inductive bias selection~\cite{baxter2000model} and an overview of the long line of research on this paradigm (Sect.~\ref{sec:background-on-dynamic-inductive-bias-selection}).
Finally, we turn into one instantiation of the dynamic selection of inductive bias paradigm, {\it surrogate models}. In our use-case, we encode models of the deployments as well as those of the phenomena into surrogate models (Sect.~\ref{sec:surrogate-models}).
\subsection{Background on supervised learning}\label{sec:background-supervised-learning}
%"In this paper we introduce and analyze a formal model of bias learning that builds upon
%
According to the PAC model of machine learning and its variants~\cite{vapnik1982necessary,valiant1984theory,blumer1989learnability}, supervised learning models typically take the following general form: the learner is supplied with a hypothesis space $\mathcal{H}$ and training data $\{(x_1, y_1), \dots, (x_m, y_m)\}$ drawn independently according to some underlying distribution $P$ on $X \times Y$. Based on the information contained in the training data, the learner's goal is to select a hypothesis $h : X \xrightarrow[]{} Y$ from $\mathcal{H}$ minimizing some measure $er_{P}(h)$ of expected loss with respect to $P$ (for example, in the case of squared loss $er_{P} (h) := \mathbb{E}_{(x,y)\sim P} (h(x),  y)^2$). In such models the learner's bias is represented by the choice of $\mathcal{H}$;
if $\mathcal{H}$ does not contain a good solution to the problem, then, regardless of how much data the learner receives, it cannot learn~\cite{baxter2000model}.
%
%"Computational learning theory models of supervised learning usually include the following ingredients:
In general, models of supervised learning include: an input space $X$ and an output space $Y$, a probability distribution $P$ on $X \times Y$, a loss function $\ell : Y \times Y \xrightarrow[]{} \mathbb{R}$ (empirical risk minimization), and a hypothesis space $\mathcal{H}$ which is a set of hypotheses or functions $h : X \xrightarrow[]{} Y$.
%\begin{itemize}
 %   \item An input space $X$ and an output space $Y$,
 %   \item a probability distribution $P$ on $X \times Y$,  
 %   \item a loss function $\ell : Y \times Y \xrightarrow[]{} \mathbb{R}$, and
  %  \item a hypothesis space $\mathcal{H}$ which is a set of hypotheses or functions $h : X \xrightarrow[]{} Y$.
%\end{itemize}

In the case of human activity recognition, on possible mapping is $X$ would be the set of observations generated by the on-body sensor nodes, $Y$ would be the set of target activities (walk, run, etc.), and the distribution $P$ would be peaked over different episodes during which the users perform one of the target activities.
The learner’s hypothesis space $\mathcal{H}$ would be a class of neural networks mapping the input space $X$ to $Y$. The loss in this case would be discrete loss: $\ell(y, y') := \Bigl\{ \begin{array}{cc}\small
    1 & \text{if } y \neq y'  \\
    0 & \text{if } y = y' 
\end{array}
$

\begin{figure}[h!]
    \centering
    \sffamily
    \def\svgwidth{0.8\columnwidth}
    \resizebox{130mm}{!}{
        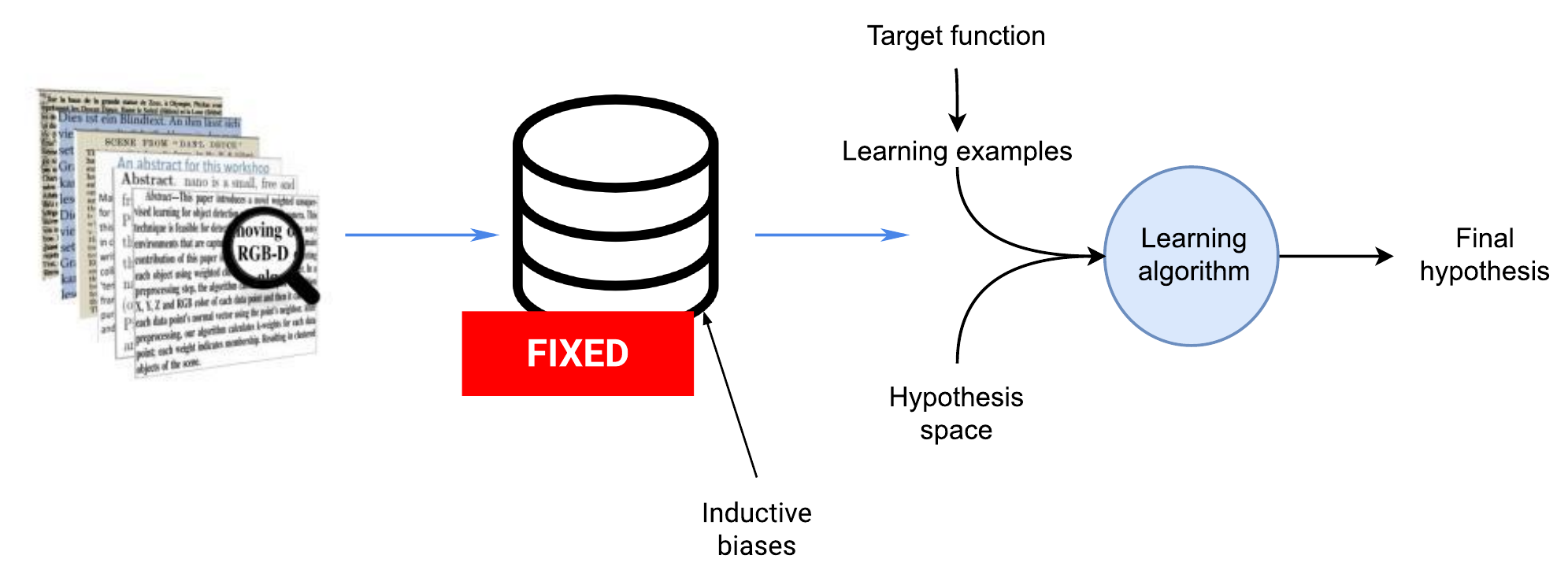
    }
    \caption{
        Basic learning setting where the learner is supplied with a fixed set of inductive biases.
        %Here, the inductive biases are fixed, which is symbolized by the repository that lies in-between the input space and the learner.
        The inductive biases guide the learner in searching for the hypothesis that explain best the set of learning examples.
    }
    \label{fig:basic-learning-setting}
\end{figure}

Figure~\ref{fig:basic-learning-setting} illustrates the basic learning setting where the learner is supplied with a fixed set of inductive biases. These inductive biases are the set of all factors that collectively influence hypothesis selection.
In the case of human activity recognition from a stream of observations, these factors include for example the pre-processing, segmentation, feature extraction, and other steps which are part of the activity recognition chain (see Section~\ref{sec:background-on-activity-recognition-chain}).
In addition to the definition of the space of hypothesis and the algorithm that searches for the optimal hypothesis,
Learning concepts from examples is depicted in Figure~\ref{fig:basic-learning-setting} as a function of two arguments, the training instances, and the bias for hypothesis preference.
The problem of selecting training examples is as important as the problem of selecting biases.
Rather, for any particular set of training instances, the biases guide the learner to choose a particular hypothesis. A program that learns concepts from examples is successful only when it has a bias that guides it to make a satisfactory selection from among the available hypotheses. Without the bias, the program has no basis for, electing one hypothesis in favor of another.
Two important features of bias are strength (reduction factor of hypothesis space) and correctness \cite{utgoff1986machine}.

%__
\subsection{Background on Activity Recognition Chain}\label{sec:background-on-activity-recognition-chain}
%
%The activity recognition chain~\cite{bulling2014tutorial} is a widely used machine learning-based inductive process in the literature. Used to model human activities, it is composed of five main steps: \textit{data acquisition}, \textit{preprocessing}, \textit{segmentation}, \textit{feature extraction}, and \textit{classification} step.

The activity recognition chain~\cite{bulling2014tutorial} is a widely used machine learning-based inductive process in the literature which is used to model human activities (our phenomenon of interest).
It is composed of five different steps: \textit{data acquisition}, \textit{preprocessing}, \textit{segmentation}, \textit{feature extraction}, and \textit{classification}.
\begin{figure}[h!]
    \centering
    \includegraphics[width=10cm]{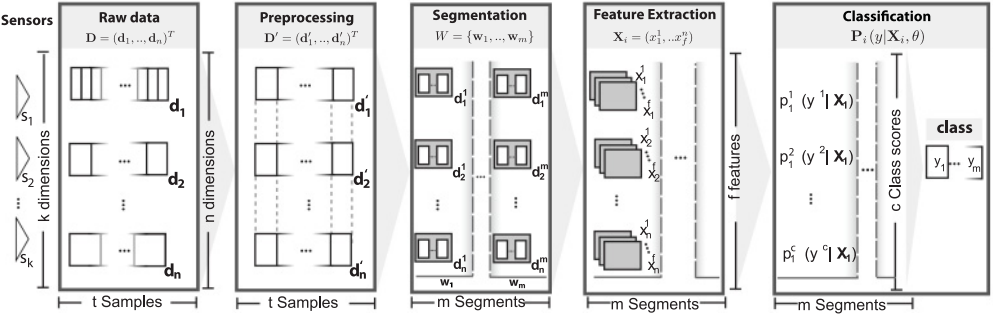}
    \caption{Activity recognition chain defined in~\cite{bulling2014tutorial} which encompasses various stages including data acquisition, signal preprocessing, segmentation, feature extraction, classification \& evaluation.
    %Raw signals (D) are first processed (D) and split into m segments (Wi) from which feature vectors (Xi) are extracted. Given features (Xi), a model with parameters $\theta$ scores c activity classes Yi = {y1,..., yc} with a confidence vector pi."
    }
    \label{fig:bulling-activity-recognition-chain}
\end{figure}

Given a collection $\mathcal{S}=\{s_1, \dots, s_M\}$ of $M$ sensors (also called data generators or data sources) carried by the user during daily activities to capture the body movements. Each sensor $s_i$ generates a stream $\mathbf{x}^{i} = (x_{1}^{i}, x_{2}^{i},\dots)$ of observations of a certain modality, which can be composed of several channels, e.g. the accelerometer modality contains tree channels ($x$, $y$, and $z$ axes).

Figure~\ref{fig:bulling-activity-recognition-chain} illustrates the steps of the activity recognition chain as defined in~\cite{bulling2014tutorial}. The goal of these steps, as presented in the following, is to build a model capable of recognizing human activities (outputs) from the streams of observations (inputs).
\subsubsection{Preprocessing}
In this step, the streams of observations generated by each sensor are being "enhanced", in some sense, and this in the perspective of features extraction. The preprocessing step makes the features extraction phase more robust.
In the case of device-free activity recognition approaches, authors in~\cite{wang2017device,wang2019survey} provide a comprehensive list of preprocessing methods widely used in the literature each of which suitable in different situations.
Some of these methods are High-pass filter (pre-emphasizing), Hampel filter~\cite{chowdhury2018using}, Phase sanitization~\cite{qian2018enabling}, phase calibration~\cite{zhou2016device}, Butterworth low-pass filter~\cite{arshad2017wi}, STFT (Heisenberg uncertainty principle~\cite{cohen1995uncertainty}), Savitzky-Golay filter~\cite{zhang2019towards}, and Birge-Massart filter~\cite{he2015wig}.

\begin{figure}[h!]
    \centering
    \includegraphics[width=9cm]{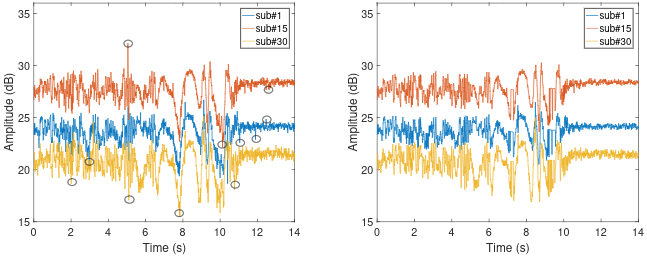}
    \caption{
    Noises generated by internal state transitions such as transmission power and rate adaptations, and thermal noises in the devices.
    Effect of Hampel Outlier Removal on three subcarriers:  (left) Raw CSI Amplitude waveforms with outliers denoted by ’black circles’, (right) Hampel filtered CSI amplitude waveforms (from~\cite{chowdhury2018using}).}
    \label{fig:effect-of-hampel-filter}
\end{figure}
%
%In addition to orientation normalization~\cite{}, many of these methods are also used in the case of inertial sensors-based activity recognition and similarly applicable depending on the situations. Low-pass filters for example are used to isolate gravity in the case of accelerometers and high-pass filters serve to remove drift in the case of gyroscopes~\cite{gjoreski2011accelerometer}.
%In the light of the constraints presented in Section~\ref{sec:contraintes-liees-aux-objets-connectes}, Figure~\ref{fig:effect-of-hampel-filter} illustrates the noises generated by internal state transitions such as transmission power and rate adaptations, and thermal noises in the devices.
%Depending on the situations, these effects require suitable methods that can cope with each situation in an optimal way.
%
\subsubsection{Segmentation}
\begin{figure}[h!]
    \centering
    \includegraphics[width=9cm]{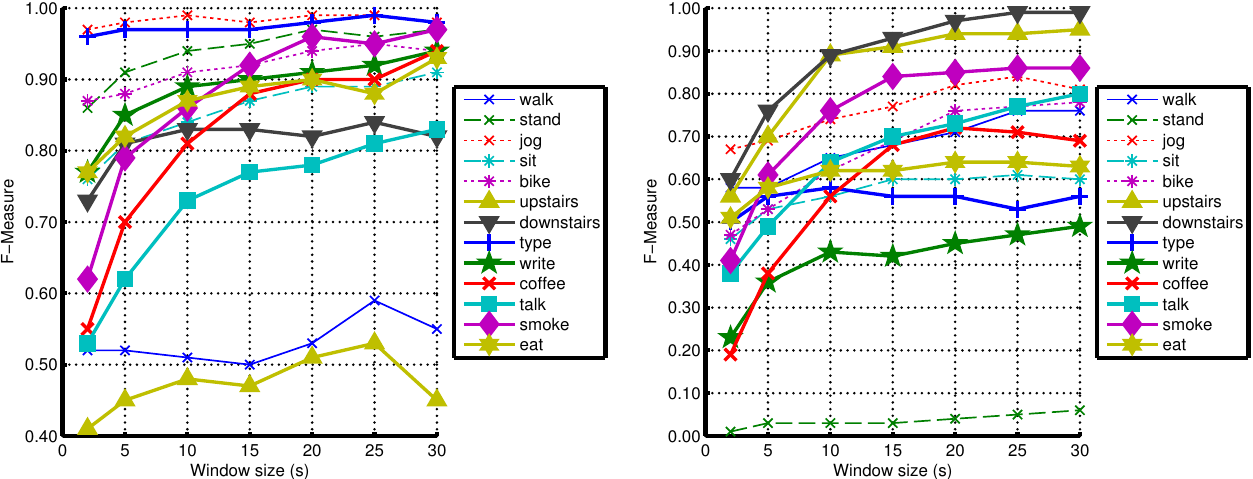}
    \caption{Effect of window size on recognition performances (from~\cite{shoaib2016complex}).}
    \label{fig:effect-of-window-size-on-recognition}
\end{figure}

During this step, the preprocessed streams of observations are divided into a set of segments which, depending on the segmentation procedure and its hyperparameters, likely contain the whole activity or parts of it.
Many different types of segmentation procedures exist in the literature around activity recognition and beyond including time-based, event-based, energy-based, etc.\cite{yala2015feature}.
Various works studied empirically the effects of different segment lengths on the recognition performances~\cite{shoaib2016complex,banos2014window}.
For example, Figure~\ref{fig:effect-of-window-size-on-recognition} shows the effect of window size on the performances (f-measure) of activity recognition models.

Issues with time-based segmentation are not circumscribed to the choice of the segment's length but are also tightly linked to the feature extraction step.
Activities that lasts for variable time constitutes an important issue.
For example, fixing the segment path can lead to spectral leakage which impacts the subsequent steps noticeably the feature extraction step from the spectral representation of the signal. Indeed, spectral leakage causes the spectrum to be noisy impacting the correct determination of frequencies, etc.
Issues go beyond the impact of the segment's length on the extracted features.
Many studies showed the impact related to the overlap of windows on the classification and evaluation steps~\cite{osmani2017platform,hammerla2015let} (see Section~\ref{sec:evaluation-and-model-selection}).
A growing line of research consider the issues that stem from the dynamic nature of the sensor deployments with regards to segmentation.
\subsubsection{Feature extraction}
Features are extracted from the preprocessed segments obtained in the previous steps and not from the whole streams of observations.
The resulting features are often impacted by the hyperparameters controlling the preceding steps. In~\cite{millecamps2015understanding}, for example, authors investigated the influence of pre-processing operations on features extracted from accelerometers in both time and frequency domains. Obtained results indicate that the preprocessing methods have to be carefully chosen as their impact is significant and disparate.
Another example is related to the impact of segmentation on the resulting frequency domain representation which is obtained using the short-time Fourier transform. Indeed, two effects at least can be mentioned: in the one hand, the tradeoff between resolution \& the Heisenberg uncertainty principle, in the other hand, spectral leakage~\cite{harris1978use}.
\subsubsection{Classification and evaluation}
The final step of the activity recognition chain consists of the classification of each individual segment of features, obtained before, into its correct class.
With regard to the PAC model of machine learning presented above, this step corresponds to electing a hypothesis that best explains the learning examples which are supplied along with the hypothesis space, i.e. the set of inductive biases ranging from the sensing and deployment models until the learning algorithm that we chose including the preprocessing, segmentation, and feature extraction steps.
For example, authors in~\cite{ma2016asynchronous} were interested in the highly dynamic nature of wearable sensor deployment, in the case of health monitoring, where changes in sensing platform (e.g., sensor upgrade) and platform   settings (e.g., sampling frequency, on-body sensor location) cause activity recognition models to degrade in terms of performances.
\subsection{Background on dynamic inductive bias selection}\label{sec:background-on-dynamic-inductive-bias-selection}
\begin{figure}[h!]
    \centering
    \sffamily
    \def\svgwidth{0.75\columnwidth}
    \resizebox{130mm}{!}{
        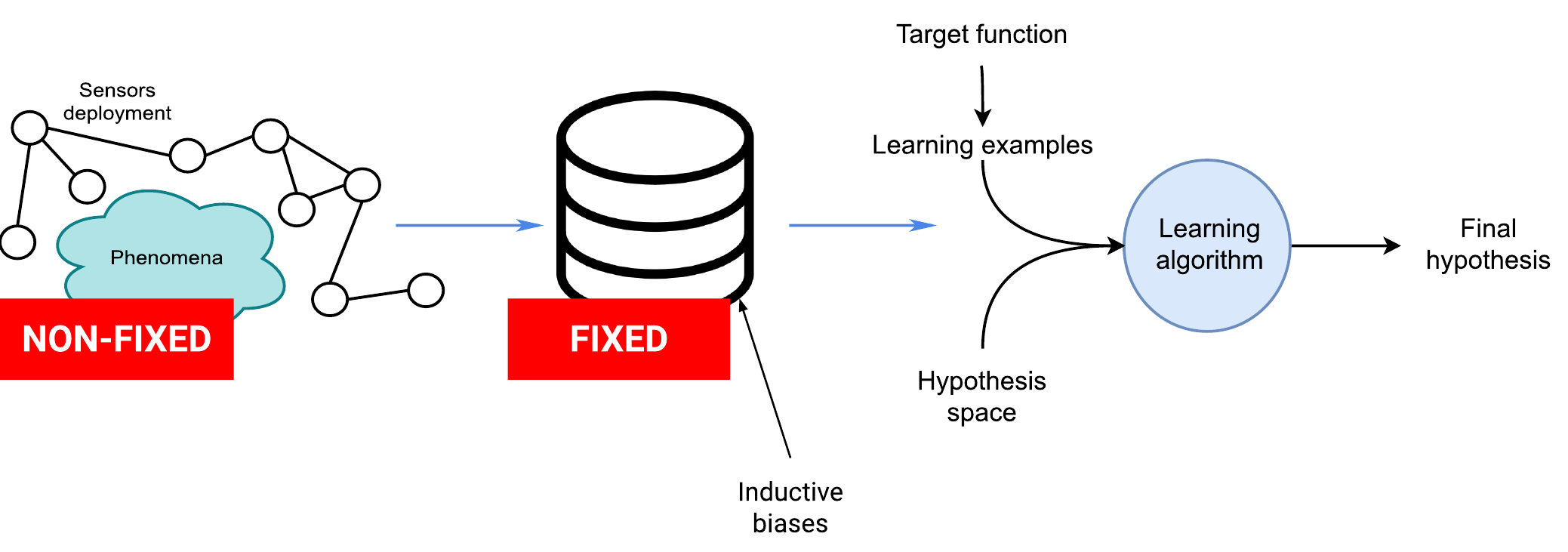
    }
    \caption{Learning setting in the case of sensor-rich environments deals with evolving and non-fixed settings materialized by the sensor deployments as well as the phenomena of interest. %The nature of the environments where the models will be deployed contradict the bias fixing paradigm. where rather than having a fixed learning setting complying with the bias fixing paradigm, we mainly 
    }
    \label{fig:dynamic-selection-of-inductive-biases}
\end{figure}

Sensor-rich environments are characterized by dynamicity.
For example,  sensors deployments often evolve and are subject to packets loss, heterogeneity, among many other issues.
While fixing inductive biases applying to specific problems can be advantageous in controlled environments, doing so during early steps of the activity recognition chain in such environments (see Figure~\ref{fig:dynamic-selection-of-inductive-biases}) leads inevitably to inefficient hypothesis space exploration but even worse, the final hypothesis that is elected may fail to explain the learning process.
A natural solution is to delay the selection of the inductive biases as late as possible and maintain concurrent hypotheses which can cope rapidly with new situations.
This leads to different implications operationally speaking, namely, maintaining a set of alternative inductive bias candidates (the domain) and exploring the space rapidly in order to elect the appropriate hypothesis (amount of supervision with learning examples).
In other words, the exploration of the hypothesis space should be structured by leveraging {\it a priori} knowledge about the sensor deployments and the phenomenon itself.

In~\cite{baxter2000model}, author proposed a model of bias learning where the learner can sample from multiple tasks, and hence it can search for a hypothesis space that contains good solutions to many of the problems in the environment.
Recall from the last paragraph of the previous section that the learner’s bias is represented by its choice of hypothesis space $\mathcal{H}$.
So to enable the learner to learn the bias, we supply it with a family or set of hypothesis spaces $\mathbb{H} := \{\mathcal{H}\}$.
Formally a learning to learn or bias learning problem consists of:
\begin{itemize}
    \item an input space $X$ and an output space $Y$ (both of which are separable metric spaces),
    \item a loss function $\ell : Y \times Y \xrightarrow[]{} \mathbb{R}$,
    \item an environment $(\mathcal{P}, Q)$ where $\mathcal{P}$ is the set of all probability distributions on $X \times Y$ and $Q$ is a distribution on $\mathcal{P}$,
    \item a hypothesis space family $\mathbb{H} = \{\mathcal{H}\}$ where each $\mathcal{H} \in \mathbb{H}$ is a set of functions $h : X \xrightarrow[]{} Y$.
\end{itemize}

In the bias learning model proposed in~\cite{baxter2000model}, the learner is embedded in an environment of related tasks, e.g. face recognition, character recognition, etc., and thus requiring fairly dissimilar inductive biases.
Here we rather consider learning configurations that describe the same phenomena (a same task) which evolve itself but also in terms of the sensor deployments used to capture it.
More formally, according to the notation in~\cite{baxter2000model} adapted to the problem we are interested in, the set of learning settings which are likely to be encountered in real-life deployments is represented by a pair $(\mathcal{P}, Q)$ where $\mathcal{P}$ is the set of all probability distributions on $X \times Y$ (i.e., $\mathcal{P}$ is the set of all possible learning problems), and $Q$ is a distribution on $\mathcal{P}$.
$Q$ controls the various scenarios which the activity recognition model will likely encounter in real-life deployment settings.
\subsection{Surrogate models}\label{sec:surrogate-models}
With the dynamic nature of sensor-rich environments we have to delay the selection of the inductive biases as late as possible and maintain concurrent hypotheses which can cope rapidly with new situations and we need models that learn and adapt quickly to new settings, new users, new activities, etc.

\begin{figure}[h!]
    \centering
    \sffamily
    \def\svgwidth{0.7\columnwidth}
    \resizebox{130mm}{!}{
        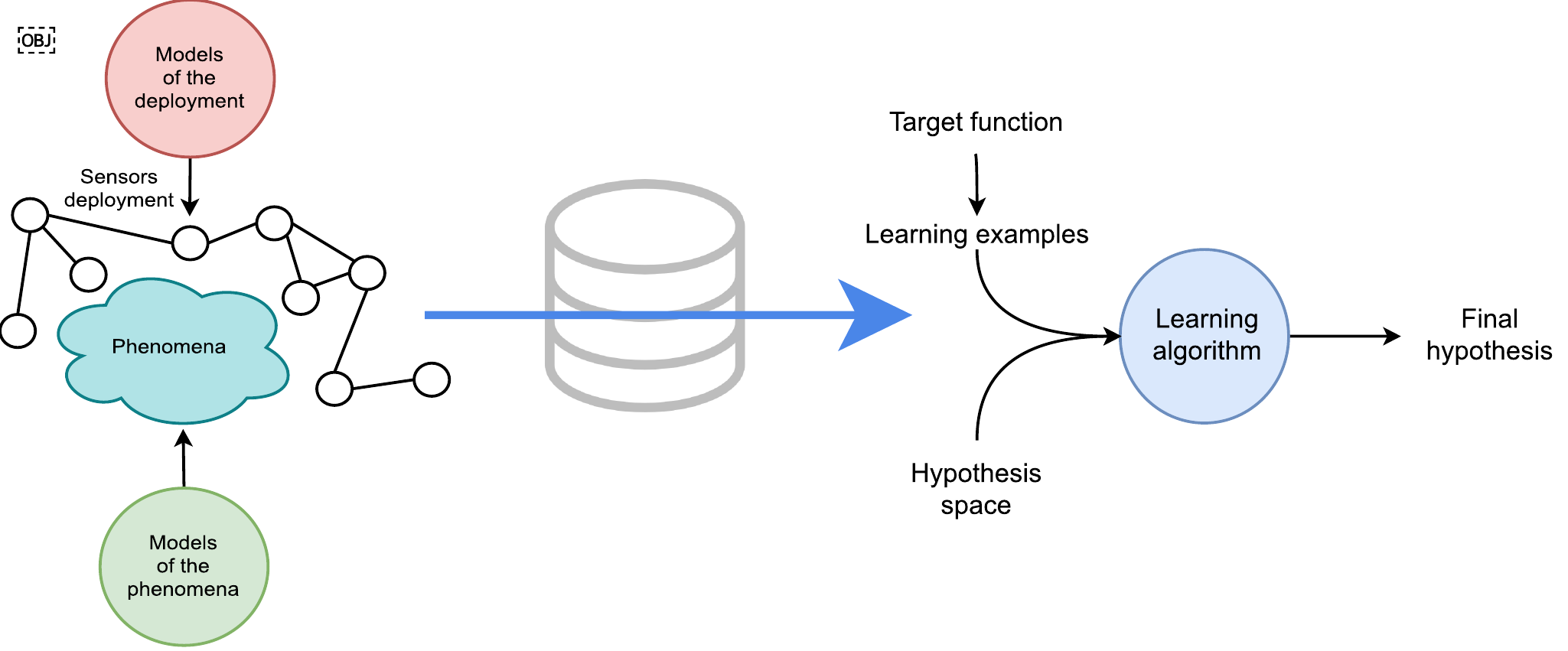
    }
    \caption{We are no longer required to fix the inductive biases. The models of both the sensors deployments and the monitored phenomena serve to guide the learning process by providing the adequate inductive biases dynamically.}
    \label{fig:domain-models-based-learning-setting}
\end{figure}

These surrogate models have larger capacity~\footnote{Capacity refer to the generalization ability (complexity, representativeness, richness, flexibility) of a given model in the sense of Vapnik's definition~\cite{vapnik1995nature}.} and involve slower extraction of information.
Framed in the multi-level structuring of meta-learning, these (surrogate) models are used to guide smaller models which, in the contrary, have generally smaller capacity and can be trained rapidly.
This can be related to the way meta-optimizers, which have been reviewed in the literature, predict the weights of other models that are involved in solving a task.
Conceptually, the idea behind our approach is to remove the barrier that imposes us to fix the inductive biases beforehand (and subsequently the hypothesis space to explore) and rather leverage surrogate models that guide the selection of inductive biases. Figure~\ref{fig:domain-models-based-learning-setting} illustrates this idea.
Models of both the deployments as well as those of the monitored phenomena are highlighted.

Beyond the meta-learning literature, this approach joins the famous Vapnik's privileged information and Hinton's distillation frameworks and can be related, in some extent, to the developments that are witnessed in context-aware models. These are detailed in the following.

{\bf Vapnik's privileged information framework.}
Authors in~\cite{vapnik2015learning} make an analogy with the fact that humans learn much faster than machines and illustrate this with the Japanese proverb \textit{"better than a thousand days of diligent study is one day with a great teacher"}.
Motivated by this insight, the authors incorporate an “intelligent teacher” into machine learning. Their solution is to consider training data formed by a collection of triplets
$\{(x_1, x^{*}_1, y_1), \dots ,(x_n, x^{*}_n, y_n)\} \sim P_n(x, x^{*}, y)$.
Here, each $(x_i, y_i)$ is a feature-label pair, and the novel element $x^{*}_i$ is additional information about the example $(x_i, y_i)$ provided by an intelligent teacher, such as to support the learning process.
Unfortunately, the learning machine will not have access to the teacher explanations $x^{*}_i$ at test time. Thus, the framework of learning using privileged information~\cite{vapnik2015learning} studies how to leverage these explanations $x^{*}_i$ at training time, to build a classifier for test time that outperforms those built on the regular features $x_i$ alone.

{\bf Hinton's distillation framework.}
Authors in~\cite{hinton2015distilling} for their part introduced the distillation framework where knowledge, in the form of class-probability predictions, is distilled from high-capacity models into low-capacity models.
The proposed framework was motivated by 
%and $\Omega : \mathbb{R} \xrightarrow[]{} \mathbb{R}$ is an increasing function which serves as a regularizer.
%When learning from real world data such as high-resolution images, ft is often an ensemble of large deep convolutional neural networks (LeCun et al., 1998a).
the computational cost of predicting new examples at test time using large models, e.g., deep convolutional neural networks in the case of high-resolution images-based learning, is often prohibitive for production systems.
Rather than training the low-capacity, production-"convenient" models using the raw (hard) labels, class-probability predictions (soft labels) generated by the high-capacity models are used instead.
%For this reason, Hinton et al. (2015) propose to distill the learned representation $ft \in Ft$ into
%are the soft predictions from ft about the training data, and Fs is a function class simpler than Ft.
%
A \textit{temperature} parameter controls how much do we want to soften or smooth the class probability predictions, and the \textit{imitation} parameter balances the importance between imitating the soft predictions and predicting the true hard labels. Higher temperatures lead to softer class-probability predictions.
%In turn, softer class-probability predictions reveal label dependencies which would be otherwise hidden as extremely large or small numbers. After distillation, we can use the simpler $f_s \in F_s$ for faster prediction at test time.
This framework represents another approach for acquiring different kinds of knowledge and structuring it in a multi-level fashion.

{\bf Context-aware models.}
Also, many links can be drawn with context-aware activity recognition models which, in some sense, leverage a kind of privileged information in order to improve both the recognition quality of recognition models and also their data requirements.
For example, various approaches were proposed to leverage domain knowledge such as a user’s tasks (e.g.,  spontaneous activity,  engaged tasks) or a user’s social environment (e.g., co-location of others, group dynamics), etc. in order to define the context~\cite{huo2020uncertainty}.
Many different applications including  activity  recognition~\cite{lan2010beyond,helaoui2012towards}, adaptive activity classification~\cite{xu2016personalized},  as well as healthcare-related applications~\cite{andreu2015wearable,spiegel2014validation} were investigated from the lances of context incorporation.
Reduction of data rates using contextual information were also investigated in~\cite{mohomed2008context} where authors proposed to exploit user’s context (both physiological and activity) to adapt the stream processing logic on the client device. This is motivated by the observation that both the medical events being observed and the ‘expected’ values for various medical parameters (e.g., heart rate) are often a function of an individual’s activity (e.g., walking vs. running) and his medical context (e.g., prescribed medication).
\subsection{A Surrogate model for activity recognition}
Construction of surrogate models can take different forms and can be instantiated using different approaches.
We take the two-level approach reviewed above and define a meta-optimizer or surrogate model based on neural architecture search (hyperparameter optimization).
Constructing the surrogate model allows us to extract meta-data and incorporate it into simpler models via sample selection.
In the following, we detail the proposed approach around three questions: \textit{which knowledge to encode?}; \textit{how knowledge is encoded?}; and \textit{how encoded knowledge is exploited?}
\subsubsection{Which knowledge to encode?}\label{sec:dynamic-inductive-bias-selection:which-knowledge-to-encode}
We chose to encode the dynamics of body movements (the phenomenon) along with the importance of the sensing nodes (the deployment) using a surrogate model (hyperparameters space)
Incorporation of prior knowledge derived, directly or indirectly, from 3D body skeleton-based representations holds an important place in the literature around activity recognition.
%__
A long line of research, e.g.~\cite{vatavu2012multi,kovalenko2014real,papadopoulos2014real,parisi2017emergence,dhiman2019skeleton}, proposed to incorporate the 3D body skeleton-based representation into activity recognition models.
%__
Specifically, authors in~\cite{parisi2017emergence}
estimate centroids for upper, middle, and lower body and use slopes of the segments delimited by these centroids in order to represent the posture in terms of the overall orientation of the upper and lower body.
%__
In~\cite{papadopoulos2014real}, authors introduced a representation based on the calculation of spherical angles between selected joints and the respective angular velocities. They used their system for real-time tracking of human activities.
%__
Other works encode prior domain knowledge using ontology-based representations~\cite{ousmer2019ontology,rodriguez2013understanding} which are then used to constrain training of activity recognition models.
%__
In~\cite{vishwakarma2019visual}, authors leverage key pose images and textural traits at various orientations and scales. These are extracted using Gabor wavelet while shape traits are computed through a multilevel approach called spatial edge distribution of gradients.
%__
In~\cite{dhiman2019robust}, authors use depth videos to capture the structural appearance of human poses and their temporal motion contents. These are used to recognize abnormal human actions such as falling.
%__
In~\cite{dhiman2019skeleton}, authors use part-wise skeleton-based motion dynamics to highlight local features of the skeleton. They partition the complete skeleton in five parts: head to spine, left leg,  right leg, left hand, right hand, and use both spatial and temporal information of action using 3D skeleton-based representations of the actions to recognize human actions. Authors, in this study, showed superior performance when using local dynamics encoded by these parts over global dynamics encoded by the full skeleton.
\subsubsection{How knowledge is concretely encoded?}\label{sec:dynamic-inductive-bias-selection:how-knowledge-is-concretely-encoded}
%\subsection{Architecture Space as Proxy for the DGP}
We use the space defined by multimodal analysis architectures as a proxy for the dynamics of the body movements. The exploration of this architecture space serves, then, to derive the knowledge about these dynamics.
In this work, we focus on two different notions that encode these dynamics: \textit{importance} of a single data source and degree of \textit{interaction} among a set of data sources.
Given a data source $s_i$ that is attached to a given body part and an activity $y$, the \textbf{importance} of $s_i$ with regards to activity $y$, denoted $\mu_i^y  \in [0, 1)$, is defined as a quantity that represents the relative involvement of that body part in the dynamics of the gestures pertaining to that activity.
An \textbf{interaction}
involves two or more data sources and is defined as their degree of dependence regarding the relative involvement of the body parts, they are attached to, in the dynamics of the gestures.
The greater the degree, the more interacting the data sources.

\begin{figure}[h!]
    \sffamily
    \centering
    \hspace*{-8mm}
    \fontsize{4}{1}\selectfont
    \def\svgwidth{0.21\columnwidth}
    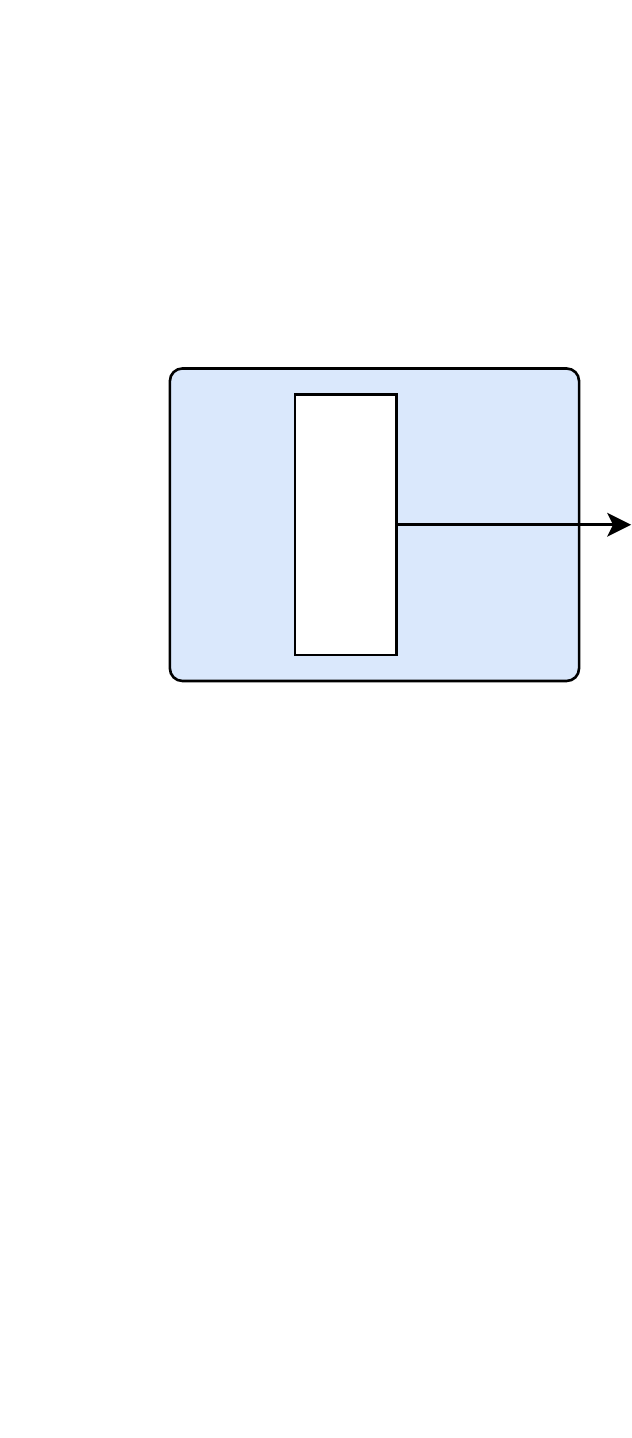
    \hspace*{5mm}
    \fontsize{5}{4}\selectfont
    \def\svgwidth{0.26\columnwidth}
    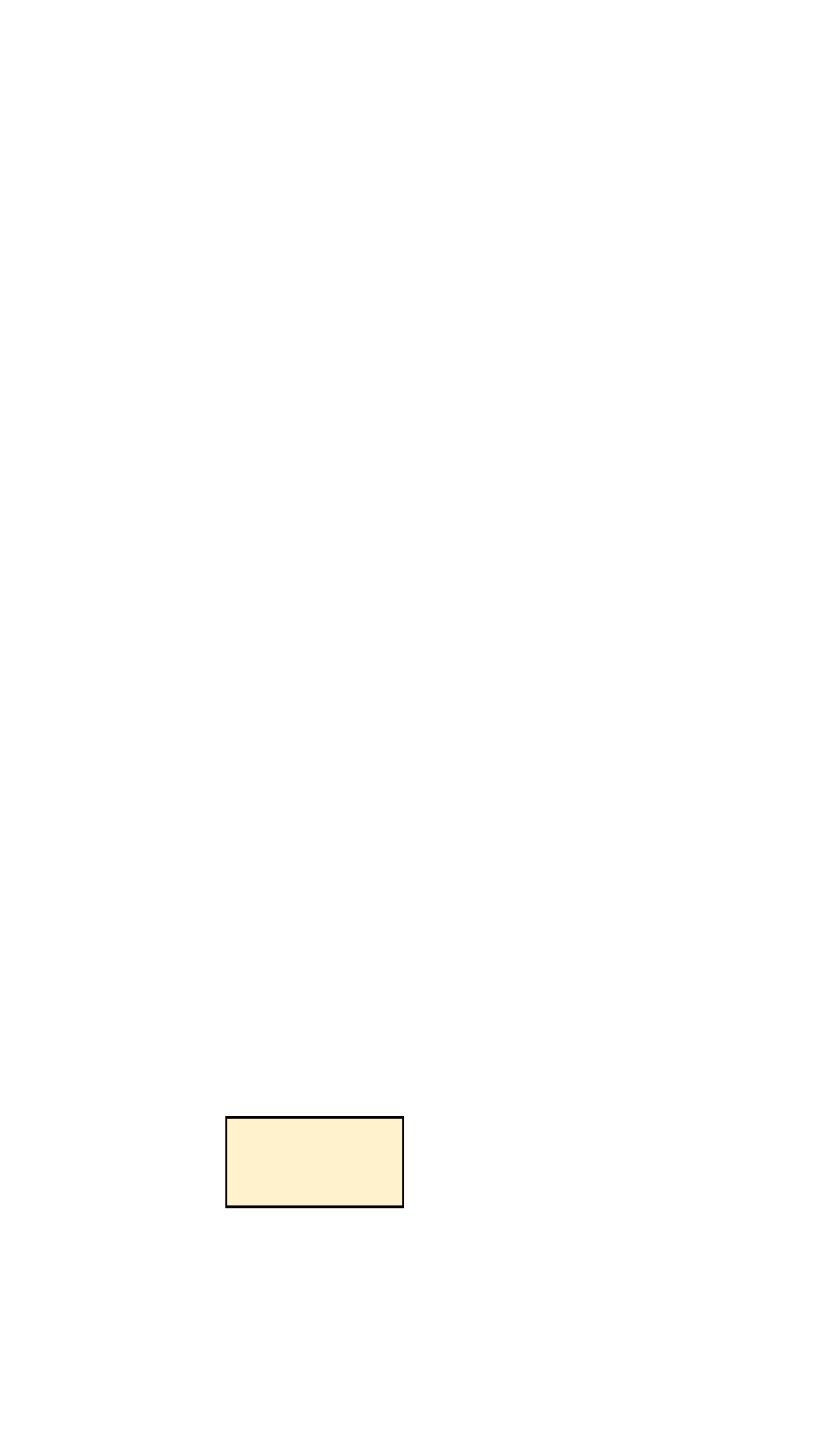
    \caption{
    \small
    (Left) feature extraction and multimodal fusion components defined in~\protect\cite{atrey2010multimodal}.
    Feature extraction (FE), feature fusion (FF), decision fusion (DF), and analysis unit (AU).
    These building blocks can be combined in order to form feature-level, decision-level, and hybrid multimodal analysis.
    Additionally, the hyperparameters $h_i$ controlling the effects of each individual input are depicted.
    (Right) An illustration of an architecture where each node corresponds to a component.
    An edge from component $C_u$ to component $C_v$ denotes that $C_v$ receives the output of $C_u$ as input.
    }
    \label{fig:kr-nas-based-framework}
\end{figure}

An architecture is defined as a set of architectural components
responsible for extracting valuable insights, in the form of features, from the observations and efficiently fusing different data sources carrying different modalities and various spatial perspectives.
We distinguish four types of architectural components: {\it feature extraction} (FE), {\it feature fusion} (FF), {\it decision fusion} (DF), and {\it analysis unit} (AU) as defined in~\cite{atrey2010multimodal}.
These are illustrated in Fig.~\ref{fig:kr-nas-based-framework} (left).
An architectural component takes as inputs either raw data, features, or decisions and outputs either a feature or a decision.
The way a given component processes each individual input is controlled by a hyperparameter.
It is convenient to represent an architecture as a directed acyclic graph where the architectural components are connected together using valued edges.
We associate a value (hyperparameter) $h_u^v$ with every edge in the directed graph that connects two components $C_u$ and $C_v$.
These values control how architectural components process each individual input and by the same occasion their influence on the overall architecture performance.
We refer to the set of all hyperparameters of a given architecture by $\mathcal{H}$.

We focus, particularly, on the insights that stem from tuning and adapting these architectures, through their hyperparameters and specifically those controlling the influence of the data sources.
At each layer of a given architecture, setting the right combination of hyperparameters is critical. In particular, choosing the right instantiation for the features learning and sensor fusion components can lead to an architecture capable of building, from the various data sources, an original set of features which is suitable for recognizing a given activity.
We take into account the following assumption:
let $\mathcal{H}_s \subsetneq \mathcal{H}$ be the set of hyperparameters controlling the impact of a given data source $s$.
The global impact of $\mathcal{H}_s$ on the recognition performances represents the impact of the data source $s$.

The problem of modeling the DGP becomes, then, an exploration of the architecture (hyperparameter) space.
This exploration is determined by three aspects: (1) a search space which defines the architectural components and the type of branching that is allowed for the architectures (e.g. convolutional layers); (2) a search strategy which decides how the exploration of the space should be carried (e.g. Bayesian optimization of the hyperparameters); and (3) a performance estimation strategy (e.g. sequence classification problem)~\cite{elsken2019neural}.

In the case of convolutional layers, for example, architectures can be constructed by stacking a series of Conv1d/ReLU/MaxPool blocks followed by Fully-Connected/ReLU layers.
Denote by $\nu_k$ the validation loss of a particular instantiation $k$ of the set of hyperparameters.
The exploration strategy tries to find an architecture $k^{*}$ that minimizes the validation loss $\nu_{k^*}(w^{*})$.
The weights $w$ associated with the architecture are obtained by optimizing the weights of the components using, for example, a gradient descent algorithm over a predefined class of functions.

Given an exploration budget $B$, the exploration strategy yields a series of validation losses $\nu_1, \dots, \nu_B$ including partial validation losses pertaining to individual activities.
The task of modeling the DGP, therefore, reduces to find a link between these validation losses and the impact of each individual data source.
Figure~\ref{fig:kr-shl-deployment-topology-running} illustrates the main steps used to derive the data generation process.
Setting described in~\cite{hamidi2020data}.
Derivation of the data generation process is framed as an exploration of an architecture space which constitutes a surrogate model.

\begin{figure}[h!]
\centering
\sffamily
\def\svgwidth{1.2\columnwidth}
\resizebox{90mm}{!}{
   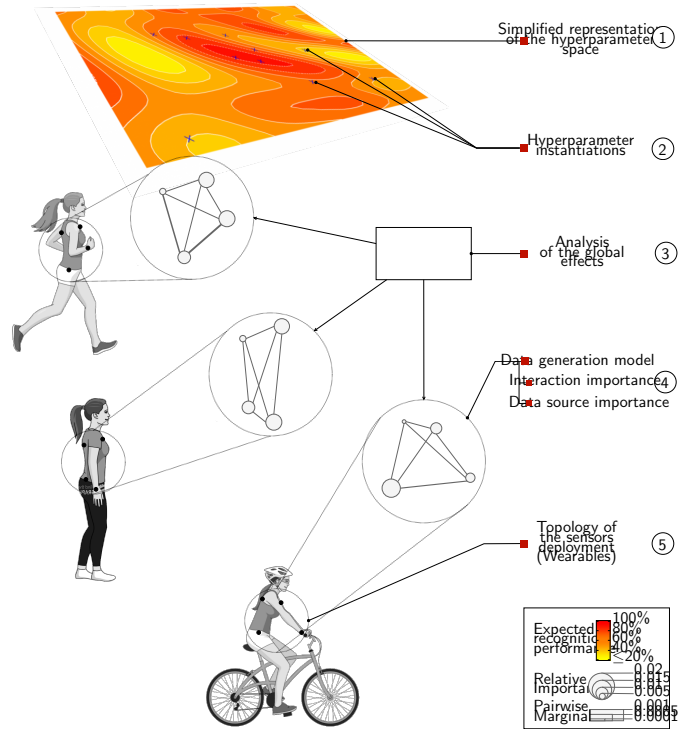
}
\caption{
    \textbf{Schematic description of used experiments.}
    Deriving the data generation model capturing the body movements dynamics of human activity is framed as an exploration of the neural architecture space. This approach allows to explore the architectures space to find more influential hyperparameters for each human activity.
    This examples shows the importance of each position and the level of interactions between modalities for the recognition of running, still, and bicycling.
}
\label{fig:sh-infrastructure-promotional}
\end{figure}
\subsubsection{How encoded knowledge is exploited?}
The constructed model is able to cope with evolution of deployments. It leverages the dynamics of the body movements and the assumption stating that each individual activity is characterized by a different set of gestures which in turn involve specific body parts equipped with data sources.  
This is why we use bottom-up approach to train simpler models via  subsets data sources selection  that are highly confident and informative regarding these dynamics, to create a curated training set for model training.
Indeed, as we have access to the best performing architectures, one can deploy it directly without going through models of low capacity. Here, we are rather interested in the overall behavior and form of the architecture space expressed by the configurations that have been explored. This behavior is what we incorporate into models that are more restricted in terms of capacity, and therefore easy to train and adapt.
Practically speaking, we select highly informative data sources to form training sets.
During the training phase, activity recognition models are encouraged to concentrate on the provided subsets of data sources to learn the corresponding human activities.
\section{Case Study}\label{sec:case-study}
Here we describe the used SHL dataset in our empirical evaluation \cite{gjoreski2018university}\footnote{The preview of the SHL data set can be downloaded from: \url{http://www.shl-dataset.org/download/}.}. It is a highly versatile and precisely annotated dataset dedicated to mobility-related human activity recognition (750 hours of labeled locomotion data). It provides, simultaneously, multimodal and multilocation locomotion data recorded in real-life settings more than most of used ones like ~\cite{zheng2010geolife,zhang2012usc,yu2014big,carpineti2018custom}. 
There are in total 16 modalities including accelerometer, gyroscope, cellular networks, WiFi networks, audio, etc. making it suitable for a wide range of applications.
Among the 16 modalities of the original dataset, we select the body-motion modalities to be included in our experiments, namely: accelerometer, gyroscope, magnetometer, linear acceleration, orientation, gravity, and in addition, ambient pressure.

\begin{figure}[h!]
    \centering
    \sffamily
    \def\svgwidth{0.7\columnwidth}
    \resizebox{40mm}{!}{
        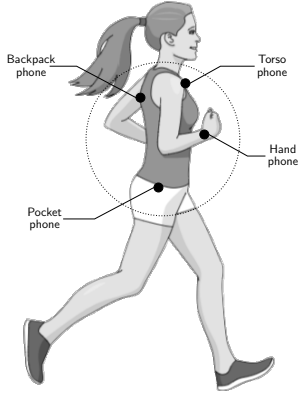
    }
    \caption{Topology of the on-body sensors deployment.}
    \label{fig:kr-shl-deployment-topology-running}
\end{figure}

Data collection was performed by each participant using four smartphones simultaneously placed in different body locations:
{\it Hand}, {\it Torso}, {\it Hips}, and {\it Bag}.
These four positions define the topology that allows us to model and leverage the dynamics of body movements for activity recognition models.
Figure~\ref{fig:kr-shl-deployment-topology-running} shows the on-body sensors deployment used during data collection.

\section{Evaluations}\label{sec:evaluations}
We conduct in this section an empirical evaluation of the dynamic inductive bias selection via two axes (i) adaptive segmentation, and (ii) adaptive sampling (Sect.~\ref{sec:evaluations}).
We particularly provide An illustration of surrogate models used to guide the segmentation and sampling.
\subsection{Basic activity recognition chain}
\label{sec:evaluations:setup}
In this first set of evaluations, we consider a basic activity recognition chain.
\paragraph{Inputs/preprocessing/segmentation} Sensor signals were sampled at 100 Hz and the frames, for training and testing dataset were generated by segmenting the whole data with a non-overlap sliding window of 1-minute length.
The subset that we use consists of 5 hours representing 16310 sequences for the training part and
5698 sequences of a duration of one minute each are reserved to test our models.
\paragraph{Feature learning and classification}
The main idea is to use architectures based on neural networks in order to overcome aforementioned limits of human expertise and to come-up with genuine features extraction and sensor fusion schemes.
Neural networks hold important properties that are advantageous to multimodal recognition tasks. They are able to construct, or learn, hierarchies of abstract features and relate efficiently modalities between them.
In our work, to replicate such capabilities within our architectures, we use convolutional neural networks.
Beyond the frequent application of convolutional neural networks for the recognition of human activities, which show, by the way, good performances, e.g.~\cite{ha2016convolutional,ordonez2016deep,radu2018multimodal,bevilacqua2018human}, these kinds of networks are being adopted, primarily, for their ability to efficiently aggregate heterogeneous data from different sources.
In~\cite{ha2016convolutional} for example, authors proposed various convolutional architectures featuring an explicit mechanism for partial and full weight sharing, by placing separate convolution kernels on each modality and in the upper layers responsible for aggregating features maps.
In this work, we construct neural architectures by stacking Conv/ReLU/MaxPool blocks.
These blocks are followed by a Fully Connected/ReLU layers.
In order to allow for the emergence of cross-modal relationships at both low and high-levels of abstractions,
we define three {\it convolutional modes} of the input sequences with each set of filters:
\begin{itemize}
    \item whole modalities grouped and convolved, referred to as {\it grouped modalities}. %See Figure.~\ref{fig:modalities-grouped};
    \item each modality convolved apart, which is designated by {\it split modalities}. %See Figure.~\ref{fig:modalities-appart};
    \item each channel convolved apart, referred to as {\it split channels}. %See Figure.~\ref{fig:channels-appart-arch}.
\end{itemize}
These various convolutional modes can be considered as different levels of sensor fusion.
From this perspective, {\it split channels} would correspond to a late fusion and, on the contrary, {\it grouped modalities} would correspond to an early fusion scheme.

In order to train a given architecture, we frame recognizing human activities as a sequence classification problem, where the goal is to learn a function $\mathcal{F}: X \xrightarrow{} Y$ mapping inputs to outputs.
As in the traditional classification setting, performance of the neural architecture is quantified with a loss function $\ell : X \times Y \xrightarrow{} \mathds{R}$, and a mapping is found via
\begin{align}
    f^{*} = \operatorname*{argmin}_{f \in \mathcal{F}} \frac{1}{N}\sum_{j=1}^{N} \ell(f(\mathbf{x}_j), y_j)
\end{align}
which can be optimized using a gradient descent algorithm over a pre-defined class of functions $\mathcal{F}$.
In our case, $\mathcal{F}$ will be convolutional networks parametrized by their weights and the loss function will be $\ell(f(\mathbf{x}_i), y_i) = \mathds{1}\{f(\mathbf{x}_i) \neq y_i\}$.
For a fixed architecture, i.e. a particular instantiation of the hyperparameters, the optimization process will tune the weights of the network and, by the same occasion, the subsequent uni-modal and multi-modal features that are extracted from the input signals.
\subsubsection{Results}
\begin{figure*}[!h]
\centering
\sffamily
\subfloat[]{
    \def\svgwidth{0.8\columnwidth}
    \resizebox{70mm}{!}{
       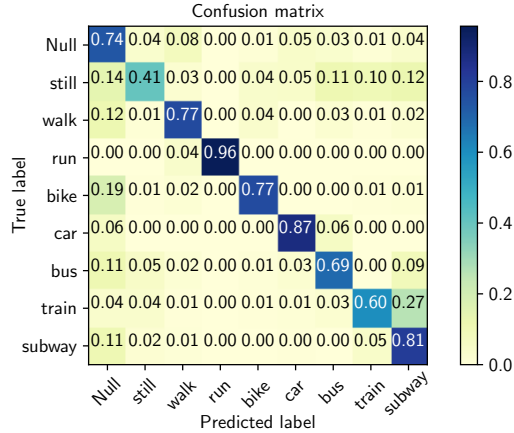
    }
    \label{fig:confusion_matrix-5-5-3-5-AllPositionsFused-10-folds-modelIter35}
}%
\caption{
    %(a)
    Confusion matrix
    %and (b) Receiver operating characteristic curve
    of the best model trained on all body locations fused together yielding 70.86 \% recognition performance measured by the f1-score.
}
\label{fig:all-positions-fused}
\end{figure*}

Figure~\ref{fig:all-positions-fused} shows the recognition performance of the best model trained on all body locations fused together yielding 70.86 \% recognition performance measured by the f1-score.
\subsubsection{Evaluation and model selection}\label{sec:evaluation-and-model-selection}
The $k$-fold cross-validation is widely used for assessing performances of a prediction model. It performs, first, a random partitioning of the dataset into $k$ distinct folds. Then, at each iteration, a different subset consisting of $k-1$ folds is used to train the model while the remaining fold is used for the purpose of validation.

\paragraph{Usual partitioning and \textit{neighborhood bias}}
The random partitioning used in the case of segmented time-series introduces a neighborhood bias~\citep[]{hammerla2015let}. This bias consists in the high probability that adjacent and overlapping frames, that are typically obtained during segmentation
%a process similar to the one we have developed in Section~\ref{sec:audio-signal-preprocessing}
and that share a great deal of characteristics fall into training and validation folds in the same time. This leads to an overestimation of the validation results and goes often disregarded in the literature. 
\begin{figure*}[!ht]
\hspace*{-4.6em}
\centering
\sffamily
%\scriptsize
\def\svgwidth{2.0\columnwidth}
\resizebox{150mm}{!}{
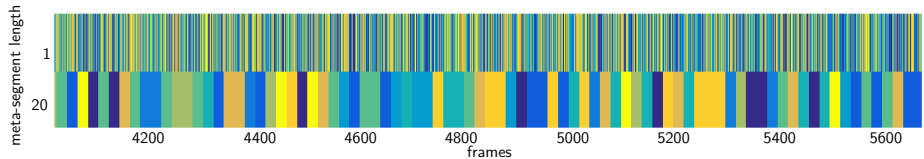
%\label{fig:associated-fold-for-each-frame-25vs1}
}
\caption{Partitioning of a portion of the final labeled dataset's frames over 10 folds using meta-segmented partitioning algorithm proposed in~\cite{hammerla2015let}. A segment length of one corresponds to the partitioning produced by the regular cross-validation procedure. Shown frames range from 4030 to 5670 are ordered by their time indexing across recordings. Each color corresponds to a different fold.}
\label{fig:associated-fold-for-each-frame}
\end{figure*}

\paragraph{Meta-segmented partitioning}
which is proposed in~\citep[Section 5]{hammerla2015let} tries to circumvent this bias by, first, grouping adjacent frames into meta-segments of a given size. These meta-segments are then distributed on each fold. Figure~\ref{fig:associated-fold-for-each-frame} shows the partitioning of a subset of the generated frames over 10 folds with a meta-segment length of 1, which corresponds to the usual partitioning procedure, and a meta-segment length of 20.\\

\begin{figure}[h!]
\sffamily
\centering
\subfloat[]{
    \def\svgwidth{1.5\columnwidth}
    \resizebox{80mm}{!}{
        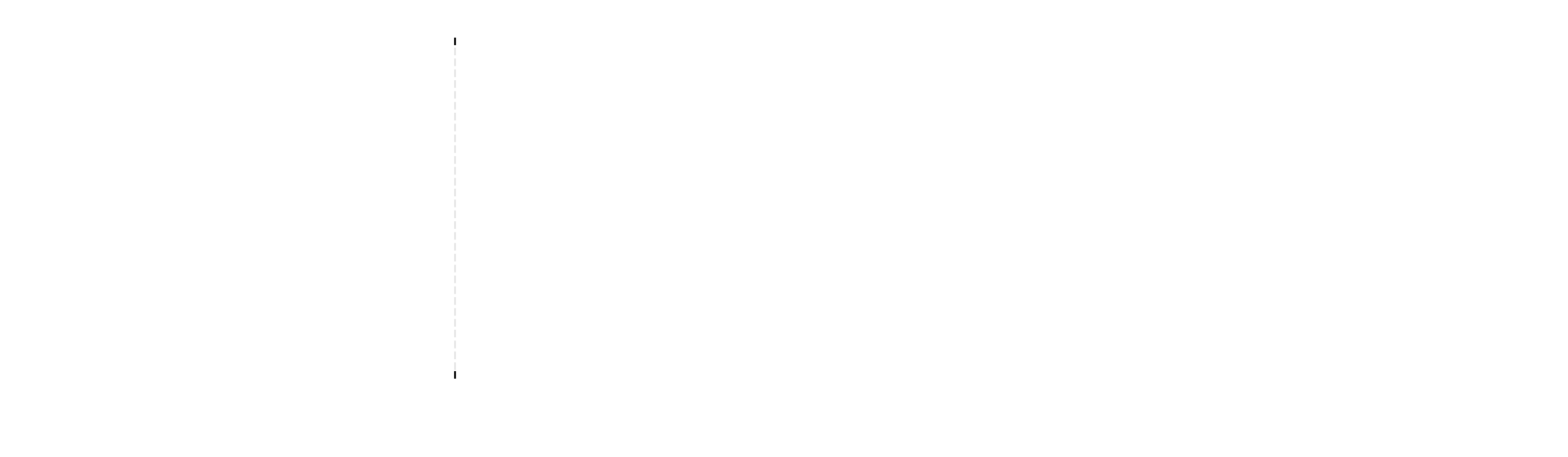
    }
    \label{fig:pitfalls-relative-bias-5-5-3-2}
}%

\subfloat[]{
    \def\svgwidth{1.5\columnwidth}
    \resizebox{80mm}{!}{
        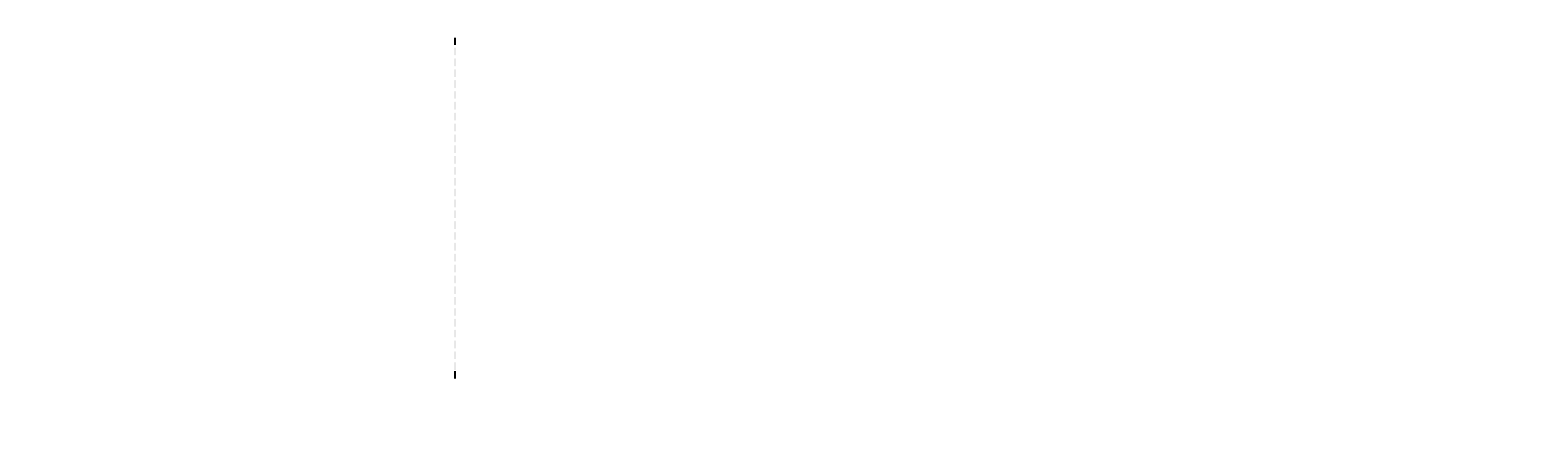
    }
    \label{fig:pitfalls-relative-bias-5-5-3-4}
}%

\subfloat[]{
    \def\svgwidth{1.5\columnwidth}
    \resizebox{80mm}{!}{
        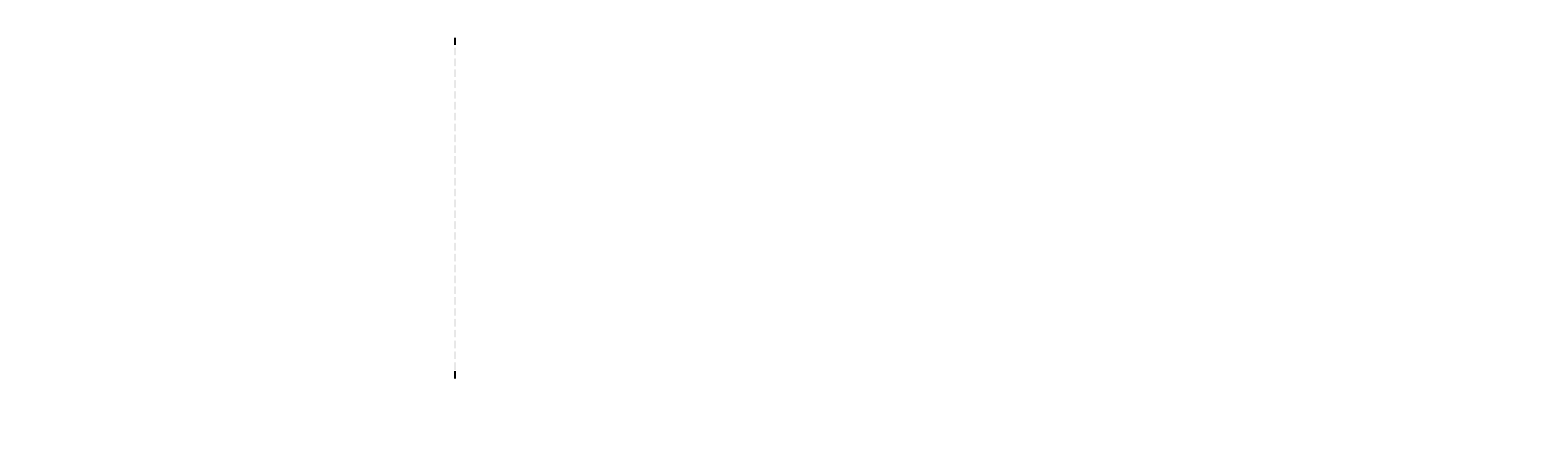
    }
    \label{fig:pitfalls-relative-bias-5-5-3-5}
}%
\caption{
    Relative bias induced by the way the f1-score is computed and its interplay with validation strategies used during experiments.
    10-folds cross-validation with the null class being (a) discarded and (b) included. (c) 5-folds meta-segmented cross-validation. From~\cite{hamidi2020domain}.
}
\label{fig:fmeasure-pitfalls}
\end{figure}
%

%\subsection{Adaptive segmentation}
\subsection{Surrogate model based on Bayesian optimization}
In this set of experiments, we evaluate an adaptive segmentation process and its impact on the final recognition performances.
The experimental setting is presented in more details in~\cite{osmani2018hybrid}.

%_ici on présente l'instantiation du surrogate model présenté dans la section 5.6 basée sur une procédure d'optimisation Bayesienne

%_présenter les composants architecturaux utilisés pour cette partie expérimentale

%_présenter la fonction d'acquisition qui contrôle l'exploration de l'espace des architectures

\subsubsection{Setup}
Here we detail precisely the different building blocks presented in Sect.~\ref{sec:evaluations:setup}.
Note that in the following, hyperparameters accompanied with a mathematical notation are subject to the BO procedure.
We use up to 3 convolutional layers in the features learning stage, each followed by a unit of max-pooling whose parameters, i.e. window size, is set to 2.
Each convolutional layer has its own set of filters that have sizes $ks_{i}$, $i \in \{1, 2, 3\}$ and are different from one layer to another.
The number of filters $n_f$ meanwhile remains the same for each layer of a given architecture.
%
%\textcolor{red}{
We use two types of activation functions, namely the rectified linear unit (ReLU) and the hyperbolic tangent (Tanh) activation functions.
%}
%
Concerning the recurrent layers, we use 2 layers of LSTM units composed of $n_{hu1}$ and $n_{hu2}$ hidden units for layers 1 and 2 respectively.
Moreover, we initialize the bias of the forget gate to 1 according to~\cite{jozefowicz2015empirical} who recommend setting the bias of this gate to relatively wide values such as 1 or 2 which allows the gradient to flow easily.

With regard to the input signals
%, according to the assumption we made in Sect.~\ref{sec:method},
we decided to discard any additional segmentation process that could potentially introduce bias into the learning process.
Inputs are, then, taken as they are without any additional segmentation process.
The entries of our different architectures are therefore of the order of 1 minute, i.e. 6000 samples, given that the sampling rate is 100 Hz.
\begin{table}[h!]
    %\sidecaption
    \centering
\caption{Summary of the different hyper-parameters assessed during Bayesian optimization procedure along with their respective bounds. From~\cite{osmani2019bayesian}.}
    \begin{tabular}{lccc}
    \toprule
    Hyper-param. (sym)                      & low & high & prior \\
    \midrule
    Learning rate ($lr$)                 & 0.001 & 0.1 & log \\
    \midrule
    Kernel size 1$^{st}$ ($ks_{1}$)         & 9 & 15 & - \\
    Kernel size 2$^{nd}$ ($ks_{2}$)         & 9 & 15 & - \\
    Kernel size 3$^{rd}$ ($ks_{3}$)         & 9 & 12 & - \\
    Number of filters ($n_f$)               & 16 & 28 & - \\
    Stride ($s$)                            & 0.5 & 0.6 & log \\
    \midrule
    Dropout probability ($p_{d}$)           & 0.1 & 0.5 & log \\
    Number of units dense layer ($n_{u}$)   & 64 & 2048 & - \\
    \midrule
    Number of hidden units 1 ($n_{hu1}$)    & 64 & 384 & - \\
    Number of hidden units 2 ($n_{hu2}$)    & 64 & 384 & - \\
    \midrule
    Inputs dropout probability ($p_{in}$)   & 0.5 & 1 & log \\
    Outputs dropout probability ($p_{ou}$)  & 0.5 & 1 & log \\
    States dropout probability ($p_{st}$)   & 0.5 & 1 & log \\
    \bottomrule
\end{tabular}
\label{tab:hyper-parameters}
\end{table}

\subsubsection{Analysis of the surrogate model's response surface}\label{sec:evaluations:analysis-of-the-surrogate-models-response-surface}
\begin{figure}
\centering
\subfloat[]{
    \def\svgwidth{0.9\columnwidth}
    \resizebox{55mm}{!}{
        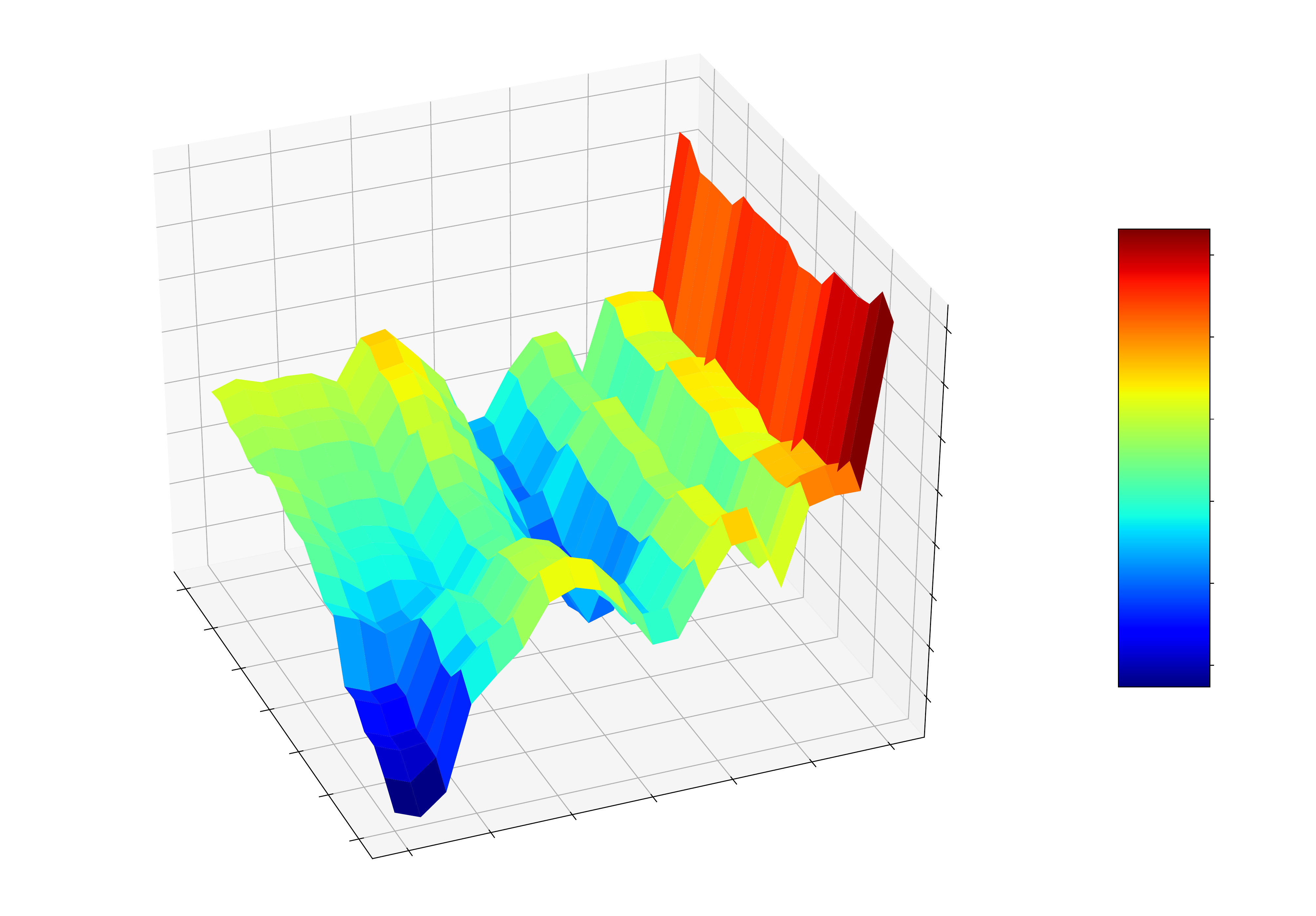
    }
}%
\subfloat[]{
    \def\svgwidth{0.9\columnwidth}
    \resizebox{55mm}{!}{
        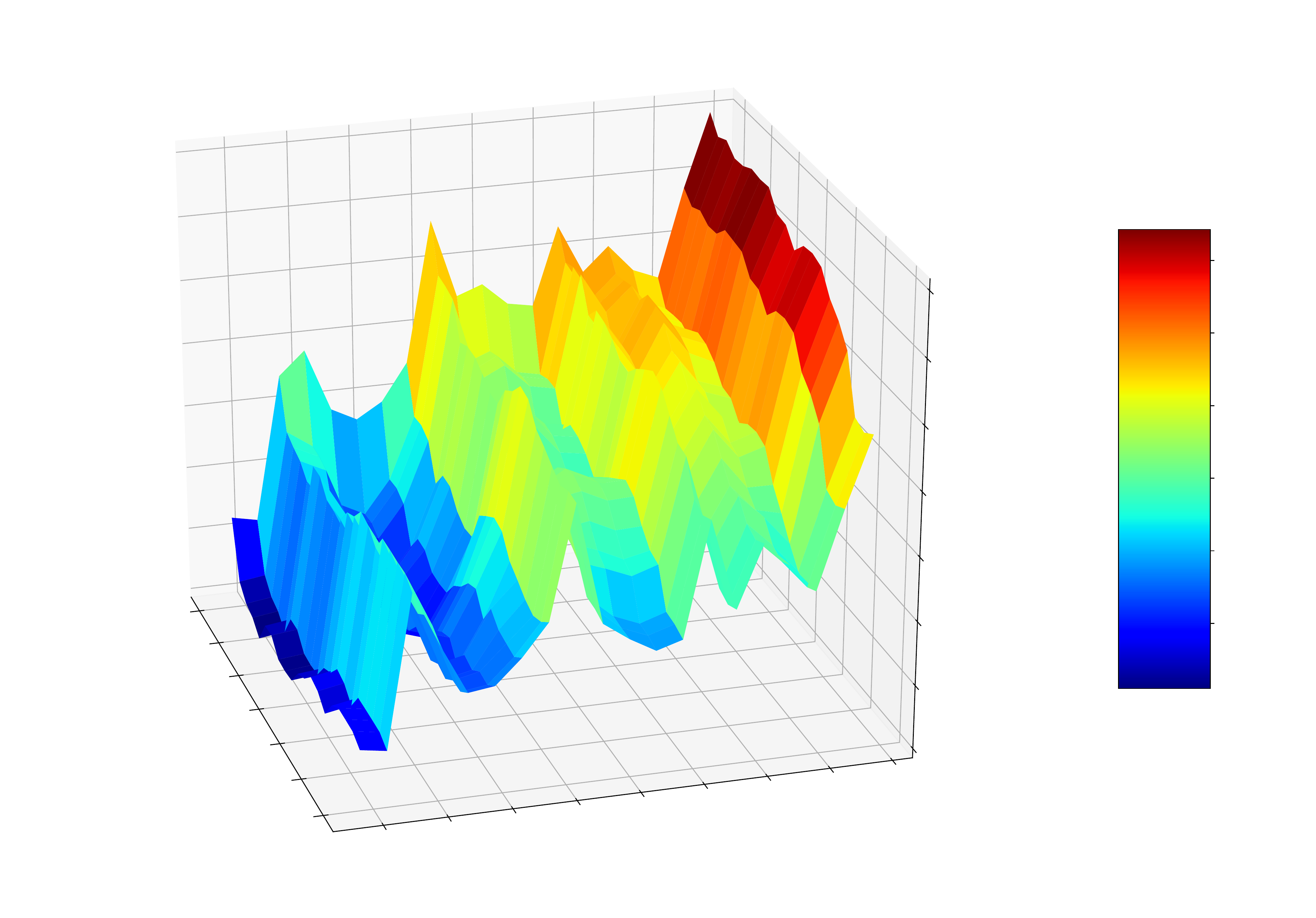
    }
}
\caption{Pairwise marginal plots produced via fANOVA framework~\cite{hoos2014efficient} for convolutional architectures. (a) kernel size 2 and kernel size 3 of convolutional layers 2 and 3 respectively, (b) number of units of the dense layer and kernel size 2 of convolutional layer 2.}
\label{fig:cnn-fanova}
\end{figure}

%Figs.~\ref{fig:convolutional-modes-cum-dist-HYBRID} and~\ref{fig:convolutional-modes-cum-dist-CONVOLUTIONAL} show recognition performances achieved by hybrid and convolutional architectures, respectively.
%These figures compare in particular the influence of the different convolutional modes being experimented in this work.
Table~\ref{tab:hyper-parameters} provides a summary of the most important hyperparameter pairwise marginals obtained on convolutional architectures.
Table~\ref{tab:most-important-pairwise-marginals-CONVOLUTIONAL-HYBRID} provides, for its part, a summary of the hyperparameters' importance obtained through the fANOVA analysis of convolutional and hybrid architectures' recognition performances.
Figure~\ref{fig:cnn-fanova} provide pairwise marginal plots of a set of hyperparameters in the case of convolutional obtained also through the fANOVA framework~\cite{hoos2014efficient}.

In the case of architectures with split modalities, throughout the BO runs we do notice that the number of architectures yielding, for example, a given recognition performance varies to a critical large extent ($\pm 0.6$ standard deviation points).
This underlines just how complex the subspace generated by the split modalities convolutional mode is.
In other words, the generated subspace can be viewed as spanning the entire recognition performances range and encompassing many plateaus of variable sizes scattered all over it, each of which yielding equivalent performing architectures.
This contrasts with the convolutional architectures where we do not get a similar variability.
%(Fig.~\ref{fig:convolutional-modes-cum-dist-CONVOLUTIONAL}).
Rather, we see clearly that, with the exception of all modalities grouped convolutional mode which shows a much more stable behavior, the variability is confined to $\pm 0.2$ standard deviation points.
This suggests that the entire hyper-parameter space induced by convolutional architectures, regardless of the convolutional mode, is characterized by abrupt topology and the absence of sustainable plateaus
which is presumably what the analysis of the pairwise marginal plots shown in Fig.~\ref{fig:cnn-fanova} tends to support to a certain extent.
This being said, we don't know which of the building blocks account for the above observations.
What we know is that the building blocks of our architectures exhibit a lot of complex interactions which influence the recognition performances, as a matter of fact, quite large.
In order to get insights about these interactions, what can be done, is to assess the respective hyper-parameters of these building blocks as well as the low-level interactions between them.

\begin{table}
    \centering
\caption{Summary of the hyper-parameters' importance obtained through the fANOVA analysis of convolutional and hybrid architectures' recognition performances.}
    \begin{tabular}{lcc}
    \toprule
                                            & \multicolumn{2}{c}{Individual importance} \\
    \cline{2-3}
    Hyper-param. (sym)                      & Convolutional                 & Hybrid \\
    \midrule
    Learning rate ($l_{r}$)                 & 0.10423                       & 0.19815 \\
    \midrule
    Kernel size 1$^{st}$ ($ks_{1}$)         & 0.01410                       & 0.00874 \\
    Kernel size 2$^{nd}$ ($ks_{2}$)         & 0.00916                       & 0.023105 \\
    Kernel size 3$^{rd}$ ($ks_{3}$)         & 0.04373                       & 0.01788 \\
    Number of filters ($n_f$)               & 0.02810                       & 0.01845 \\
    Stride ($s$)                            & 0.08092                       & 0.06236 \\
    \midrule
    Dropout probability ($p_{d}$)           & 0.03279                       & - \\
    Number of units dense layer ($n_{u}$)   & 0.16748                       & - \\
    \midrule
    Number of hidden units 1 ($n_{hu1}$)    & -                             & 0.06324 \\
    Number of hidden units 2 ($n_{hu2}$)    & -                             & 0.02478 \\
    \midrule
    Inputs dropout probability ($p_{in}$)   & -                             & 0.04047 \\
    Outputs dropout probability ($p_{ou}$)  & -                             & 0.01056 \\
    States dropout probability ($p_{st}$)   & -                             & 0.01991 \\
    \bottomrule
\end{tabular}
\label{tab:most-important-pairwise-marginals-CONVOLUTIONAL-HYBRID}
\end{table}
The main difference between hybrid and convolutional architectures reside in the type of output layers that accounts for discovering combinations of features which correspond to a given activity pattern.
In particular, we suspect that the LSTM layers may be, either directly or indirectly, responsible for the large variability that characterizes hybrid architectures' recognition performances.
Indeed, the comparison between hybrid and convolutional architectures in terms of the respective hyper-parameters of the features learning stage, i.e. kernel sizes, stride, number of filters,
reveals that these have roughly the same influence on the recognition performances and are by far characterized by relatively insignificant interactions among them
%, as shown in Table~\ref{tab:most-important-pairwise-marginals-CONVOLUTIONAL} and~\ref{tab:most-important-pairwise-marginals-HYBRID}
, which translates, presumably, to a kind of stability characterizing this building block.
In the other hand, dense layers' hyper-parameters exhibit an extremely different impact.
Indeed, the number of units in the dense layer $n_u$ is found to be the most important hyper-parameter in the case of convolutional architectures and accounts for more than 16\% of the recognition performances variability.
In addition, the number of units in the dense layer has the largest low-level interactions
%(see Table~\ref{tab:most-important-pairwise-marginals-CONVOLUTIONAL})
, which confirms the important impact of the dense layer in recognizing activity patterns.
This contrasts with the quantified importance of the number of hidden units of LSTM layers which is 6\% and 2\% for the first and second layer respectively.
We notice instead that the learning rate is taking advantage over the number of hidden units and accounts for more than 19\% of the recognition performances variability in the case of hybrid architectures.
%Although this has to be taken with care and need to be further investigated as it involves an optimization procedure which, we should recall, is characterized by non-deterministic behavior,
%our findings would seem to imply that the dense layers exhibit more representational power compared to LSTM layers,
%the latter being disadvantaged during the BO procedure in favor of the learning rate.
%This observation corroborates the findings of~\cite{greff2017lstm} and~\cite{hammerla2016deep} in both a general learning setting and a HAR-specific one respectively.

\begin{figure}[h!]
    \centering
    \sffamily
\subfloat[]{
    \def\svgwidth{0.8\columnwidth}
    \resizebox{55mm}{!}{
       \includegraphics[width=5cm]{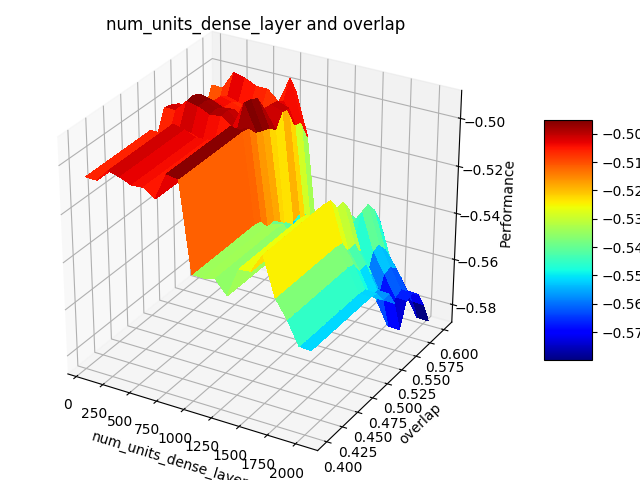}
    }
    \label{fig:num_units_dense_layer_overlap}
}%
\subfloat[]{
    \def\svgwidth{0.8\columnwidth}
    \resizebox{55mm}{!}{
        \includegraphics[width=5cm]{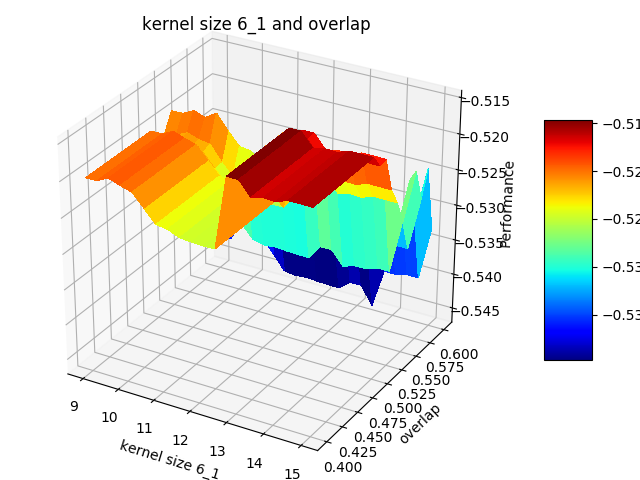}
    }
    \label{fig:kernel_size_6_1_overlap}
}
    \caption{Pairwise marginal plots produced via fANOVA framework~\cite{hoos2014efficient} illustrating the impact of the stride hyperparameter on the recognition performances.}
    \label{fig:fanova-overlap-impact}
\end{figure}

\paragraph{Segmentation via the \textit{stride} hyperparameter}
%Les hyperparametres peuvent être étendu à d'autres aspects de la chaine d'apprentissage et notamment l'étape de segmentation. Par exemple, nous pouvons exploiter pour cela l'hyperparmaetre stride, que nous avons déjà défini dans le tableau résumant les hyperparametres optimisés (voir Table~\ref{tab:hyper-parameters}).
The hyperparameters can be extended to virtually any aspect of the activity recognition chain, including for example the segmentation stage. Indeed, we can leverage for this the stride hyperparameter, which we have already defined in the table summarizing the optimized hyperparameters (see Table~\ref{tab:hyper-parameters}).
%\textbf{Overlap/Stride.}
%Its relation to segmentation \dots
Figure~\ref{fig:stride-conv-architecture} illustrates how each portion of an input signal (here the $y$ dimension of a given modality) is processed by the input layer of the neural architecture.
The portions are determined by the value of the stride which represents in our setting as a hyperparameter to be optimized, just as any other weight or hyperparameter of the neural architecture.
Figure~\ref{fig:fanova-overlap-impact} shows the impact of the overlap/stride hyperparameter on the recognition performances.
%The impact is computed using the fANOVA framework~\cite{hoos2014efficient}.
%
\begin{figure}[h!]
    \centering
    \sffamily
    \def\svgwidth{0.8\columnwidth}
    \resizebox{55mm}{!}{
       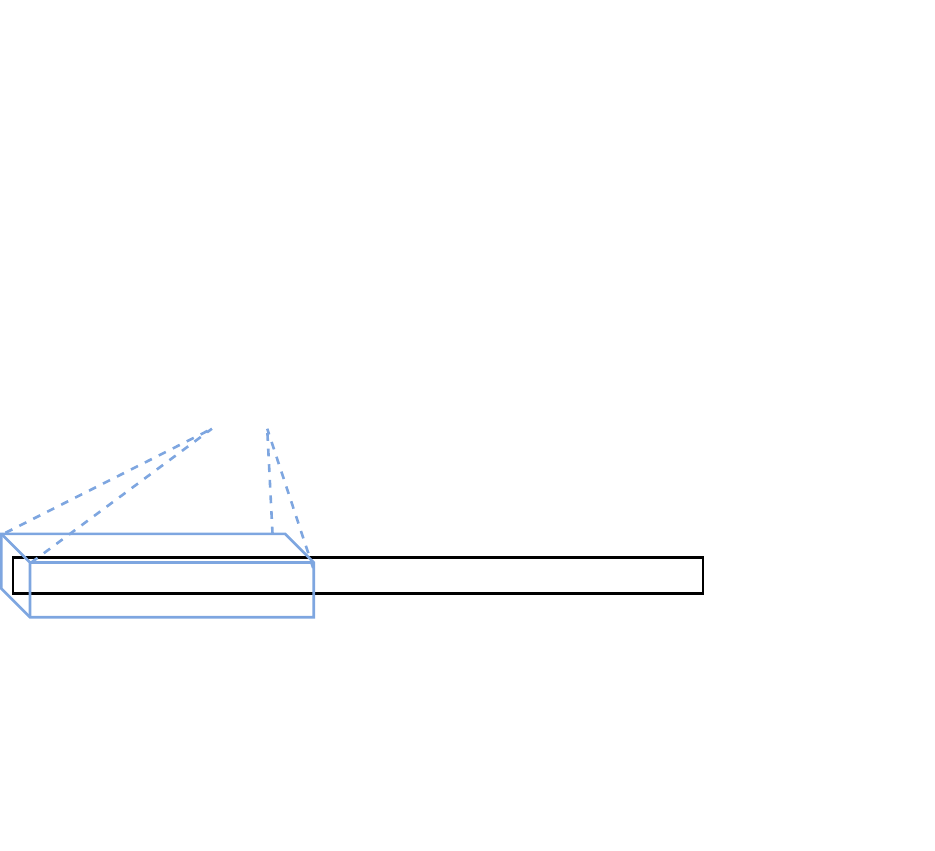
    }
    \label{fig:stride-conv-architecture}
    \caption{Interpretation of the stride hyperparameter as controlling the segmentation step.}
    \label{fig:stride-conv-architecture}
\end{figure}

\subsection{Adaptive sampling}\label{sec:evaluations:adaptive-sampling}
% data generation process
Given the surrogate model, the task now is to incorporate the derived knowledge into low-capacity, data-efficient, and deployment-ready models.
As we mentioned before, human activities are largely determined by the dynamics of the gestures.
Indeed, each activity is characterized by a different set of gestures which in turn involve specific body parts.
%
%In the case of wearable technologies, where these body parts are equipped with data sources, often, focusing on these specific data sources, allows recognizing a given activity precisely.
%Therefore,  our approach attempts  to  select  subsets of data sources that are highly confident and informative with regards to these dynamics, to create a curated training set for model training.
%In this work, we focus on two different notions that encode these dynamics: importance of a single data source and degree of interaction among a set of data sources.
%
Here, we describe the experimental setup used to exhibit and then incorporate the dynamics of body movements.
The experimental setting is presented in more details in~\cite{hamidi2020data}.

\subsubsection{Setup}
Different exploration strategies will lead to different sets of hyperparameter instantiations.
In our experiments, we instantiate our approach with various exploration strategies.
We use the Microsoft-NNI toolkit~\footnote{\url{https://github.com/microsoft/nni}} which provides a comprehensive list of exploration strategies, in particular, those based on hyperparameter tuning, including (1) exhaustive search (random search~\cite{bergstra2012random}, and grid search); (2) heuristic search (naive evolution~\cite{real2017large}, anneal~\cite{bergstra2011algorithms}, and hyperband~\cite{li2017hyperband}); and (3) sequential model-based optimization (Bayesian optimization hyperband~\cite{falkner2018bohb}, tree-structured Parzen Estimator~\cite{bergstra2011algorithms}, and Gaussian process tuner~\cite{bergstra2011algorithms}).

We quantify the influence of data sources using the efficient implementation of fANOVA proposed in~\cite{hoos2014efficient}, which is based on a linear-time algorithm for computing marginal predictions in random forests.
Interaction structure of the data sources is estimated using fanova-graph~\cite{muehlenstaedt2012data}.

In addition, we define two thresholds, $\tau_{imp}\in [0,1)$ and $\tau_{int}\in [0,1)$, above which a given set of data sources $S\subset\mathcal{S}$ can be selected and included into the sample.

{\bf Datasets.}
We use the SHL dataset primarily to derive the data generation model.
The derived model is then incorporated into the SHL dataset itself and three other datasets including
(1) {\it USC-HAD}~\cite{zhang2012usc} containing body-motion modalities of 12 daily activities collected from 14 subjects (7 male,7 female) using MotionNode, a 6-DOF inertial measurement unit, that integrates a 3-axis accelerometer,  3-axis gyroscope,  and a 3-axis magnetometer;
(2) {\it HTC-TMD}~\cite{yu2014big} containing accelerometer, gyroscope, and magnetometer data all sampled at 30Hz from smartphone built-in sensors in the context of energy footprint reduction;
and (3) {\it US-TMD}~\cite{carpineti2018custom} featuring motion data collected from 13 subjects (9 male, 4 female) using smartphone built-in sensors.

\subsubsection{Results}
In this second experiment, we incorporate the derived data generation model into activity recognition models via sample selection.
We select highly informative data sources to form training sets.
During
the training phase, activity recognition models are encouraged to concentrate on the provided subsets of data sources to learn the corresponding human activities. We refer to this setting as {\it w-DGP}, which stands for, with data generation process.

For this, we construct activity recognition models based on neural networks, similar to the architectures used to derive the data generation model, but restricted to 3 Conv1d/ReLU/MaxPool stacked blocks. These blocks are followed by a Fully Connected/ReLU layers.
The weights of the layers corresponding to all inputs are optimized during training without distinction, the constraining being specified via data augmentation.
Indeed, in this setting,
for each subset of interacting data sources, we perform data augmentation by assigning values, drawn from a normal distribution, to the unimportant data sources.
The goal is to make the neural network insensitive to the remaining inputs.
We provide training examples to the neural network according to the given
subsets of interacting data sources that we extract from the derived model.
Furthermore, we experiment with different values of $\tau_{int}$ and $\tau_{imp}$ to extract the subsets of data sources.
For comparison, we train the activity recognition models on the whole data sources of each dataset, i.e., without incorporation of the derived data generation model. These models constitute our baselines and we refer to this setting as {\it wo-DGP}.
In addition, we also incorporate the data generation model based on human expertise (HExp). We refer to this setting as {\it w-HExp}.
%Table~\ref{tab:recognition-performances-w-and-wo-dgp} compares recognition performances obtained, on each dataset, using these settings.
Obtained recognition performances on each dataset are compared and overall, we obtain substantial improvements for all datasets when incorporating a data generation process (either w-HExp or w-DGP).
%\textcolor{red}{
%It is to note, though, that for HTC-TMD, we get a smaller improvement compared to the other datasets.
%This could be related to the limited number of modalities and unavailability of the precise location of the data sources.
%}

\begin{figure}[h!]
    \centering
    \sffamily
    \def\svgwidth{0.8\columnwidth}
    \resizebox{80mm}{!}{
        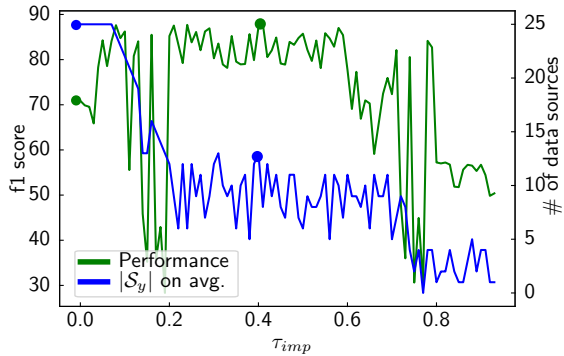
    }
    \caption{
        \small
        Recognition performances as a function of the data source importance threshold $\tau_{imp}$.
        In parallel, the cardinality on average of the subsets $|\mathcal{S}_y|$ used to train the models is shown.
        The left-most points correspond to a configuration where all data sources are used, i.e., no DGP.
    }
    \label{fig:perf_constraining}
\end{figure}

Figure~\ref{fig:perf_constraining} shows the evolution of the obtained recognition performances depending on the parameters $\tau_{int}$ and $\tau_{imp}$.
In addition, this figure illustrates the average number of data sources, that are included in the subsets, depending on these two thresholds.
In particular, when, for example, $\tau_{imp}$ and $\tau_{int}$ are set to 0, all data sources are included.
We find that the neural networks trained with smaller subsets of data sources perform better than the baseline and most of the settings which rely on a higher number of data sources.
Noticeably, we get a recognition performance of 88.7\%$\pm$0.6, measured by the f1-score, using subsets containing on average 12 data sources.
Thus, an improvement over the baseline of 17.84\% in terms of recognition performances and a reduction of one-half concerning the required quantities of data.
Surprisingly, we do not see a lot of bad subsets of interacting data sources for $0.2 \leq \tau_{imp} \leq 0.6$, where the number of data sources per subset is confined between 5 and 12.
It is also worthy to note that in some configurations where $|\mathcal{S}_y|=13$, the trained model performs badly (less than 40\%$\pm$0.16 f1-score).
In the contrary, for smaller subsets ($|\mathcal{S}_y|\leq 5$), trained models get high recognition performances (more than 80\%$\pm$0.05 f1-score).
A Deeper inspection of these configurations reveals that the location of selected data sources plays an important role, in particular, the latter subsets are mainly composed of hips data sources.

\paragraph{Alternative Exploration Strategies}
In the previous experiment, we constrain training of activity recognition models using data generation model derived using the Gaussian process tuner as it had the highest degree of agreement with HExp.
Since the exploration strategies tend to favor different regions of the architecture space, we hypothesize that the derived models will be characterized by variety in terms of combinations of data sources but will still hold the same property, which is being highly informative with regards to the dynamics of body movements.
Here we evaluate the effectiveness of the data generation models derived using the other exploration strategies.
Table~\ref{tab:recognition-performances-dgp-derived-w-different-exploration-strategies} presents the results obtained for this setting on each individual dataset.

\setlength{\belowrulesep}{0pt}
\begin{table*}[t!]
    \caption{
        \small
        Recognition performances of activity recognition models while incorporating the data generation models derived using different space exploration strategies.
    }
    \small
    \centering
    \begin{tabular}{@{\extracolsep{0pt}}lcccccccc@{}}
    \toprule
    \multirow{3}{*}{Dataset}  & \multicolumn{2}{c}{\multirow{2}{*}{\bf Exhaustive}} & \multicolumn{3}{c}{\multirow{2}{*}{\bf Heuristic}} & \multicolumn{3}{c}{\multirow{2}{*}{\bf Sequential}}\\[3pt]
                              & \multicolumn{2}{c}{\bf search}                      & \multicolumn{3}{c}{\bf search}                     & \multicolumn{3}{c}{\bf model-based} \\\cline{2-3}\cline{4-6}\cline{7-9}
                              & Random & Grid & Naive & Anneal & HB & BOHB & TPE & GP Tun.\\
    \midrule
    USC & 79.28\% & 79.58\% & 80.76\% & 83.56\% & 85.27\% & 86.66\% & 82.37\% & {\bf 89.33\%} \\
    \midrule
    HTC-TMD & 76.34\% & 75.17\% & 74.98\% & 73.18\% & 77.45\% & 75.86\% & {\bf 80.13\%} & 78.9\% \\
    \midrule
    US-TMD & 74.14\% & 72.21\% & 79.71\% & 81.13\% & 80.80\% & 79.17\% & {\bf 84.39\%} & 83.64\% \\
    \midrule
    \midrule
    SHL & 72.2\% & 71.32\% & 79.46\% & 84.16\% & 82.33\% & 84.22\% & 86.7\% & {\bf 88.7\%}   \\
    \bottomrule
    \end{tabular}
    \label{tab:recognition-performances-dgp-derived-w-different-exploration-strategies}
\end{table*}

Note that TPE outperforms GP tuner in the case of HTC-TMD and US-TMD datasets.
It is also interesting to note that even though exhaustive search strategies have a low degree of agreement with HExp, incorporation of their corresponding data generation models is competitive for both HTC-TMD and USC-HAD, which
can be explained by the ability of our approach to derive knowledge that is hardly captured by the sole human expertise.

\begin{figure*}[h!]
\captionsetup[subfigure]{labelformat=empty}
\centering
\sffamily
\subfloat[]{
    \def\svgwidth{1.6\columnwidth}
    \resizebox{110mm}{!}{
        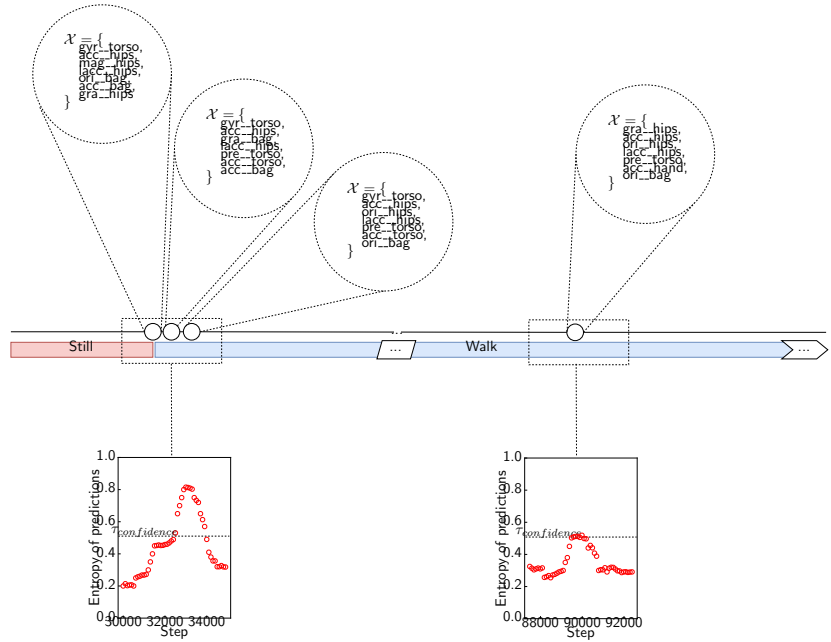
    }
}
\caption{
    %Real monitoring scenario using the proposed continual learning model.
    Evaluation of a continual scenario, in the form of a timeline, showing, on the left, one case of true transition (between still in red and walk in blue) that is correctly detected and, on the right, one case of false alarm.
    Each time a transition (true transition or false alarm) is detected, various sets of data sources (circles on top of the timeline) are selected in order to recognize the current activity.
    At the bottom of the timeline, the evolution of the entropy, as outputted by the continual learning model, and the confidence threshold ($\tau_{confidence}$) are shown.
}
\label{fig:kr-shl-real-scenario}
\end{figure*}
%

%%
%%
%%
%%
%%
%%%%%%%%%%%%%%%%%%%%%%%%%%%%%%%%
%%%%%%%%%%%%%%%%%%%%%%%%%%%%%%%%
%%
%%
%%
%%
%%
%\newpage
\section{Discussion}\label{sec:discussion}
%Notre approche pourrait se résumer en deux niveaux (1) explicitation des biais inductifs au niveau des modèles surrogate et (2) maintien de configurations d'apprentissage (biais inductifs) alternatifs ou concurrents.
The dynamic inductive bias selection perspective that we propose to apply to human activity recognition could be framed into two levels: (1) making explicit the inductive biases related to the complete activity recognition chain (domain knowledge) in the form of surrogate models and (2) maintaining alternative or competing learning configurations (inductive biases) by allowing easy and rapid adaptation to new configurations.
The discussion here is framed around the importance of domain knowledge and the three pillars of our proposed approach which can be stated as questions: (i) which knowledge to encode? (ii) how to encode it? and (iii) How to incorporate it into deployed models?

\subsection{Which knowledge to encode?}
%\bigskip
In our approach, we investigated the benefits of incorporating prior knowledge into neural networks to improve data source integration.
%__
Precisely, we leverage the dynamics of the body movements and the assumption stating that each individual activity is characterized by a different set of gestures which in turn involve specific body parts.
In the case of wearable technologies, where these body parts are equipped with data sources, often, focusing on these specific data sources, allows recognizing a given activity precisely.
As discussed in Section~\ref{sec:dynamic-inductive-bias-selection:which-knowledge-to-encode},
Incorporation of prior knowledge derived, directly or indirectly, from 3D body skeleton-based representations holds an important place in the literature around activity recognition.
Long lines of research proposed to leverage 3D body skeleton-based representation~\cite{vatavu2012multi,kovalenko2014real,papadopoulos2014real,parisi2017emergence,dhiman2019skeleton}, ontology-based representations~\cite{ousmer2019ontology,rodriguez2013understanding}, etc.

This being said, other aspects can be modeled and incorporated into activity recognition models and can affect virtually every step in the recognition chain from the measurement process to the topology of the sensor deployment and the transmission protocols used between the nodes of these deployments.
As seen in Section~\ref{sec:evaluations:analysis-of-the-surrogate-models-response-surface} for the case of segmentation, exhibiting the hyperparameters can be extended to aspects other than the importance and interaction of data sources.
%
%En effet, l'intérêt à terme serait d'expliciter les biais de toutes les étapes de la chaine de reconnaissance d'activités en allant jusqu'aux fonctions de transfers qui, comme illustrés plus-haut, constituent des biais à part entière.
Indeed, the long-term interest would be to make explicit the biases of all the stages of the activity recognition chain by going as far as the transfer functions which, as illustrated in Section~\ref{sec:sensing-constraints} and the introduction of Section~\ref{sec:dynamic-inductive-bias-selection}, constitute biases in their own.

%__
\subsection{How to encode it?}
%\subsubsection{Neural architecture search}
%\bigskip
%__
In our approach, derivation of the dynamics of body movements is based on a neural architecture space.
%__
We use this model to derive subsets of important and interacting data sources based on the hyperparameters controlling the effects they have on the recognition of each individual activity.
%__
These values control how architectural components process each individual input and by the same occasion their influence on the overall architecture performance.
%__
We focused, particularly, on the insights that stem from tuning and adapting these architectures, through their hyperparameters and specifically those controlling the influence of the data sources.
At each layer of a given architecture, setting the right combination of hyperparameters is critical. In particular, choosing the right instantiation for the features learning and sensor fusion components can lead to an architecture capable of building, from the various data sources, an original set of features which is suitable for recognizing a given activity.
%\textcolor{red}{
%
%Derived subsets of data sources are found to impact substantially the recognition of different human activities.
%The effectiveness of the obtained domain models is demonstrated via a setting that exploits these subsets in order to constrain the learning process of a neural network.
%
%The proposed setting showed substantial improvements over the baseline where all data sources of the sensor-rich environment are used.
%Noticeably, we get an improvement of the recognition performances, measured by the f1-score, of up to 17.84\% over the baseline and this, using solely 12 data sources among those available.
%}

%__
In our proposed approach, the neural architecture space is assimilated to a high-capacity
%~\footnote{Capacity refer to the generalization ability (complexity, representativeness, richness, flexibility) of a given model in the sense of Vapnik's definition~\cite{vapnik1995nature}.}
model (proxy).
We use this high-capacity model for its ability to encode domain knowledge and store lots of experience.
Making explicit these biases via hyperparameters is motivated by several aspects, the most important being their capacity to play the role of inductive biases far more than the parameters of a model.
Indeed, the biases of the architecture (e.g. CNN for vision, LSTM for time-dependent sequences, etc.) are decisive for the tasks for which they were originally designed.
More importantly, several empirical results are backing the fact that the hyperparameters are playing a more important role in the final recognition performances than the models' parameters~\cite{gaier2019weight}.
These results are also in the direction of rapid adaptation and the use of few learning steps since the weights (parameters) of the models are less influential than the hyperparameters.

%Toute la question est de concevoir de bonnes fonctions d'acquisition qui à la fois ont un bon compromis entre exploration et exploitation (qui permet d'obtenir une image assez parlante de l'espace des architectures) et reflètent en même temps la connaissance métier visée (l'exploration doit avoir l'information que l'aspect visé est le biais lié à la segmentation, aux sources de données, au pre-processing, etc.). Concernant ce point, l'architecture multi-modale présentée ici va dans ce sense (les hyperparametres ont été conçu de manière à refléter l'impact des sources de données).
As we saw, the exploration of the architecture space is often based on an acquisition function (responsible of choosing the next configuration to explore), the whole issue is to design good acquisition functions that both have a good compromise between exploration and exploitation (which gives a fairly meaningful picture of the space of architectures) and at the same time reflect the targeted domain knowledge (the exploration must have the information that, for example, the targeted aspect is the bias related to segmentation, data sources, pre-processing, etc.). Concerning this point, the multi-modal architecture presented in Section~\ref{sec:dynamic-inductive-bias-selection:how-knowledge-is-concretely-encoded} goes in this direction (the hyperparameters have been designed to reflect the impact of the data sources and their interactions).

%__
\subsection{How to incorporate knowledge into deployed models?}
The way incorporating knowledge, in a principled fashion, into deployed models remains also an open question as it is the case for the other aspects that we are investigating.
%
%En effet, dans l'approche que nous avons proposé, ce n'était pas tant la meilleure architecture qui nous intéressait (et qui aurait été déployée directement sans passer par des modèles de moindre capacité) mais plutôt le comportement global des architectures explorées. Ce comportement a été ensuite incorporé à des modèles plus restreints en terms de capacité, et donc facile à entrainer et à adapter (peu d'entrées, peu d'entropie, c'est un lien qui peut être exploré aussi). Ceci dit, d'autres manières de faire, qui se baseraient sur des techniques existantes, pourraient être mise en oeuvre. Dans ce qui suit, nous présentons quelques pistes qui peuvent être investiguées en ce sense.
Indeed, in the approach we proposed, it was not the best architecture that we have been interested in (and which would have been deployed directly without going through models of lower capacity) but rather the overall behavior of the architectures explored. This behavior was then incorporated into models that were more restricted in terms of capacity, and therefore easy to train and adapt.
%(few inputs, little entropy, it is a link that can be explored too).
That said, other ways of achieving this, by leveraging on existing techniques, could be implemented. In what follows, we present some avenues that can be investigated in this sense.

%Various approaches have been developed \dots

\textbf{Regularization.}
Regularization techniques have been investigated as ways of incorporating domain knowledge into machine learning models.
%__
Beyond activity recognition, many other applications leverage domain models to enforce certain conditions or equations, which are part of prior knowledge, within machine learning models.
%__
In~\cite{stewart2017label,nabian2020physics}, authors propose to incorporate domain knowledge, like known laws of physics, by constraining neural networks via regularization.
Their settings introduce new challenges for encoding knowledge into appropriate loss functions and avoiding trivial solutions in the constraint space.

\textbf{Attention mechanisms.}
Some other works use additional computational levels, like attention mechanisms, first used in natural language processing~\cite{bahdanau2014neural}, in order to help models focus on specific entries.
In this vein, authors in~\cite{zeng2018understanding} propose two  attention  models  for  human  activity  recognition:  temporal attention and sensor attention.  These two mechanisms adaptively focus on important signals and sensor modalities.
%__
In this sense, our approach is based on the selection of relevant and more informative data sources for the recognition of each individual activities.
Our approach can be related to that of~\cite{zeng2018understanding}, with the difference that selection is performed beforehand, in a separate procedure, before incorporating them into the main model.

\textbf{Pruning.}
Sparsifying neural networks via pruning is also a way to go in order to incorporate accumulated knowledge and at the same time. In~\cite{tartaglione2018learning}, for example, authors exploited the sensitivity between inputs and outputs in order to eliminate model's weights which are not responsive enough to the input-output pairs stimulus during training. This process can be adapted to our proposed approach.

%\textbf{Similarity control and Knowledge transfer.}
%Vapnik \& Vashist (2009); Vapnik \& Izmailov (2015) offer two strategies to learn using privileged information: similarity control and knowledge transfer.
%
%In~\cite{abdallah2018activity}, authors explored transfer learning to deal with evolution of the learning configurations ("model personalization to best fit a specific user; model adaptation by adding new activities or deleting abandoned ones.").

\textbf{Neuromodulation in neural networks.}
In~\cite{kovalenko2014real}, for example, authors construct an ontology that serves as a basis for constructing a network of Bayesian inference while in~\cite{parisi2017emergence}, the constructed representations, which obey to certain constraints, help the neural networks to self-organize.
The proposed hierarchical processing of visual inputs allows obtaining progressively specialized neurons encoding latent spatiotemporal dynamics of the input data sources.
In~\cite{vecoven2020introducing}, authors draw inspiration from cellular neuromodulation to construct a new deep neural network architecture that is specifically designed to learn adaptive behaviors. The neural architecture comprises two neural networks: a main network and a neuromodulatory network. The neuromodulatory network processes feedback and contextual data whereas the main network is in charge of processing other inputs.

%\section*{References}

%\bibliography{biblio}

\end{document}

%% file: img/basic-learning-setting.pdf_tex
%% Accompanies image file 'basic-learning-setting.pdf' (pdf, eps, ps)
\begingroup%
  \makeatletter%
  \providecommand\color[2][]{%
    \errmessage{(Inkscape) Color is used for the text in Inkscape, but the package 'color.sty' is not loaded}%
    \renewcommand\color[2][]{}%
  }%
  \providecommand\transparent[1]{%
    \errmessage{(Inkscape) Transparency is used (non-zero) for the text in Inkscape, but the package 'transparent.sty' is not loaded}%
    \renewcommand\transparent[1]{}%
  }%
  \providecommand\rotatebox[2]{#2}%
  \newcommand*\fsize{\dimexpr\f@size pt\relax}%
  \newcommand*\lineheight[1]{\fontsize{\fsize}{#1\fsize}\selectfont}%
  \ifx\svgwidth\undefined%
    \setlength{\unitlength}{573.2204625bp}%
    \ifx\svgscale\undefined%
      \relax%
    \else%
      \setlength{\unitlength}{\unitlength * \real{\svgscale}}%
    \fi%
  \else%
    \setlength{\unitlength}{\svgwidth}%
  \fi%
  \global\let\svgwidth\undefined%
  \global\let\svgscale\undefined%
  \makeatother%
  \begin{picture}(1,0.36363823)%
    \lineheight{1}%
    \setlength\tabcolsep{0pt}%
    \put(0,0){\includegraphics[width=\unitlength,page=1]{basic-learning-setting.pdf}}%
  \end{picture}%
\endgroup%

%% file: img/dynamic-selection-of-inductive-biases.pdf_tex
%% Accompanies image file 'dynamic-selection-of-inductive-biases.pdf' (pdf, eps, ps)
\begingroup%
  \makeatletter%
  \providecommand\color[2][]{%
    \errmessage{(Inkscape) Color is used for the text in Inkscape, but the package 'color.sty' is not loaded}%
    \renewcommand\color[2][]{}%
  }%
  \providecommand\transparent[1]{%
    \errmessage{(Inkscape) Transparency is used (non-zero) for the text in Inkscape, but the package 'transparent.sty' is not loaded}%
    \renewcommand\transparent[1]{}%
  }%
  \providecommand\rotatebox[2]{#2}%
  \newcommand*\fsize{\dimexpr\f@size pt\relax}%
  \newcommand*\lineheight[1]{\fontsize{\fsize}{#1\fsize}\selectfont}%
  \ifx\svgwidth\undefined%
    \setlength{\unitlength}{597.936975bp}%
    \ifx\svgscale\undefined%
      \relax%
    \else%
      \setlength{\unitlength}{\unitlength * \real{\svgscale}}%
    \fi%
  \else%
    \setlength{\unitlength}{\svgwidth}%
  \fi%
  \global\let\svgwidth\undefined%
  \global\let\svgscale\undefined%
  \makeatother%
  \begin{picture}(1,0.35180017)%
    \lineheight{1}%
    \setlength\tabcolsep{0pt}%
    \put(0,0){\includegraphics[width=\unitlength,page=1]{dynamic-selection-of-inductive-biases.pdf}}%
  \end{picture}%
\endgroup%

%% file: img/domain-models-based-learning-setting.pdf_tex
%% Accompanies image file 'domain-models-based-learning-setting.pdf' (pdf, eps, ps)
\begingroup%
  \makeatletter%
  \providecommand\color[2][]{%
    \errmessage{(Inkscape) Color is used for the text in Inkscape, but the package 'color.sty' is not loaded}%
    \renewcommand\color[2][]{}%
  }%
  \providecommand\transparent[1]{%
    \errmessage{(Inkscape) Transparency is used (non-zero) for the text in Inkscape, but the package 'transparent.sty' is not loaded}%
    \renewcommand\transparent[1]{}%
  }%
  \providecommand\rotatebox[2]{#2}%
  \newcommand*\fsize{\dimexpr\f@size pt\relax}%
  \newcommand*\lineheight[1]{\fontsize{\fsize}{#1\fsize}\selectfont}%
  \ifx\svgwidth\undefined%
    \setlength{\unitlength}{589.9409625bp}%
    \ifx\svgscale\undefined%
      \relax%
    \else%
      \setlength{\unitlength}{\unitlength * \real{\svgscale}}%
    \fi%
  \else%
    \setlength{\unitlength}{\svgwidth}%
  \fi%
  \global\let\svgwidth\undefined%
  \global\let\svgscale\undefined%
  \makeatother%
  \begin{picture}(1,0.41497213)%
    \lineheight{1}%
    \setlength\tabcolsep{0pt}%
    \put(0,0){\includegraphics[width=\unitlength,page=1]{domain-models-based-learning-setting.pdf}}%
  \end{picture}%
\endgroup%

%% file: img/kr-nas-based-framework.pdf_tex
%% Accompanies image file 'kr-nas-based-framework.pdf' (pdf, eps, ps)
\begingroup%
  \makeatletter%
  \providecommand\color[2][]{%
    \errmessage{(Inkscape) Color is used for the text in Inkscape, but the package 'color.sty' is not loaded}%
    \renewcommand\color[2][]{}%
  }%
  \providecommand\transparent[1]{%
    \errmessage{(Inkscape) Transparency is used (non-zero) for the text in Inkscape, but the package 'transparent.sty' is not loaded}%
    \renewcommand\transparent[1]{}%
  }%
  \providecommand\rotatebox[2]{#2}%
  \newcommand*\fsize{\dimexpr\f@size pt\relax}%
  \newcommand*\lineheight[1]{\fontsize{\fsize}{#1\fsize}\selectfont}%
  \ifx\svgwidth\undefined%
    \setlength{\unitlength}{181.77806501bp}%
    \ifx\svgscale\undefined%
      \relax%
    \else%
      \setlength{\unitlength}{\unitlength * \real{\svgscale}}%
    \fi%
  \else%
    \setlength{\unitlength}{\svgwidth}%
  \fi%
  \global\let\svgwidth\undefined%
  \global\let\svgscale\undefined%
  \makeatother%
  \begin{picture}(1,2.27750251)%
    \lineheight{1}%
    \setlength\tabcolsep{0pt}%
    \put(0,0){\includegraphics[width=\unitlength,page=1]{kr-nas-based-framework.pdf}}%
    \put(0.54562162,1.42756498){\makebox(0,0)[t]{\lineheight{1.25}\smash{\begin{tabular}[t]{c}FF\end{tabular}}}}%
    \put(0,0){\includegraphics[width=\unitlength,page=2]{kr-nas-based-framework.pdf}}%
    \put(0.76689415,1.46882408){\makebox(0,0)[t]{\lineheight{1.25}\smash{\begin{tabular}[t]{c}$F_{1,n}$\end{tabular}}}}%
    \put(0,0){\includegraphics[width=\unitlength,page=3]{kr-nas-based-framework.pdf}}%
    \put(0.17065887,1.61323095){\makebox(0,0)[t]{\lineheight{1.25}\smash{\begin{tabular}[t]{c}$F_1$\end{tabular}}}}%
    \put(0,0){\includegraphics[width=\unitlength,page=4]{kr-nas-based-framework.pdf}}%
    \put(0.17065887,1.53896456){\makebox(0,0)[t]{\lineheight{1.25}\smash{\begin{tabular}[t]{c}$F_2$\end{tabular}}}}%
    \put(0,0){\includegraphics[width=\unitlength,page=5]{kr-nas-based-framework.pdf}}%
    \put(0.17065887,1.26665448){\makebox(0,0)[t]{\lineheight{1.25}\smash{\begin{tabular}[t]{c}$F_n$\end{tabular}}}}%
    \put(0,0){\includegraphics[width=\unitlength,page=6]{kr-nas-based-framework.pdf}}%
    \put(0.54562162,0.8375598){\makebox(0,0)[t]{\lineheight{1.25}\smash{\begin{tabular}[t]{c}DF\end{tabular}}}}%
    \put(0,0){\includegraphics[width=\unitlength,page=7]{kr-nas-based-framework.pdf}}%
    \put(0.75913749,0.8788189){\makebox(0,0)[t]{\lineheight{1.25}\smash{\begin{tabular}[t]{c}$D_{1,n}$\end{tabular}}}}%
    \put(0,0){\includegraphics[width=\unitlength,page=8]{kr-nas-based-framework.pdf}}%
    \put(0.54921115,0.23105098){\makebox(0,0)[t]{\lineheight{1.25}\smash{\begin{tabular}[t]{c}AU\end{tabular}}}}%
    \put(0,0){\includegraphics[width=\unitlength,page=9]{kr-nas-based-framework.pdf}}%
    \put(0.77097883,0.26405826){\makebox(0,0)[t]{\lineheight{1.25}\smash{\begin{tabular}[t]{c}$D$\end{tabular}}}}%
    \put(0,0){\includegraphics[width=\unitlength,page=10]{kr-nas-based-framework.pdf}}%
    \put(0.14854399,0.23105098){\makebox(0,0)[t]{\lineheight{1.25}\smash{\begin{tabular}[t]{c}$F_1$/$D_1$\end{tabular}}}}%
    \put(0,0){\includegraphics[width=\unitlength,page=11]{kr-nas-based-framework.pdf}}%
    \put(0.17065887,1.01909986){\makebox(0,0)[t]{\lineheight{1.25}\smash{\begin{tabular}[t]{c}$D_1$\end{tabular}}}}%
    \put(0,0){\includegraphics[width=\unitlength,page=12]{kr-nas-based-framework.pdf}}%
    \put(0.17065887,0.94483347){\makebox(0,0)[t]{\lineheight{1.25}\smash{\begin{tabular}[t]{c}$D_2$\end{tabular}}}}%
    \put(0,0){\includegraphics[width=\unitlength,page=13]{kr-nas-based-framework.pdf}}%
    \put(0.17065887,0.6766493){\makebox(0,0)[t]{\lineheight{1.25}\smash{\begin{tabular}[t]{c}$D_n$\end{tabular}}}}%
    \put(0,0){\includegraphics[width=\unitlength,page=14]{kr-nas-based-framework.pdf}}%
    \put(0.54925238,2.00931834){\makebox(0,0)[t]{\lineheight{1.25}\smash{\begin{tabular}[t]{c}FE\end{tabular}}}}%
    \put(0,0){\includegraphics[width=\unitlength,page=15]{kr-nas-based-framework.pdf}}%
    \put(0.77102006,2.04232562){\makebox(0,0)[t]{\lineheight{1.25}\smash{\begin{tabular}[t]{c}$F$\end{tabular}}}}%
    \put(0,0){\includegraphics[width=\unitlength,page=16]{kr-nas-based-framework.pdf}}%
    \put(0.13587745,2.00931834){\makebox(0,0)[t]{\lineheight{1.25}\smash{\begin{tabular}[t]{c}$I_1$\end{tabular}}}}%
    \put(0,0){\includegraphics[width=\unitlength,page=17]{kr-nas-based-framework.pdf}}%
    \put(0.37080679,2.04438858){\makebox(0,0)[t]{\lineheight{1.25}\smash{\begin{tabular}[t]{c}$h_1$\end{tabular}}}}%
    \put(0,0){\includegraphics[width=\unitlength,page=18]{kr-nas-based-framework.pdf}}%
    \put(0.37076552,1.64417527){\makebox(0,0)[t]{\lineheight{1.25}\smash{\begin{tabular}[t]{c}$h_1$\end{tabular}}}}%
    \put(0,0){\includegraphics[width=\unitlength,page=19]{kr-nas-based-framework.pdf}}%
    \put(0.37076552,1.56990889){\makebox(0,0)[t]{\lineheight{1.25}\smash{\begin{tabular}[t]{c}$h_2$\end{tabular}}}}%
    \put(0,0){\includegraphics[width=\unitlength,page=20]{kr-nas-based-framework.pdf}}%
    \put(0.37076552,1.29759881){\makebox(0,0)[t]{\lineheight{1.25}\smash{\begin{tabular}[t]{c}$h_n$\end{tabular}}}}%
    \put(0,0){\includegraphics[width=\unitlength,page=21]{kr-nas-based-framework.pdf}}%
    \put(0.37076552,1.05417009){\makebox(0,0)[t]{\lineheight{1.25}\smash{\begin{tabular}[t]{c}$h_1$\end{tabular}}}}%
    \put(0,0){\includegraphics[width=\unitlength,page=22]{kr-nas-based-framework.pdf}}%
    \put(0.37076552,0.97990371){\makebox(0,0)[t]{\lineheight{1.25}\smash{\begin{tabular}[t]{c}$h_2$\end{tabular}}}}%
    \put(0,0){\includegraphics[width=\unitlength,page=23]{kr-nas-based-framework.pdf}}%
    \put(0.37076552,0.71171954){\makebox(0,0)[t]{\lineheight{1.25}\smash{\begin{tabular}[t]{c}$h_n$\end{tabular}}}}%
    \put(0,0){\includegraphics[width=\unitlength,page=24]{kr-nas-based-framework.pdf}}%
    \put(0.37076552,0.26612122){\makebox(0,0)[t]{\lineheight{1.25}\smash{\begin{tabular}[t]{c}$h_1$\end{tabular}}}}%
  \end{picture}%
\endgroup%

%% file: img/kr-nas-based-framework-architecture-space.pdf_tex
%% Accompanies image file 'kr-nas-based-framework-architecture-space.pdf' (pdf, eps, ps)
\begingroup%
  \makeatletter%
  \providecommand\color[2][]{%
    \errmessage{(Inkscape) Color is used for the text in Inkscape, but the package 'color.sty' is not loaded}%
    \renewcommand\color[2][]{}%
  }%
  \providecommand\transparent[1]{%
    \errmessage{(Inkscape) Transparency is used (non-zero) for the text in Inkscape, but the package 'transparent.sty' is not loaded}%
    \renewcommand\transparent[1]{}%
  }%
  \providecommand\rotatebox[2]{#2}%
  \newcommand*\fsize{\dimexpr\f@size pt\relax}%
  \newcommand*\lineheight[1]{\fontsize{\fsize}{#1\fsize}\selectfont}%
  \ifx\svgwidth\undefined%
    \setlength{\unitlength}{254.46005515bp}%
    \ifx\svgscale\undefined%
      \relax%
    \else%
      \setlength{\unitlength}{\unitlength * \real{\svgscale}}%
    \fi%
  \else%
    \setlength{\unitlength}{\svgwidth}%
  \fi%
  \global\let\svgwidth\undefined%
  \global\let\svgscale\undefined%
  \makeatother%
  \begin{picture}(1,1.76845045)%
    \lineheight{1}%
    \setlength\tabcolsep{0pt}%
    \put(0,0){\includegraphics[width=\unitlength,page=1]{kr-nas-based-framework-architecture-space.pdf}}%
    \put(0.38104234,0.34337413){\makebox(0,0)[t]{\lineheight{1.25}\smash{\begin{tabular}[t]{c}$C_2$\end{tabular}}}}%
    \put(0,0){\includegraphics[width=\unitlength,page=2]{kr-nas-based-framework-architecture-space.pdf}}%
    \put(0.63009911,0.34337413){\makebox(0,0)[t]{\lineheight{1.25}\smash{\begin{tabular}[t]{c}$C_3$\end{tabular}}}}%
    \put(0,0){\includegraphics[width=\unitlength,page=3]{kr-nas-based-framework-architecture-space.pdf}}%
    \put(0.75073689,0.78548674){\makebox(0,0)[t]{\lineheight{1.25}\smash{\begin{tabular}[t]{c}$C_6$\end{tabular}}}}%
    \put(0,0){\includegraphics[width=\unitlength,page=4]{kr-nas-based-framework-architecture-space.pdf}}%
    \put(0.12314332,0.34337413){\makebox(0,0)[t]{\lineheight{1.25}\smash{\begin{tabular}[t]{c}$C_1$\end{tabular}}}}%
    \put(0,0){\includegraphics[width=\unitlength,page=5]{kr-nas-based-framework-architecture-space.pdf}}%
    \put(0.43831064,1.05075431){\makebox(0,0)[t]{\lineheight{1.25}\smash{\begin{tabular}[t]{c}$C_7$\end{tabular}}}}%
    \put(0,0){\includegraphics[width=\unitlength,page=6]{kr-nas-based-framework-architecture-space.pdf}}%
    \put(0.75112007,1.05075431){\makebox(0,0)[t]{\lineheight{1.25}\smash{\begin{tabular}[t]{c}$C_8$\end{tabular}}}}%
    \put(0,0){\includegraphics[width=\unitlength,page=7]{kr-nas-based-framework-architecture-space.pdf}}%
    \put(0.38104234,1.31602188){\makebox(0,0)[t]{\lineheight{1.25}\smash{\begin{tabular}[t]{c}$C_9$\end{tabular}}}}%
    \put(0,0){\includegraphics[width=\unitlength,page=8]{kr-nas-based-framework-architecture-space.pdf}}%
    \put(0.59620381,1.5223411){\makebox(0,0)[t]{\lineheight{1.25}\smash{\begin{tabular}[t]{c}$C_{10}$\end{tabular}}}}%
    \put(0,0){\includegraphics[width=\unitlength,page=9]{kr-nas-based-framework-architecture-space.pdf}}%
    \put(0.59620381,1.69918614){\makebox(0,0)[t]{\lineheight{1.25}\smash{\begin{tabular}[t]{c}output\end{tabular}}}}%
    \put(0,0){\includegraphics[width=\unitlength,page=10]{kr-nas-based-framework-architecture-space.pdf}}%
    \put(0.573715,0.04273755){\makebox(0,0)[t]{\lineheight{1.25}\smash{\begin{tabular}[t]{c}$s_j$\end{tabular}}}}%
    \put(0,0){\includegraphics[width=\unitlength,page=11]{kr-nas-based-framework-architecture-space.pdf}}%
    \put(0.12314332,0.04273755){\makebox(0,0)[t]{\lineheight{1.25}\smash{\begin{tabular}[t]{c}$s_i$\end{tabular}}}}%
    \put(0,0){\includegraphics[width=\unitlength,page=12]{kr-nas-based-framework-architecture-space.pdf}}%
    \put(0.25282968,0.78548674){\makebox(0,0)[t]{\lineheight{1.25}\smash{\begin{tabular}[t]{c}$C_5$\end{tabular}}}}%
    \put(0,0){\includegraphics[width=\unitlength,page=13]{kr-nas-based-framework-architecture-space.pdf}}%
    \put(0.88947185,0.34337413){\makebox(0,0)[t]{\lineheight{1.25}\smash{\begin{tabular}[t]{c}$C_4$\end{tabular}}}}%
  \end{picture}%
\endgroup%

%% file: img/kr-shl-infrastructure-promotional-bis.pdf_tex
%% Accompanies image file 'kr-shl-infrastructure-promotional-bis.pdf' (pdf, eps, ps)
\begingroup%
  \makeatletter%
  \providecommand\color[2][]{%
    \errmessage{(Inkscape) Color is used for the text in Inkscape, but the package 'color.sty' is not loaded}%
    \renewcommand\color[2][]{}%
  }%
  \providecommand\transparent[1]{%
    \errmessage{(Inkscape) Transparency is used (non-zero) for the text in Inkscape, but the package 'transparent.sty' is not loaded}%
    \renewcommand\transparent[1]{}%
  }%
  \providecommand\rotatebox[2]{#2}%
  \newcommand*\fsize{\dimexpr\f@size pt\relax}%
  \newcommand*\lineheight[1]{\fontsize{\fsize}{#1\fsize}\selectfont}%
  \ifx\svgwidth\undefined%
    \setlength{\unitlength}{1338.75bp}%
    \ifx\svgscale\undefined%
      \relax%
    \else%
      \setlength{\unitlength}{\unitlength * \real{\svgscale}}%
    \fi%
  \else%
    \setlength{\unitlength}{\svgwidth}%
  \fi%
  \global\let\svgwidth\undefined%
  \global\let\svgscale\undefined%
  \makeatother%
  \begin{picture}(1,1.08963585)%
    \lineheight{1}%
    \setlength\tabcolsep{0pt}%
    \put(0,0){\includegraphics[width=\unitlength,page=1]{kr-shl-infrastructure-promotional-bis.pdf}}%
    \put(0.85546218,1.04313725){\makebox(0,0)[t]{\lineheight{1.25}\smash{\begin{tabular}[t]{c}Simplified representation\end{tabular}}}}%
    \put(0.85546218,1.02857143){\makebox(0,0)[t]{\lineheight{1.25}\smash{\begin{tabular}[t]{c}of the hyperparameter\end{tabular}}}}%
    \put(0.85546218,1.0140056){\makebox(0,0)[t]{\lineheight{1.25}\smash{\begin{tabular}[t]{c}space\end{tabular}}}}%
    \put(0,0){\includegraphics[width=\unitlength,page=2]{kr-shl-infrastructure-promotional-bis.pdf}}%
    \put(0.85546218,0.87731092){\makebox(0,0)[t]{\lineheight{1.25}\smash{\begin{tabular}[t]{c}Hyperparameter\end{tabular}}}}%
    \put(0.85546218,0.8627451){\makebox(0,0)[t]{\lineheight{1.25}\smash{\begin{tabular}[t]{c}instantiations\end{tabular}}}}%
    \put(0,0){\includegraphics[width=\unitlength,page=3]{kr-shl-infrastructure-promotional-bis.pdf}}%
    \put(0.85770308,0.72380952){\makebox(0,0)[t]{\lineheight{1.25}\smash{\begin{tabular}[t]{c}Analysis\end{tabular}}}}%
    \put(0.85770308,0.7092437){\makebox(0,0)[t]{\lineheight{1.25}\smash{\begin{tabular}[t]{c}of the global\end{tabular}}}}%
    \put(0.85770308,0.69467787){\makebox(0,0)[t]{\lineheight{1.25}\smash{\begin{tabular}[t]{c}effects\end{tabular}}}}%
    \put(0,0){\includegraphics[width=\unitlength,page=4]{kr-shl-infrastructure-promotional-bis.pdf}}%
    \put(0.78431373,0.06806723){\makebox(0,0)[lt]{\lineheight{1.25}\smash{\begin{tabular}[t]{l}Relative\end{tabular}}}}%
    \put(0.78431373,0.0535014){\makebox(0,0)[lt]{\lineheight{1.25}\smash{\begin{tabular}[t]{l}Importance\end{tabular}}}}%
    \put(0,0){\includegraphics[width=\unitlength,page=5]{kr-shl-infrastructure-promotional-bis.pdf}}%
    \put(0.93557423,0.08179272){\makebox(0,0)[lt]{\lineheight{1.25}\smash{\begin{tabular}[t]{l}0.02\end{tabular}}}}%
    \put(0,0){\includegraphics[width=\unitlength,page=6]{kr-shl-infrastructure-promotional-bis.pdf}}%
    \put(0.93557423,0.07086835){\makebox(0,0)[lt]{\lineheight{1.25}\smash{\begin{tabular}[t]{l}0.015\end{tabular}}}}%
    \put(0,0){\includegraphics[width=\unitlength,page=7]{kr-shl-infrastructure-promotional-bis.pdf}}%
    \put(0.93585434,0.05014006){\makebox(0,0)[lt]{\lineheight{1.25}\smash{\begin{tabular}[t]{l}0.005\end{tabular}}}}%
    \put(0,0){\includegraphics[width=\unitlength,page=8]{kr-shl-infrastructure-promotional-bis.pdf}}%
    \put(0.93557423,0.0605042){\makebox(0,0)[lt]{\lineheight{1.25}\smash{\begin{tabular}[t]{l}0.01\end{tabular}}}}%
    \put(0,0){\includegraphics[width=\unitlength,page=9]{kr-shl-infrastructure-promotional-bis.pdf}}%
    \put(0.93529412,0.02829132){\makebox(0,0)[lt]{\lineheight{1.25}\smash{\begin{tabular}[t]{l}0.001\end{tabular}}}}%
    \put(0,0){\includegraphics[width=\unitlength,page=10]{kr-shl-infrastructure-promotional-bis.pdf}}%
    \put(0.93529412,0.01988796){\makebox(0,0)[lt]{\lineheight{1.25}\smash{\begin{tabular}[t]{l}0.0005\end{tabular}}}}%
    \put(0,0){\includegraphics[width=\unitlength,page=11]{kr-shl-infrastructure-promotional-bis.pdf}}%
    \put(0.93529412,0.01204482){\makebox(0,0)[lt]{\lineheight{1.25}\smash{\begin{tabular}[t]{l}0.0001\end{tabular}}}}%
    \put(0,0){\includegraphics[width=\unitlength,page=12]{kr-shl-infrastructure-promotional-bis.pdf}}%
    \put(0.78431373,0.02661064){\makebox(0,0)[lt]{\lineheight{1.25}\smash{\begin{tabular}[t]{l}Pairwise\end{tabular}}}}%
    \put(0.78431373,0.01204482){\makebox(0,0)[lt]{\lineheight{1.25}\smash{\begin{tabular}[t]{l}Marginal\end{tabular}}}}%
    \put(0,0){\includegraphics[width=\unitlength,page=13]{kr-shl-infrastructure-promotional-bis.pdf}}%
    \put(0.85078433,0.54733894){\makebox(0,0)[t]{\lineheight{1.25}\smash{\begin{tabular}[t]{c}Data generation model\end{tabular}}}}%
    \put(0,0){\includegraphics[width=\unitlength,page=14]{kr-shl-infrastructure-promotional-bis.pdf}}%
    \put(0.84929972,0.29551821){\makebox(0,0)[t]{\lineheight{1.25}\smash{\begin{tabular}[t]{c}Topology of\end{tabular}}}}%
    \put(0.84929972,0.28095238){\makebox(0,0)[t]{\lineheight{1.25}\smash{\begin{tabular}[t]{c}the sensors\end{tabular}}}}%
    \put(0.84929972,0.26638655){\makebox(0,0)[t]{\lineheight{1.25}\smash{\begin{tabular}[t]{c}deployment\end{tabular}}}}%
    \put(0.84929972,0.25182073){\makebox(0,0)[t]{\lineheight{1.25}\smash{\begin{tabular}[t]{c}(Wearables)\end{tabular}}}}%
    \put(0,0){\includegraphics[width=\unitlength,page=15]{kr-shl-infrastructure-promotional-bis.pdf}}%
    \put(0.78431373,0.14313725){\makebox(0,0)[lt]{\lineheight{1.25}\smash{\begin{tabular}[t]{l}Expected\end{tabular}}}}%
    \put(0.78431373,0.12857143){\makebox(0,0)[lt]{\lineheight{1.25}\smash{\begin{tabular}[t]{l}recognition\end{tabular}}}}%
    \put(0.78431373,0.1140056){\makebox(0,0)[lt]{\lineheight{1.25}\smash{\begin{tabular}[t]{l}performance\end{tabular}}}}%
    \put(0,0){\includegraphics[width=\unitlength,page=16]{kr-shl-infrastructure-promotional-bis.pdf}}%
    \put(0.90308123,0.15910364){\makebox(0,0)[lt]{\lineheight{1.25}\smash{\begin{tabular}[t]{l}100\%\end{tabular}}}}%
    \put(0,0){\includegraphics[width=\unitlength,page=17]{kr-shl-infrastructure-promotional-bis.pdf}}%
    \put(0.90308123,0.14453782){\makebox(0,0)[lt]{\lineheight{1.25}\smash{\begin{tabular}[t]{l}80\%\end{tabular}}}}%
    \put(0,0){\includegraphics[width=\unitlength,page=18]{kr-shl-infrastructure-promotional-bis.pdf}}%
    \put(0.90252101,0.13109244){\makebox(0,0)[lt]{\lineheight{1.25}\smash{\begin{tabular}[t]{l}60\%\end{tabular}}}}%
    \put(0,0){\includegraphics[width=\unitlength,page=19]{kr-shl-infrastructure-promotional-bis.pdf}}%
    \put(0.90252101,0.11540616){\makebox(0,0)[lt]{\lineheight{1.25}\smash{\begin{tabular}[t]{l}40\%\end{tabular}}}}%
    \put(0,0){\includegraphics[width=\unitlength,page=20]{kr-shl-infrastructure-promotional-bis.pdf}}%
    \put(0.90252101,0.10196078){\makebox(0,0)[lt]{\lineheight{1.25}\smash{\begin{tabular}[t]{l}$\leq$20\%\end{tabular}}}}%
    \put(0,0){\includegraphics[width=\unitlength,page=21]{kr-shl-infrastructure-promotional-bis.pdf}}%
    \put(0.97843137,1.02969188){\makebox(0,0)[t]{\lineheight{1.25}\smash{\begin{tabular}[t]{c}1\end{tabular}}}}%
    \put(0,0){\includegraphics[width=\unitlength,page=22]{kr-shl-infrastructure-promotional-bis.pdf}}%
    \put(0.97787115,0.86218487){\makebox(0,0)[t]{\lineheight{1.25}\smash{\begin{tabular}[t]{c}2\end{tabular}}}}%
    \put(0,0){\includegraphics[width=\unitlength,page=23]{kr-shl-infrastructure-promotional-bis.pdf}}%
    \put(0.98263305,0.70560224){\makebox(0,0)[t]{\lineheight{1.25}\smash{\begin{tabular}[t]{c}3\end{tabular}}}}%
    \put(0,0){\includegraphics[width=\unitlength,page=24]{kr-shl-infrastructure-promotional-bis.pdf}}%
    \put(0.98151261,0.51176471){\makebox(0,0)[t]{\lineheight{1.25}\smash{\begin{tabular}[t]{c}4\end{tabular}}}}%
    \put(0,0){\includegraphics[width=\unitlength,page=25]{kr-shl-infrastructure-promotional-bis.pdf}}%
    \put(0.97787115,0.26918768){\makebox(0,0)[t]{\lineheight{1.25}\smash{\begin{tabular}[t]{c}5\end{tabular}}}}%
    \put(0,0){\includegraphics[width=\unitlength,page=26]{kr-shl-infrastructure-promotional-bis.pdf}}%
    \put(0.86190476,0.5162465){\makebox(0,0)[t]{\lineheight{1.25}\smash{\begin{tabular}[t]{c}Interaction importance\end{tabular}}}}%
    \put(0,0){\includegraphics[width=\unitlength,page=27]{kr-shl-infrastructure-promotional-bis.pdf}}%
    \put(0.86778711,0.4837535){\makebox(0,0)[t]{\lineheight{1.25}\smash{\begin{tabular}[t]{c}Data source importance\end{tabular}}}}%
    \put(0,0){\includegraphics[width=\unitlength,page=28]{kr-shl-infrastructure-promotional-bis.pdf}}%
  \end{picture}%
\endgroup%

%% file: img/kr-shl-deployment-topology-running.pdf_tex
%% Accompanies image file 'kr-shl-deployment-topology-running.pdf' (pdf, eps, ps)
\begingroup%
  \makeatletter%
  \providecommand\color[2][]{%
    \errmessage{(Inkscape) Color is used for the text in Inkscape, but the package 'color.sty' is not loaded}%
    \renewcommand\color[2][]{}%
  }%
  \providecommand\transparent[1]{%
    \errmessage{(Inkscape) Transparency is used (non-zero) for the text in Inkscape, but the package 'transparent.sty' is not loaded}%
    \renewcommand\transparent[1]{}%
  }%
  \providecommand\rotatebox[2]{#2}%
  \newcommand*\fsize{\dimexpr\f@size pt\relax}%
  \newcommand*\lineheight[1]{\fontsize{\fsize}{#1\fsize}\selectfont}%
  \ifx\svgwidth\undefined%
    \setlength{\unitlength}{244.23724219bp}%
    \ifx\svgscale\undefined%
      \relax%
    \else%
      \setlength{\unitlength}{\unitlength * \real{\svgscale}}%
    \fi%
  \else%
    \setlength{\unitlength}{\svgwidth}%
  \fi%
  \global\let\svgwidth\undefined%
  \global\let\svgscale\undefined%
  \makeatother%
  \begin{picture}(1,1.41256099)%
    \lineheight{1}%
    \setlength\tabcolsep{0pt}%
    \put(0,0){\includegraphics[width=\unitlength,page=1]{kr-shl-deployment-topology-running.pdf}}%
    \put(0.9001995,1.14540271){\makebox(0,0)[t]{\lineheight{1.25}\smash{\begin{tabular}[t]{c}Torso\end{tabular}}}}%
    \put(0.9001995,1.10241173){\makebox(0,0)[t]{\lineheight{1.25}\smash{\begin{tabular}[t]{c}phone\end{tabular}}}}%
    \put(0,0){\includegraphics[width=\unitlength,page=2]{kr-shl-deployment-topology-running.pdf}}%
    \put(0.93704891,0.83832424){\makebox(0,0)[t]{\lineheight{1.25}\smash{\begin{tabular}[t]{c}Hand\end{tabular}}}}%
    \put(0.93704891,0.79533325){\makebox(0,0)[t]{\lineheight{1.25}\smash{\begin{tabular}[t]{c}phone\end{tabular}}}}%
    \put(0,0){\includegraphics[width=\unitlength,page=3]{kr-shl-deployment-topology-running.pdf}}%
    \put(0.12329095,0.62336931){\makebox(0,0)[t]{\lineheight{1.25}\smash{\begin{tabular}[t]{c}Pocket\end{tabular}}}}%
    \put(0.12329095,0.58037832){\makebox(0,0)[t]{\lineheight{1.25}\smash{\begin{tabular}[t]{c}phone\end{tabular}}}}%
    \put(0,0){\includegraphics[width=\unitlength,page=4]{kr-shl-deployment-topology-running.pdf}}%
    \put(0.07722918,1.14540271){\makebox(0,0)[t]{\lineheight{1.25}\smash{\begin{tabular}[t]{c}Backpack\end{tabular}}}}%
    \put(0.07722918,1.10241173){\makebox(0,0)[t]{\lineheight{1.25}\smash{\begin{tabular}[t]{c}phone\end{tabular}}}}%
  \end{picture}%
\endgroup%

%% file: img/confusion_matrix-5-5-3-5-AllPositionsFused-10-folds-modelIter35.pdf_tex
%% Accompanies image file 'confusion_matrix-5-5-3-5-AllPositionsFused-10-folds-modelIter35.pdf' (pdf, eps, ps)
\begingroup%
  \makeatletter%
  \providecommand\color[2][]{%
    \errmessage{(Inkscape) Color is used for the text in Inkscape, but the package 'color.sty' is not loaded}%
    \renewcommand\color[2][]{}%
  }%
  \providecommand\transparent[1]{%
    \errmessage{(Inkscape) Transparency is used (non-zero) for the text in Inkscape, but the package 'transparent.sty' is not loaded}%
    \renewcommand\transparent[1]{}%
  }%
  \providecommand\rotatebox[2]{#2}%
  \newcommand*\fsize{\dimexpr\f@size pt\relax}%
  \newcommand*\lineheight[1]{\fontsize{\fsize}{#1\fsize}\selectfont}%
  \ifx\svgwidth\undefined%
    \setlength{\unitlength}{317.93824694bp}%
    \ifx\svgscale\undefined%
      \relax%
    \else%
      \setlength{\unitlength}{\unitlength * \real{\svgscale}}%
    \fi%
  \else%
    \setlength{\unitlength}{\svgwidth}%
  \fi%
  \global\let\svgwidth\undefined%
  \global\let\svgscale\undefined%
  \makeatother%
  \begin{picture}(1,0.86888303)%
    \lineheight{1}%
    \setlength\tabcolsep{0pt}%
    \put(0,0){\includegraphics[width=\unitlength,page=1]{confusion_matrix-5-5-3-5-AllPositionsFused-10-folds-modelIter35.pdf}}%
    \put(0.17341706,0.04377949){\rotatebox{45}{\makebox(0,0)[lt]{\lineheight{1.25}\smash{\begin{tabular}[t]{l}Null\end{tabular}}}}}%
    \put(0,0){\includegraphics[width=\unitlength,page=2]{confusion_matrix-5-5-3-5-AllPositionsFused-10-folds-modelIter35.pdf}}%
    \put(0.24874246,0.0480295){\rotatebox{45}{\makebox(0,0)[lt]{\lineheight{1.25}\smash{\begin{tabular}[t]{l}still\end{tabular}}}}}%
    \put(0,0){\includegraphics[width=\unitlength,page=3]{confusion_matrix-5-5-3-5-AllPositionsFused-10-folds-modelIter35.pdf}}%
    \put(0.3159258,0.03599535){\rotatebox{45}{\makebox(0,0)[lt]{\lineheight{1.25}\smash{\begin{tabular}[t]{l}walk\end{tabular}}}}}%
    \put(0,0){\includegraphics[width=\unitlength,page=4]{confusion_matrix-5-5-3-5-AllPositionsFused-10-folds-modelIter35.pdf}}%
    \put(0.39589735,0.04953767){\rotatebox{45}{\makebox(0,0)[lt]{\lineheight{1.25}\smash{\begin{tabular}[t]{l}run\end{tabular}}}}}%
    \put(0,0){\includegraphics[width=\unitlength,page=5]{confusion_matrix-5-5-3-5-AllPositionsFused-10-folds-modelIter35.pdf}}%
    \put(0.46433866,0.04001948){\rotatebox{45}{\makebox(0,0)[lt]{\lineheight{1.25}\smash{\begin{tabular}[t]{l}bike\end{tabular}}}}}%
    \put(0,0){\includegraphics[width=\unitlength,page=6]{confusion_matrix-5-5-3-5-AllPositionsFused-10-folds-modelIter35.pdf}}%
    \put(0.54346403,0.05186943){\rotatebox{45}{\makebox(0,0)[lt]{\lineheight{1.25}\smash{\begin{tabular}[t]{l}car\end{tabular}}}}}%
    \put(0,0){\includegraphics[width=\unitlength,page=7]{confusion_matrix-5-5-3-5-AllPositionsFused-10-folds-modelIter35.pdf}}%
    \put(0.61426494,0.04707037){\rotatebox{45}{\makebox(0,0)[lt]{\lineheight{1.25}\smash{\begin{tabular}[t]{l}bus\end{tabular}}}}}%
    \put(0,0){\includegraphics[width=\unitlength,page=8]{confusion_matrix-5-5-3-5-AllPositionsFused-10-folds-modelIter35.pdf}}%
    \put(0.68148301,0.03510573){\rotatebox{45}{\makebox(0,0)[lt]{\lineheight{1.25}\smash{\begin{tabular}[t]{l}train\end{tabular}}}}}%
    \put(0,0){\includegraphics[width=\unitlength,page=9]{confusion_matrix-5-5-3-5-AllPositionsFused-10-folds-modelIter35.pdf}}%
    \put(0.73817513,0.02567869){\rotatebox{45}{\makebox(0,0)[lt]{\lineheight{1.25}\smash{\begin{tabular}[t]{l}subway\end{tabular}}}}}%
    \put(0.48162709,0.0034448){\makebox(0,0)[t]{\lineheight{1.25}\smash{\begin{tabular}[t]{c}Predicted label\end{tabular}}}}%
    \put(0,0){\includegraphics[width=\unitlength,page=10]{confusion_matrix-5-5-3-5-AllPositionsFused-10-folds-modelIter35.pdf}}%
    \put(0.13020844,0.73604017){\makebox(0,0)[rt]{\lineheight{1.25}\smash{\begin{tabular}[t]{r}Null\end{tabular}}}}%
    \put(0,0){\includegraphics[width=\unitlength,page=11]{confusion_matrix-5-5-3-5-AllPositionsFused-10-folds-modelIter35.pdf}}%
    \put(0.13020844,0.66283977){\makebox(0,0)[rt]{\lineheight{1.25}\smash{\begin{tabular}[t]{r}still\end{tabular}}}}%
    \put(0,0){\includegraphics[width=\unitlength,page=12]{confusion_matrix-5-5-3-5-AllPositionsFused-10-folds-modelIter35.pdf}}%
    \put(0.13020844,0.58963936){\makebox(0,0)[rt]{\lineheight{1.25}\smash{\begin{tabular}[t]{r}walk\end{tabular}}}}%
    \put(0,0){\includegraphics[width=\unitlength,page=13]{confusion_matrix-5-5-3-5-AllPositionsFused-10-folds-modelIter35.pdf}}%
    \put(0.13020844,0.51643896){\makebox(0,0)[rt]{\lineheight{1.25}\smash{\begin{tabular}[t]{r}run\end{tabular}}}}%
    \put(0,0){\includegraphics[width=\unitlength,page=14]{confusion_matrix-5-5-3-5-AllPositionsFused-10-folds-modelIter35.pdf}}%
    \put(0.13020844,0.44323855){\makebox(0,0)[rt]{\lineheight{1.25}\smash{\begin{tabular}[t]{r}bike\end{tabular}}}}%
    \put(0,0){\includegraphics[width=\unitlength,page=15]{confusion_matrix-5-5-3-5-AllPositionsFused-10-folds-modelIter35.pdf}}%
    \put(0.13020844,0.37003815){\makebox(0,0)[rt]{\lineheight{1.25}\smash{\begin{tabular}[t]{r}car\end{tabular}}}}%
    \put(0,0){\includegraphics[width=\unitlength,page=16]{confusion_matrix-5-5-3-5-AllPositionsFused-10-folds-modelIter35.pdf}}%
    \put(0.13020844,0.29683774){\makebox(0,0)[rt]{\lineheight{1.25}\smash{\begin{tabular}[t]{r}bus\end{tabular}}}}%
    \put(0,0){\includegraphics[width=\unitlength,page=17]{confusion_matrix-5-5-3-5-AllPositionsFused-10-folds-modelIter35.pdf}}%
    \put(0.13020844,0.22363734){\makebox(0,0)[rt]{\lineheight{1.25}\smash{\begin{tabular}[t]{r}train\end{tabular}}}}%
    \put(0,0){\includegraphics[width=\unitlength,page=18]{confusion_matrix-5-5-3-5-AllPositionsFused-10-folds-modelIter35.pdf}}%
    \put(0.13020844,0.15043694){\makebox(0,0)[rt]{\lineheight{1.25}\smash{\begin{tabular}[t]{r}subway\end{tabular}}}}%
    \put(0.02893087,0.45518811){\rotatebox{90}{\makebox(0,0)[t]{\lineheight{1.25}\smash{\begin{tabular}[t]{c}True label\end{tabular}}}}}%
    \put(0,0){\includegraphics[width=\unitlength,page=19]{confusion_matrix-5-5-3-5-AllPositionsFused-10-folds-modelIter35.pdf}}%
    \put(0.18882547,0.74798973){\color[rgb]{1,1,1}\makebox(0,0)[t]{\lineheight{1.25}\smash{\begin{tabular}[t]{c}0.74\end{tabular}}}}%
    \put(0.26202588,0.74798973){\makebox(0,0)[t]{\lineheight{1.25}\smash{\begin{tabular}[t]{c}0.04\end{tabular}}}}%
    \put(0.33522628,0.74798973){\makebox(0,0)[t]{\lineheight{1.25}\smash{\begin{tabular}[t]{c}0.08\end{tabular}}}}%
    \put(0.40842669,0.74798973){\makebox(0,0)[t]{\lineheight{1.25}\smash{\begin{tabular}[t]{c}0.00\end{tabular}}}}%
    \put(0.48162709,0.74798973){\makebox(0,0)[t]{\lineheight{1.25}\smash{\begin{tabular}[t]{c}0.01\end{tabular}}}}%
    \put(0.55482745,0.74798973){\makebox(0,0)[t]{\lineheight{1.25}\smash{\begin{tabular}[t]{c}0.05\end{tabular}}}}%
    \put(0.6280279,0.74798973){\makebox(0,0)[t]{\lineheight{1.25}\smash{\begin{tabular}[t]{c}0.03\end{tabular}}}}%
    \put(0.70122826,0.74798973){\makebox(0,0)[t]{\lineheight{1.25}\smash{\begin{tabular}[t]{c}0.01\end{tabular}}}}%
    \put(0.77442871,0.74798973){\makebox(0,0)[t]{\lineheight{1.25}\smash{\begin{tabular}[t]{c}0.04\end{tabular}}}}%
    \put(0.18882547,0.67478933){\makebox(0,0)[t]{\lineheight{1.25}\smash{\begin{tabular}[t]{c}0.14\end{tabular}}}}%
    \put(0.26202588,0.67478933){\makebox(0,0)[t]{\lineheight{1.25}\smash{\begin{tabular}[t]{c}0.41\end{tabular}}}}%
    \put(0.33522628,0.67478933){\makebox(0,0)[t]{\lineheight{1.25}\smash{\begin{tabular}[t]{c}0.03\end{tabular}}}}%
    \put(0.40842669,0.67478933){\makebox(0,0)[t]{\lineheight{1.25}\smash{\begin{tabular}[t]{c}0.00\end{tabular}}}}%
    \put(0.48162709,0.67478933){\makebox(0,0)[t]{\lineheight{1.25}\smash{\begin{tabular}[t]{c}0.04\end{tabular}}}}%
    \put(0.55482745,0.67478933){\makebox(0,0)[t]{\lineheight{1.25}\smash{\begin{tabular}[t]{c}0.05\end{tabular}}}}%
    \put(0.6280279,0.67478933){\makebox(0,0)[t]{\lineheight{1.25}\smash{\begin{tabular}[t]{c}0.11\end{tabular}}}}%
    \put(0.70122826,0.67478933){\makebox(0,0)[t]{\lineheight{1.25}\smash{\begin{tabular}[t]{c}0.10\end{tabular}}}}%
    \put(0.77442871,0.67478933){\makebox(0,0)[t]{\lineheight{1.25}\smash{\begin{tabular}[t]{c}0.12\end{tabular}}}}%
    \put(0.18882547,0.60158891){\makebox(0,0)[t]{\lineheight{1.25}\smash{\begin{tabular}[t]{c}0.12\end{tabular}}}}%
    \put(0.26202588,0.60158891){\makebox(0,0)[t]{\lineheight{1.25}\smash{\begin{tabular}[t]{c}0.01\end{tabular}}}}%
    \put(0.33522628,0.60158891){\color[rgb]{1,1,1}\makebox(0,0)[t]{\lineheight{1.25}\smash{\begin{tabular}[t]{c}0.77\end{tabular}}}}%
    \put(0.40842669,0.60158891){\makebox(0,0)[t]{\lineheight{1.25}\smash{\begin{tabular}[t]{c}0.00\end{tabular}}}}%
    \put(0.48162709,0.60158891){\makebox(0,0)[t]{\lineheight{1.25}\smash{\begin{tabular}[t]{c}0.04\end{tabular}}}}%
    \put(0.55482745,0.60158891){\makebox(0,0)[t]{\lineheight{1.25}\smash{\begin{tabular}[t]{c}0.00\end{tabular}}}}%
    \put(0.6280279,0.60158891){\makebox(0,0)[t]{\lineheight{1.25}\smash{\begin{tabular}[t]{c}0.03\end{tabular}}}}%
    \put(0.70122826,0.60158891){\makebox(0,0)[t]{\lineheight{1.25}\smash{\begin{tabular}[t]{c}0.01\end{tabular}}}}%
    \put(0.77442871,0.60158891){\makebox(0,0)[t]{\lineheight{1.25}\smash{\begin{tabular}[t]{c}0.02\end{tabular}}}}%
    \put(0.18882547,0.5283885){\makebox(0,0)[t]{\lineheight{1.25}\smash{\begin{tabular}[t]{c}0.00\end{tabular}}}}%
    \put(0.26202588,0.5283885){\makebox(0,0)[t]{\lineheight{1.25}\smash{\begin{tabular}[t]{c}0.00\end{tabular}}}}%
    \put(0.33522628,0.5283885){\makebox(0,0)[t]{\lineheight{1.25}\smash{\begin{tabular}[t]{c}0.04\end{tabular}}}}%
    \put(0.40842669,0.5283885){\color[rgb]{1,1,1}\makebox(0,0)[t]{\lineheight{1.25}\smash{\begin{tabular}[t]{c}0.96\end{tabular}}}}%
    \put(0.48162709,0.5283885){\makebox(0,0)[t]{\lineheight{1.25}\smash{\begin{tabular}[t]{c}0.00\end{tabular}}}}%
    \put(0.55482745,0.5283885){\makebox(0,0)[t]{\lineheight{1.25}\smash{\begin{tabular}[t]{c}0.00\end{tabular}}}}%
    \put(0.6280279,0.5283885){\makebox(0,0)[t]{\lineheight{1.25}\smash{\begin{tabular}[t]{c}0.00\end{tabular}}}}%
    \put(0.70122826,0.5283885){\makebox(0,0)[t]{\lineheight{1.25}\smash{\begin{tabular}[t]{c}0.00\end{tabular}}}}%
    \put(0.77442871,0.5283885){\makebox(0,0)[t]{\lineheight{1.25}\smash{\begin{tabular}[t]{c}0.00\end{tabular}}}}%
    \put(0.18882547,0.45518812){\makebox(0,0)[t]{\lineheight{1.25}\smash{\begin{tabular}[t]{c}0.19\end{tabular}}}}%
    \put(0.26202588,0.45518812){\makebox(0,0)[t]{\lineheight{1.25}\smash{\begin{tabular}[t]{c}0.01\end{tabular}}}}%
    \put(0.33522628,0.45518812){\makebox(0,0)[t]{\lineheight{1.25}\smash{\begin{tabular}[t]{c}0.02\end{tabular}}}}%
    \put(0.40842669,0.45518812){\makebox(0,0)[t]{\lineheight{1.25}\smash{\begin{tabular}[t]{c}0.00\end{tabular}}}}%
    \put(0.48162709,0.45518812){\color[rgb]{1,1,1}\makebox(0,0)[t]{\lineheight{1.25}\smash{\begin{tabular}[t]{c}0.77\end{tabular}}}}%
    \put(0.55482745,0.45518812){\makebox(0,0)[t]{\lineheight{1.25}\smash{\begin{tabular}[t]{c}0.00\end{tabular}}}}%
    \put(0.6280279,0.45518812){\makebox(0,0)[t]{\lineheight{1.25}\smash{\begin{tabular}[t]{c}0.00\end{tabular}}}}%
    \put(0.70122826,0.45518812){\makebox(0,0)[t]{\lineheight{1.25}\smash{\begin{tabular}[t]{c}0.01\end{tabular}}}}%
    \put(0.77442871,0.45518812){\makebox(0,0)[t]{\lineheight{1.25}\smash{\begin{tabular}[t]{c}0.01\end{tabular}}}}%
    \put(0.18882547,0.38198772){\makebox(0,0)[t]{\lineheight{1.25}\smash{\begin{tabular}[t]{c}0.06\end{tabular}}}}%
    \put(0.26202588,0.38198772){\makebox(0,0)[t]{\lineheight{1.25}\smash{\begin{tabular}[t]{c}0.00\end{tabular}}}}%
    \put(0.33522628,0.38198772){\makebox(0,0)[t]{\lineheight{1.25}\smash{\begin{tabular}[t]{c}0.00\end{tabular}}}}%
    \put(0.40842669,0.38198772){\makebox(0,0)[t]{\lineheight{1.25}\smash{\begin{tabular}[t]{c}0.00\end{tabular}}}}%
    \put(0.48162709,0.38198772){\makebox(0,0)[t]{\lineheight{1.25}\smash{\begin{tabular}[t]{c}0.00\end{tabular}}}}%
    \put(0.55482745,0.38198772){\color[rgb]{1,1,1}\makebox(0,0)[t]{\lineheight{1.25}\smash{\begin{tabular}[t]{c}0.87\end{tabular}}}}%
    \put(0.6280279,0.38198772){\makebox(0,0)[t]{\lineheight{1.25}\smash{\begin{tabular}[t]{c}0.06\end{tabular}}}}%
    \put(0.70122826,0.38198772){\makebox(0,0)[t]{\lineheight{1.25}\smash{\begin{tabular}[t]{c}0.00\end{tabular}}}}%
    \put(0.77442871,0.38198772){\makebox(0,0)[t]{\lineheight{1.25}\smash{\begin{tabular}[t]{c}0.00\end{tabular}}}}%
    \put(0.18882547,0.30878731){\makebox(0,0)[t]{\lineheight{1.25}\smash{\begin{tabular}[t]{c}0.11\end{tabular}}}}%
    \put(0.26202588,0.30878731){\makebox(0,0)[t]{\lineheight{1.25}\smash{\begin{tabular}[t]{c}0.05\end{tabular}}}}%
    \put(0.33522628,0.30878731){\makebox(0,0)[t]{\lineheight{1.25}\smash{\begin{tabular}[t]{c}0.02\end{tabular}}}}%
    \put(0.40842669,0.30878731){\makebox(0,0)[t]{\lineheight{1.25}\smash{\begin{tabular}[t]{c}0.00\end{tabular}}}}%
    \put(0.48162709,0.30878731){\makebox(0,0)[t]{\lineheight{1.25}\smash{\begin{tabular}[t]{c}0.01\end{tabular}}}}%
    \put(0.55482745,0.30878731){\makebox(0,0)[t]{\lineheight{1.25}\smash{\begin{tabular}[t]{c}0.03\end{tabular}}}}%
    \put(0.6280279,0.30878731){\color[rgb]{1,1,1}\makebox(0,0)[t]{\lineheight{1.25}\smash{\begin{tabular}[t]{c}0.69\end{tabular}}}}%
    \put(0.70122826,0.30878731){\makebox(0,0)[t]{\lineheight{1.25}\smash{\begin{tabular}[t]{c}0.00\end{tabular}}}}%
    \put(0.77442871,0.30878731){\makebox(0,0)[t]{\lineheight{1.25}\smash{\begin{tabular}[t]{c}0.09\end{tabular}}}}%
    \put(0.18882547,0.23558691){\makebox(0,0)[t]{\lineheight{1.25}\smash{\begin{tabular}[t]{c}0.04\end{tabular}}}}%
    \put(0.26202588,0.23558691){\makebox(0,0)[t]{\lineheight{1.25}\smash{\begin{tabular}[t]{c}0.04\end{tabular}}}}%
    \put(0.33522628,0.23558691){\makebox(0,0)[t]{\lineheight{1.25}\smash{\begin{tabular}[t]{c}0.01\end{tabular}}}}%
    \put(0.40842669,0.23558691){\makebox(0,0)[t]{\lineheight{1.25}\smash{\begin{tabular}[t]{c}0.00\end{tabular}}}}%
    \put(0.48162709,0.23558691){\makebox(0,0)[t]{\lineheight{1.25}\smash{\begin{tabular}[t]{c}0.01\end{tabular}}}}%
    \put(0.55482745,0.23558691){\makebox(0,0)[t]{\lineheight{1.25}\smash{\begin{tabular}[t]{c}0.01\end{tabular}}}}%
    \put(0.6280279,0.23558691){\makebox(0,0)[t]{\lineheight{1.25}\smash{\begin{tabular}[t]{c}0.03\end{tabular}}}}%
    \put(0.70122826,0.23558691){\color[rgb]{1,1,1}\makebox(0,0)[t]{\lineheight{1.25}\smash{\begin{tabular}[t]{c}0.60\end{tabular}}}}%
    \put(0.77442871,0.23558691){\makebox(0,0)[t]{\lineheight{1.25}\smash{\begin{tabular}[t]{c}0.27\end{tabular}}}}%
    \put(0.18882547,0.1623865){\makebox(0,0)[t]{\lineheight{1.25}\smash{\begin{tabular}[t]{c}0.11\end{tabular}}}}%
    \put(0.26202588,0.1623865){\makebox(0,0)[t]{\lineheight{1.25}\smash{\begin{tabular}[t]{c}0.02\end{tabular}}}}%
    \put(0.33522628,0.1623865){\makebox(0,0)[t]{\lineheight{1.25}\smash{\begin{tabular}[t]{c}0.01\end{tabular}}}}%
    \put(0.40842669,0.1623865){\makebox(0,0)[t]{\lineheight{1.25}\smash{\begin{tabular}[t]{c}0.00\end{tabular}}}}%
    \put(0.48162709,0.1623865){\makebox(0,0)[t]{\lineheight{1.25}\smash{\begin{tabular}[t]{c}0.00\end{tabular}}}}%
    \put(0.55482745,0.1623865){\makebox(0,0)[t]{\lineheight{1.25}\smash{\begin{tabular}[t]{c}0.00\end{tabular}}}}%
    \put(0.6280279,0.1623865){\makebox(0,0)[t]{\lineheight{1.25}\smash{\begin{tabular}[t]{c}0.00\end{tabular}}}}%
    \put(0.70122826,0.1623865){\makebox(0,0)[t]{\lineheight{1.25}\smash{\begin{tabular}[t]{c}0.05\end{tabular}}}}%
    \put(0.77442871,0.1623865){\color[rgb]{1,1,1}\makebox(0,0)[t]{\lineheight{1.25}\smash{\begin{tabular}[t]{c}0.81\end{tabular}}}}%
    \put(0.48162709,0.80346152){\makebox(0,0)[t]{\lineheight{1.25}\smash{\begin{tabular}[t]{c}Confusion matrix\end{tabular}}}}%
    \put(0,0){\includegraphics[width=\unitlength,page=20]{confusion_matrix-5-5-3-5-AllPositionsFused-10-folds-modelIter35.pdf}}%
    \put(0.93052678,0.11383676){\makebox(0,0)[lt]{\lineheight{1.25}\smash{\begin{tabular}[t]{l}0.0\end{tabular}}}}%
    \put(0,0){\includegraphics[width=\unitlength,page=21]{confusion_matrix-5-5-3-5-AllPositionsFused-10-folds-modelIter35.pdf}}%
    \put(0.93052678,0.25167875){\makebox(0,0)[lt]{\lineheight{1.25}\smash{\begin{tabular}[t]{l}0.2\end{tabular}}}}%
    \put(0,0){\includegraphics[width=\unitlength,page=22]{confusion_matrix-5-5-3-5-AllPositionsFused-10-folds-modelIter35.pdf}}%
    \put(0.93052678,0.38952074){\makebox(0,0)[lt]{\lineheight{1.25}\smash{\begin{tabular}[t]{l}0.4\end{tabular}}}}%
    \put(0,0){\includegraphics[width=\unitlength,page=23]{confusion_matrix-5-5-3-5-AllPositionsFused-10-folds-modelIter35.pdf}}%
    \put(0.93052678,0.52736273){\makebox(0,0)[lt]{\lineheight{1.25}\smash{\begin{tabular}[t]{l}0.6\end{tabular}}}}%
    \put(0,0){\includegraphics[width=\unitlength,page=24]{confusion_matrix-5-5-3-5-AllPositionsFused-10-folds-modelIter35.pdf}}%
    \put(0.93052678,0.66520472){\makebox(0,0)[lt]{\lineheight{1.25}\smash{\begin{tabular}[t]{l}0.8\end{tabular}}}}%
    \put(0,0){\includegraphics[width=\unitlength,page=25]{confusion_matrix-5-5-3-5-AllPositionsFused-10-folds-modelIter35.pdf}}%
  \end{picture}%
\endgroup%

%% file: img/associated-fold-for-each-frame-25vs1.pdf_tex
%% Creator: Inkscape inkscape 0.48.3.1, www.inkscape.org
%% Accompanies image file 'associated-fold-for-each-frame-25vs1.pdf' (pdf, eps, ps)
\begingroup%
  \makeatletter%
  \providecommand\color[2][]{%
    \errmessage{(Inkscape) Color is used for the text in Inkscape, but the package 'color.sty' is not loaded}%
    \renewcommand\color[2][]{}%
  }%
  \providecommand\transparent[1]{%
    \errmessage{(Inkscape) Transparency is used (non-zero) for the text in Inkscape, but the package 'transparent.sty' is not loaded}%
    \renewcommand\transparent[1]{}%
  }%
  \providecommand\rotatebox[2]{#2}%
  \ifx\svgwidth\undefined%
    \setlength{\unitlength}{1348.8bp}%
    \ifx\svgscale\undefined%
      \relax%
    \else%
      \setlength{\unitlength}{\unitlength * \real{\svgscale}}%
    \fi%
  \else%
    \setlength{\unitlength}{\svgwidth}%
  \fi%
  \global\let\svgwidth\undefined%
  \global\let\svgscale\undefined%
  \makeatother%
  \begin{picture}(1,0.12455516)%
    \put(0.00100000,0.01300000){\includegraphics[width=\unitlength]{associated-fold-for-each-frame-25vs1.pdf}}%
	\put(0.10000000,0.01000000){\rotatebox{90}{\makebox(0,0)[lb]{\smash{meta-segment length}}}}%
    \put(0.11015634,0.04275605){\makebox(0,0)[lb]{\smash{20}}}%
    \put(0.12015634,0.08775605){\makebox(0,0)[lb]{\smash{1}}}%
    \put(0.50000000,0.00000614){\makebox(0,0)[lb]{\smash{frames}}}%
    \put(0.20015634,0.01275605){\makebox(0,0)[lb]{\smash{4200}}}%
    \put(0.30015634,0.01275605){\makebox(0,0)[lb]{\smash{4400}}}%
    \put(0.39015634,0.01275605){\makebox(0,0)[lb]{\smash{4600}}}%
    \put(0.48015634,0.01275605){\makebox(0,0)[lb]{\smash{4800}}}%
    \put(0.58015634,0.01275605){\makebox(0,0)[lb]{\smash{5000}}}%
    \put(0.67015634,0.01275605){\makebox(0,0)[lb]{\smash{5200}}}%
    \put(0.76515634,0.01275605){\makebox(0,0)[lb]{\smash{5400}}}%
    \put(0.86015634,0.01275605){\makebox(0,0)[lb]{\smash{5600}}}%

  \end{picture}%
\endgroup%

%% file: img/pitfalls-relative-bias-5-5-3-2-bis.pdf_tex
%% Accompanies image file 'pitfalls-relative-bias-5-5-3-2-bis.pdf' (pdf, eps, ps)
\begingroup%
  \makeatletter%
  \providecommand\color[2][]{%
    \errmessage{(Inkscape) Color is used for the text in Inkscape, but the package 'color.sty' is not loaded}%
    \renewcommand\color[2][]{}%
  }%
  \providecommand\transparent[1]{%
    \errmessage{(Inkscape) Transparency is used (non-zero) for the text in Inkscape, but the package 'transparent.sty' is not loaded}%
    \renewcommand\transparent[1]{}%
  }%
  \providecommand\rotatebox[2]{#2}%
  \newcommand*\fsize{\dimexpr\f@size pt\relax}%
  \newcommand*\lineheight[1]{\fontsize{\fsize}{#1\fsize}\selectfont}%
  \ifx\svgwidth\undefined%
    \setlength{\unitlength}{720.00000464bp}%
    \ifx\svgscale\undefined%
      \relax%
    \else%
      \setlength{\unitlength}{\unitlength * \real{\svgscale}}%
    \fi%
  \else%
    \setlength{\unitlength}{\svgwidth}%
  \fi%
  \global\let\svgwidth\undefined%
  \global\let\svgscale\undefined%
  \makeatother%
  \begin{picture}(1,0.29153482)%
    \lineheight{1}%
    \setlength\tabcolsep{0pt}%
    \put(0,0){\includegraphics[width=\unitlength,page=1]{pitfalls-relative-bias-5-5-3-2-bis.pdf}}%
    \put(0.29017997,0.03028699){\makebox(0,0)[t]{\lineheight{1.25}\smash{\begin{tabular}[t]{c}20\end{tabular}}}}%
    \put(0,0){\includegraphics[width=\unitlength,page=2]{pitfalls-relative-bias-5-5-3-2-bis.pdf}}%
    \put(0.45865334,0.03028699){\makebox(0,0)[t]{\lineheight{1.25}\smash{\begin{tabular}[t]{c}40\end{tabular}}}}%
    \put(0,0){\includegraphics[width=\unitlength,page=3]{pitfalls-relative-bias-5-5-3-2-bis.pdf}}%
    \put(0.62712669,0.03028699){\makebox(0,0)[t]{\lineheight{1.25}\smash{\begin{tabular}[t]{c}60\end{tabular}}}}%
    \put(0,0){\includegraphics[width=\unitlength,page=4]{pitfalls-relative-bias-5-5-3-2-bis.pdf}}%
    \put(0.79560004,0.03028699){\makebox(0,0)[t]{\lineheight{1.25}\smash{\begin{tabular}[t]{c}80\end{tabular}}}}%
    \put(0.5125,0.00503959){\makebox(0,0)[t]{\lineheight{1.25}\smash{\begin{tabular}[t]{c}f1-score\end{tabular}}}}%
    \put(0,0){\includegraphics[width=\unitlength,page=5]{pitfalls-relative-bias-5-5-3-2-bis.pdf}}%
    \put(0.12013889,0.11096231){\makebox(0,0)[rt]{\lineheight{1.25}\smash{\begin{tabular}[t]{r}−1\end{tabular}}}}%
    \put(0.90486111,0.11096231){\makebox(0,0)[lt]{\lineheight{1.25}\smash{\begin{tabular}[t]{l}−1\end{tabular}}}}%
    \put(0,0){\includegraphics[width=\unitlength,page=6]{pitfalls-relative-bias-5-5-3-2-bis.pdf}}%
    \put(0.12013889,0.15122708){\makebox(0,0)[rt]{\lineheight{1.25}\smash{\begin{tabular}[t]{r}0\end{tabular}}}}%
    \put(0.90486111,0.15122708){\makebox(0,0)[lt]{\lineheight{1.25}\smash{\begin{tabular}[t]{l}0\end{tabular}}}}%
    \put(0,0){\includegraphics[width=\unitlength,page=7]{pitfalls-relative-bias-5-5-3-2-bis.pdf}}%
    \put(0.12013889,0.19149184){\makebox(0,0)[rt]{\lineheight{1.25}\smash{\begin{tabular}[t]{r}1\end{tabular}}}}%
    \put(0.90486111,0.19149184){\makebox(0,0)[lt]{\lineheight{1.25}\smash{\begin{tabular}[t]{l}1\end{tabular}}}}%
    \put(0,0){\includegraphics[width=\unitlength,page=8]{pitfalls-relative-bias-5-5-3-2-bis.pdf}}%
    \put(0.12013889,0.2317566){\makebox(0,0)[rt]{\lineheight{1.25}\smash{\begin{tabular}[t]{r}2\end{tabular}}}}%
    \put(0.90486111,0.2317566){\makebox(0,0)[lt]{\lineheight{1.25}\smash{\begin{tabular}[t]{l}2\end{tabular}}}}%
    \put(0.09121962,0.15453482){\rotatebox{90}{\makebox(0,0)[t]{\lineheight{1.25}\smash{\begin{tabular}[t]{c}Relative Bias (\%)\end{tabular}}}}}%
    \put(0,0){\includegraphics[width=\unitlength,page=9]{pitfalls-relative-bias-5-5-3-2-bis.pdf}}%
    \put(0.12013889,0.07037544){\makebox(0,0)[rt]{\lineheight{1.25}\smash{\begin{tabular}[t]{r}−2\end{tabular}}}}%
    \put(0.90486111,0.07037544){\makebox(0,0)[lt]{\lineheight{1.25}\smash{\begin{tabular}[t]{l}−2\end{tabular}}}}%
    \put(0,0){\includegraphics[width=\unitlength,page=10]{pitfalls-relative-bias-5-5-3-2-bis.pdf}}%
    \put(0.5,0.27278742){\makebox(0,0)[t]{\lineheight{1.25}\smash{\begin{tabular}[t]{c}10-Folds Cross-Validation (Null-Class discarded)\end{tabular}}}}%
  \end{picture}%
\endgroup%

%% file: img/pitfalls-relative-bias-5-5-3-4-bis.pdf_tex
%% Accompanies image file 'pitfalls-relative-bias-5-5-3-4-bis.pdf' (pdf, eps, ps)
\begingroup%
  \makeatletter%
  \providecommand\color[2][]{%
    \errmessage{(Inkscape) Color is used for the text in Inkscape, but the package 'color.sty' is not loaded}%
    \renewcommand\color[2][]{}%
  }%
  \providecommand\transparent[1]{%
    \errmessage{(Inkscape) Transparency is used (non-zero) for the text in Inkscape, but the package 'transparent.sty' is not loaded}%
    \renewcommand\transparent[1]{}%
  }%
  \providecommand\rotatebox[2]{#2}%
  \newcommand*\fsize{\dimexpr\f@size pt\relax}%
  \newcommand*\lineheight[1]{\fontsize{\fsize}{#1\fsize}\selectfont}%
  \ifx\svgwidth\undefined%
    \setlength{\unitlength}{719.99998304bp}%
    \ifx\svgscale\undefined%
      \relax%
    \else%
      \setlength{\unitlength}{\unitlength * \real{\svgscale}}%
    \fi%
  \else%
    \setlength{\unitlength}{\svgwidth}%
  \fi%
  \global\let\svgwidth\undefined%
  \global\let\svgscale\undefined%
  \makeatother%
  \begin{picture}(1,0.29153483)%
    \lineheight{1}%
    \setlength\tabcolsep{0pt}%
    \put(0,0){\includegraphics[width=\unitlength,page=1]{pitfalls-relative-bias-5-5-3-4-bis.pdf}}%
    \put(0.29017998,0.030287){\makebox(0,0)[t]{\lineheight{1.25}\smash{\begin{tabular}[t]{c}20\end{tabular}}}}%
    \put(0,0){\includegraphics[width=\unitlength,page=2]{pitfalls-relative-bias-5-5-3-4-bis.pdf}}%
    \put(0.45865336,0.030287){\makebox(0,0)[t]{\lineheight{1.25}\smash{\begin{tabular}[t]{c}40\end{tabular}}}}%
    \put(0,0){\includegraphics[width=\unitlength,page=3]{pitfalls-relative-bias-5-5-3-4-bis.pdf}}%
    \put(0.62712671,0.030287){\makebox(0,0)[t]{\lineheight{1.25}\smash{\begin{tabular}[t]{c}60\end{tabular}}}}%
    \put(0,0){\includegraphics[width=\unitlength,page=4]{pitfalls-relative-bias-5-5-3-4-bis.pdf}}%
    \put(0.79560007,0.030287){\makebox(0,0)[t]{\lineheight{1.25}\smash{\begin{tabular}[t]{c}80\end{tabular}}}}%
    \put(0.51250002,0.00503959){\makebox(0,0)[t]{\lineheight{1.25}\smash{\begin{tabular}[t]{c}f1-score\end{tabular}}}}%
    \put(0,0){\includegraphics[width=\unitlength,page=5]{pitfalls-relative-bias-5-5-3-4-bis.pdf}}%
    \put(0.12013889,0.11096231){\makebox(0,0)[rt]{\lineheight{1.25}\smash{\begin{tabular}[t]{r}−1\end{tabular}}}}%
    \put(0.90486114,0.11096231){\makebox(0,0)[lt]{\lineheight{1.25}\smash{\begin{tabular}[t]{l}−1\end{tabular}}}}%
    \put(0,0){\includegraphics[width=\unitlength,page=6]{pitfalls-relative-bias-5-5-3-4-bis.pdf}}%
    \put(0.12013889,0.15122708){\makebox(0,0)[rt]{\lineheight{1.25}\smash{\begin{tabular}[t]{r}0\end{tabular}}}}%
    \put(0.90486114,0.15122708){\makebox(0,0)[lt]{\lineheight{1.25}\smash{\begin{tabular}[t]{l}0\end{tabular}}}}%
    \put(0,0){\includegraphics[width=\unitlength,page=7]{pitfalls-relative-bias-5-5-3-4-bis.pdf}}%
    \put(0.12013889,0.19149184){\makebox(0,0)[rt]{\lineheight{1.25}\smash{\begin{tabular}[t]{r}1\end{tabular}}}}%
    \put(0.90486114,0.19149184){\makebox(0,0)[lt]{\lineheight{1.25}\smash{\begin{tabular}[t]{l}1\end{tabular}}}}%
    \put(0,0){\includegraphics[width=\unitlength,page=8]{pitfalls-relative-bias-5-5-3-4-bis.pdf}}%
    \put(0.12013889,0.23175661){\makebox(0,0)[rt]{\lineheight{1.25}\smash{\begin{tabular}[t]{r}2\end{tabular}}}}%
    \put(0.90486114,0.23175661){\makebox(0,0)[lt]{\lineheight{1.25}\smash{\begin{tabular}[t]{l}2\end{tabular}}}}%
    \put(0.09121962,0.15453482){\rotatebox{90}{\makebox(0,0)[t]{\lineheight{1.25}\smash{\begin{tabular}[t]{c}Relative Bias (\%)\end{tabular}}}}}%
    \put(0,0){\includegraphics[width=\unitlength,page=9]{pitfalls-relative-bias-5-5-3-4-bis.pdf}}%
    \put(0.12013889,0.07037544){\makebox(0,0)[rt]{\lineheight{1.25}\smash{\begin{tabular}[t]{r}−2\end{tabular}}}}%
    \put(0.90486114,0.07037544){\makebox(0,0)[lt]{\lineheight{1.25}\smash{\begin{tabular}[t]{l}−2\end{tabular}}}}%
    \put(0,0){\includegraphics[width=\unitlength,page=10]{pitfalls-relative-bias-5-5-3-4-bis.pdf}}%
    \put(0.50000002,0.27463131){\makebox(0,0)[t]{\lineheight{1.25}\smash{\begin{tabular}[t]{c}10-Folds Cross-Validation\end{tabular}}}}%
  \end{picture}%
\endgroup%

%% file: img/pitfalls-relative-bias-5-5-3-5-bis.pdf_tex
%% Accompanies image file 'pitfalls-relative-bias-5-5-3-5-bis.pdf' (pdf, eps, ps)
\begingroup%
  \makeatletter%
  \providecommand\color[2][]{%
    \errmessage{(Inkscape) Color is used for the text in Inkscape, but the package 'color.sty' is not loaded}%
    \renewcommand\color[2][]{}%
  }%
  \providecommand\transparent[1]{%
    \errmessage{(Inkscape) Transparency is used (non-zero) for the text in Inkscape, but the package 'transparent.sty' is not loaded}%
    \renewcommand\transparent[1]{}%
  }%
  \providecommand\rotatebox[2]{#2}%
  \newcommand*\fsize{\dimexpr\f@size pt\relax}%
  \newcommand*\lineheight[1]{\fontsize{\fsize}{#1\fsize}\selectfont}%
  \ifx\svgwidth\undefined%
    \setlength{\unitlength}{719.99998519bp}%
    \ifx\svgscale\undefined%
      \relax%
    \else%
      \setlength{\unitlength}{\unitlength * \real{\svgscale}}%
    \fi%
  \else%
    \setlength{\unitlength}{\svgwidth}%
  \fi%
  \global\let\svgwidth\undefined%
  \global\let\svgscale\undefined%
  \makeatother%
  \begin{picture}(1,0.29153483)%
    \lineheight{1}%
    \setlength\tabcolsep{0pt}%
    \put(0,0){\includegraphics[width=\unitlength,page=1]{pitfalls-relative-bias-5-5-3-5-bis.pdf}}%
    \put(0.29017998,0.03028699){\makebox(0,0)[t]{\lineheight{1.25}\smash{\begin{tabular}[t]{c}20\end{tabular}}}}%
    \put(0,0){\includegraphics[width=\unitlength,page=2]{pitfalls-relative-bias-5-5-3-5-bis.pdf}}%
    \put(0.45865336,0.03028699){\makebox(0,0)[t]{\lineheight{1.25}\smash{\begin{tabular}[t]{c}40\end{tabular}}}}%
    \put(0,0){\includegraphics[width=\unitlength,page=3]{pitfalls-relative-bias-5-5-3-5-bis.pdf}}%
    \put(0.62712671,0.03028699){\makebox(0,0)[t]{\lineheight{1.25}\smash{\begin{tabular}[t]{c}60\end{tabular}}}}%
    \put(0,0){\includegraphics[width=\unitlength,page=4]{pitfalls-relative-bias-5-5-3-5-bis.pdf}}%
    \put(0.79560007,0.03028699){\makebox(0,0)[t]{\lineheight{1.25}\smash{\begin{tabular}[t]{c}80\end{tabular}}}}%
    \put(0.51250002,0.00503959){\makebox(0,0)[t]{\lineheight{1.25}\smash{\begin{tabular}[t]{c}f1-score\end{tabular}}}}%
    \put(0,0){\includegraphics[width=\unitlength,page=5]{pitfalls-relative-bias-5-5-3-5-bis.pdf}}%
    \put(0.12013889,0.11096231){\makebox(0,0)[rt]{\lineheight{1.25}\smash{\begin{tabular}[t]{r}−1\end{tabular}}}}%
    \put(0.90486114,0.11096231){\makebox(0,0)[lt]{\lineheight{1.25}\smash{\begin{tabular}[t]{l}−1\end{tabular}}}}%
    \put(0,0){\includegraphics[width=\unitlength,page=6]{pitfalls-relative-bias-5-5-3-5-bis.pdf}}%
    \put(0.12013889,0.15122708){\makebox(0,0)[rt]{\lineheight{1.25}\smash{\begin{tabular}[t]{r}0\end{tabular}}}}%
    \put(0.90486114,0.15122708){\makebox(0,0)[lt]{\lineheight{1.25}\smash{\begin{tabular}[t]{l}0\end{tabular}}}}%
    \put(0,0){\includegraphics[width=\unitlength,page=7]{pitfalls-relative-bias-5-5-3-5-bis.pdf}}%
    \put(0.12013889,0.19149184){\makebox(0,0)[rt]{\lineheight{1.25}\smash{\begin{tabular}[t]{r}1\end{tabular}}}}%
    \put(0.90486114,0.19149184){\makebox(0,0)[lt]{\lineheight{1.25}\smash{\begin{tabular}[t]{l}1\end{tabular}}}}%
    \put(0,0){\includegraphics[width=\unitlength,page=8]{pitfalls-relative-bias-5-5-3-5-bis.pdf}}%
    \put(0.12013889,0.23175661){\makebox(0,0)[rt]{\lineheight{1.25}\smash{\begin{tabular}[t]{r}2\end{tabular}}}}%
    \put(0.90486114,0.23175661){\makebox(0,0)[lt]{\lineheight{1.25}\smash{\begin{tabular}[t]{l}2\end{tabular}}}}%
    \put(0.09121962,0.15453482){\rotatebox{90}{\makebox(0,0)[t]{\lineheight{1.25}\smash{\begin{tabular}[t]{c}Relative Bias (\%)\end{tabular}}}}}%
    \put(0,0){\includegraphics[width=\unitlength,page=9]{pitfalls-relative-bias-5-5-3-5-bis.pdf}}%
    \put(0.12013889,0.07037544){\makebox(0,0)[rt]{\lineheight{1.25}\smash{\begin{tabular}[t]{r}−2\end{tabular}}}}%
    \put(0.90486114,0.07037544){\makebox(0,0)[lt]{\lineheight{1.25}\smash{\begin{tabular}[t]{l}−2\end{tabular}}}}%
    \put(0,0){\includegraphics[width=\unitlength,page=10]{pitfalls-relative-bias-5-5-3-5-bis.pdf}}%
    \put(0.50000002,0.27671464){\makebox(0,0)[t]{\lineheight{1.25}\smash{\begin{tabular}[t]{c}5-Folds Meta-Segmented Cross-Validation\end{tabular}}}}%
  \end{picture}%
\endgroup%

%% file: img/conv-kernel_size-2-and-3.pdf_tex
%% Creator: Inkscape inkscape 0.48.5, www.inkscape.org
%% Accompanies image file 'conv-kernel_size-2-and-3.pdf' (pdf, eps, ps)
\begingroup%
  \makeatletter%
  \providecommand\color[2][]{%
    \errmessage{(Inkscape) Color is used for the text in Inkscape, but the package 'color.sty' is not loaded}%
    \renewcommand\color[2][]{}%
  }%
  \providecommand\transparent[1]{%
    \errmessage{(Inkscape) Transparency is used (non-zero) for the text in Inkscape, but the package 'transparent.sty' is not loaded}%
    \renewcommand\transparent[1]{}%
  }%
  \providecommand\rotatebox[2]{#2}%
  \ifx\svgwidth\undefined%
    \setlength{\unitlength}{1155.59998597bp}%
    \ifx\svgscale\undefined%
      \relax%
    \else%
      \setlength{\unitlength}{\unitlength * \real{\svgscale}}%
    \fi%
  \else%
    \setlength{\unitlength}{\svgwidth}%
  \fi%
  \global\let\svgwidth\undefined%
  \global\let\svgscale\undefined%
  \makeatother%
  \begin{picture}(1,0.69595016)%
    \put(0,0){\includegraphics[width=\unitlength]{conv-kernel_size-2-and-3.pdf}}%
    \put(0.10459893,0.1559332){\rotatebox{-55.319779}{\makebox(0,0)[lb]{\smash{kernel size 3}}}}%
    \put(0.11124379,0.23204399){\makebox(0,0)[b]{\smash{9.0}}}%
    \put(0.14180827,0.2020512){\makebox(0,0)[b]{\smash{9.5}}}%
    \put(0.16283851,0.17137918){\makebox(0,0)[b]{\smash{10.0}}}%
    \put(0.18435049,0.14000449){\makebox(0,0)[b]{\smash{10.5}}}%
    \put(0.20636099,0.10790278){\makebox(0,0)[b]{\smash{11.0}}}%
    \put(0.22888751,0.07504843){\makebox(0,0)[b]{\smash{11.5}}}%
    \put(0.25194843,0.04141471){\makebox(0,0)[b]{\smash{12.0}}}%
    \put(0.47949368,0.04213951){\rotatebox{12.412994}{\makebox(0,0)[lb]{\smash{kernel size 2}}}}%
    \put(0.31496171,0.02893636){\makebox(0,0)[b]{\smash{9}}}%
    \put(0.37743375,0.04275842){\makebox(0,0)[b]{\smash{10}}}%
    \put(0.43930936,0.05644855){\makebox(0,0)[b]{\smash{11}}}%
    \put(0.50059703,0.07000854){\makebox(0,0)[b]{\smash{12}}}%
    \put(0.56130509,0.08344034){\makebox(0,0)[b]{\smash{13}}}%
    \put(0.62144173,0.0967457){\makebox(0,0)[b]{\smash{14}}}%
    \put(0.68101497,0.10992641){\makebox(0,0)[b]{\smash{15}}}%
    \put(0.78634025,0.26021345){\rotatebox{86.85448}{\makebox(0,0)[lb]{\smash{Performance}}}}%
    \put(0.71946902,0.15602649){\makebox(0,0)[b]{\smash{-0.775}}}%
    \put(0.72168548,0.19432763){\makebox(0,0)[b]{\smash{-0.770}}}%
    \put(0.723934,0.23318283){\makebox(0,0)[b]{\smash{-0.765}}}%
    \put(0.72621526,0.27260419){\makebox(0,0)[b]{\smash{-0.760}}}%
    \put(0.72853001,0.31260419){\makebox(0,0)[b]{\smash{-0.755}}}%
    \put(0.73087899,0.35319566){\makebox(0,0)[b]{\smash{-0.750}}}%
    \put(0.73326298,0.39439177){\makebox(0,0)[b]{\smash{-0.745}}}%
    \put(0.73568273,0.43620619){\makebox(0,0)[b]{\smash{-0.740}}}%
    \put(0.39999999,0.64546625){\makebox(0,0)[b]{\smash{kernel size 3 and kernel size 2}}}%
    \put(0.92565244,0.18704234){\makebox(0,0)[lb]{\smash{-0.775}}}%
    \put(0.92565244,0.24940922){\makebox(0,0)[lb]{\smash{-0.770}}}%
    \put(0.92565244,0.31177615){\makebox(0,0)[lb]{\smash{-0.765}}}%
    \put(0.92565244,0.37414306){\makebox(0,0)[lb]{\smash{-0.760}}}%
    \put(0.92565244,0.43650999){\makebox(0,0)[lb]{\smash{-0.755}}}%
    \put(0.92565244,0.4988769){\makebox(0,0)[lb]{\smash{-0.750}}}%
  \end{picture}%
\endgroup%

%% file: img/conv-kernel_size-2-and-num_units_dense_layer.pdf_tex
%% Creator: Inkscape inkscape 0.48.5, www.inkscape.org
%% Accompanies image file 'conv-kernel_size-2-and-num_units_dense_layer.pdf' (pdf, eps, ps)
\begingroup%
  \makeatletter%
  \providecommand\color[2][]{%
    \errmessage{(Inkscape) Color is used for the text in Inkscape, but the package 'color.sty' is not loaded}%
    \renewcommand\color[2][]{}%
  }%
  \providecommand\transparent[1]{%
    \errmessage{(Inkscape) Transparency is used (non-zero) for the text in Inkscape, but the package 'transparent.sty' is not loaded}%
    \renewcommand\transparent[1]{}%
  }%
  \providecommand\rotatebox[2]{#2}%
  \ifx\svgwidth\undefined%
    \setlength{\unitlength}{1156.31992187bp}%
    \ifx\svgscale\undefined%
      \relax%
    \else%
      \setlength{\unitlength}{\unitlength * \real{\svgscale}}%
    \fi%
  \else%
    \setlength{\unitlength}{\svgwidth}%
  \fi%
  \global\let\svgwidth\undefined%
  \global\let\svgscale\undefined%
  \makeatother%
  \begin{picture}(1,0.69738481)%
    \put(0,0){\includegraphics[width=\unitlength]{conv-kernel_size-2-and-num_units_dense_layer.pdf}}%
    \put(0.11703824,0.15894678){\rotatebox{-58.810798}{\makebox(0,0)[lb]{\smash{kernel size 2}}}}%
    \put(0.22439404,0.06084026){\makebox(0,0)[b]{\smash{9}}}%
    \put(0.20782048,0.08860631){\makebox(0,0)[b]{\smash{10}}}%
    \put(0.19168355,0.11564083){\makebox(0,0)[b]{\smash{11}}}%
    \put(0.17596623,0.14197243){\makebox(0,0)[b]{\smash{12}}}%
    \put(0.16065236,0.16762808){\makebox(0,0)[b]{\smash{13}}}%
    \put(0.1457266,0.19263353){\makebox(0,0)[b]{\smash{14}}}%
    \put(0.13117437,0.21701318){\makebox(0,0)[b]{\smash{15}}}%
    \put(0.43370002,0.04529267){\rotatebox{7.271678}{\makebox(0,0)[lb]{\smash{num\_units\_dense\_layer}}}}%
    \put(0.68179281,0.09960016){\makebox(0,0)[b]{\smash{0}}}%
    \put(0.6344078,0.09351184){\makebox(0,0)[b]{\smash{250}}}%
    \put(0.58673934,0.08738716){\makebox(0,0)[b]{\smash{500}}}%
    \put(0.53878479,0.08122569){\makebox(0,0)[b]{\smash{750}}}%
    \put(0.49054162,0.07502711){\makebox(0,0)[b]{\smash{1000}}}%
    \put(0.44200718,0.06879116){\makebox(0,0)[b]{\smash{1250}}}%
    \put(0.39317885,0.06251741){\makebox(0,0)[b]{\smash{1500}}}%
    \put(0.34405397,0.05620561){\makebox(0,0)[b]{\smash{1750}}}%
    \put(0.29462981,0.04985528){\makebox(0,0)[b]{\smash{2000}}}%
    \put(0.77451291,0.26500011){\rotatebox{87.899616}{\makebox(0,0)[lb]{\smash{Performance}}}}%
    \put(0.70856305,0.11881558){\makebox(0,0)[b]{\smash{-0.82}}}%
    \put(0.71041397,0.16673418){\makebox(0,0)[b]{\smash{-0.80}}}%
    \put(0.71228801,0.21525226){\makebox(0,0)[b]{\smash{-0.78}}}%
    \put(0.7141856,0.26438109){\makebox(0,0)[b]{\smash{-0.76}}}%
    \put(0.7161073,0.31413233){\makebox(0,0)[b]{\smash{-0.74}}}%
    \put(0.71805345,0.36451787){\makebox(0,0)[b]{\smash{-0.72}}}%
    \put(0.72002461,0.41554991){\makebox(0,0)[b]{\smash{-0.70}}}%
    \put(0.72202122,0.46724099){\makebox(0,0)[b]{\smash{-0.68}}}%
    \put(0.40000001,0.6467829){\makebox(0,0)[b]{\smash{kernel size 2 and num\_units\_dense\_layer}}}%
    \put(0.92579214,0.22052944){\makebox(0,0)[lb]{\smash{-0.80}}}%
    \put(0.92579214,0.27567306){\makebox(0,0)[lb]{\smash{-0.78}}}%
    \put(0.92579214,0.33081668){\makebox(0,0)[lb]{\smash{-0.76}}}%
    \put(0.92579214,0.38596028){\makebox(0,0)[lb]{\smash{-0.74}}}%
    \put(0.92579214,0.4411039){\makebox(0,0)[lb]{\smash{-0.72}}}%
    \put(0.92579214,0.49624749){\makebox(0,0)[lb]{\smash{-0.70}}}%
  \end{picture}%
\endgroup%

%% file: img/stride-conv-architecture.pdf_tex
%% Accompanies image file 'stride-conv-architecture.pdf' (pdf, eps, ps)
\begingroup%
  \makeatletter%
  \providecommand\color[2][]{%
    \errmessage{(Inkscape) Color is used for the text in Inkscape, but the package 'color.sty' is not loaded}%
    \renewcommand\color[2][]{}%
  }%
  \providecommand\transparent[1]{%
    \errmessage{(Inkscape) Transparency is used (non-zero) for the text in Inkscape, but the package 'transparent.sty' is not loaded}%
    \renewcommand\transparent[1]{}%
  }%
  \providecommand\rotatebox[2]{#2}%
  \newcommand*\fsize{\dimexpr\f@size pt\relax}%
  \newcommand*\lineheight[1]{\fontsize{\fsize}{#1\fsize}\selectfont}%
  \ifx\svgwidth\undefined%
    \setlength{\unitlength}{269.25bp}%
    \ifx\svgscale\undefined%
      \relax%
    \else%
      \setlength{\unitlength}{\unitlength * \real{\svgscale}}%
    \fi%
  \else%
    \setlength{\unitlength}{\svgwidth}%
  \fi%
  \global\let\svgwidth\undefined%
  \global\let\svgscale\undefined%
  \makeatother%
  \begin{picture}(1,0.90983118)%
    \lineheight{1}%
    \setlength\tabcolsep{0pt}%
    \put(0,0){\includegraphics[width=\unitlength,page=1]{stride-conv-architecture.pdf}}%
    \put(0.81197772,0.28166583){\makebox(0,0)[t]{\lineheight{1.25}\smash{\begin{tabular}[t]{c}$y$\end{tabular}}}}%
    \put(0,0){\includegraphics[width=\unitlength,page=2]{stride-conv-architecture.pdf}}%
    \put(0.81754875,0.55186082){\makebox(0,0)[t]{\lineheight{1.25}\smash{\begin{tabular}[t]{c}Convolution layer\end{tabular}}}}%
    \put(0.81754875,0.51843464){\makebox(0,0)[t]{\lineheight{1.25}\smash{\begin{tabular}[t]{c}(ReLu)\end{tabular}}}}%
    \put(0,0){\includegraphics[width=\unitlength,page=3]{stride-conv-architecture.pdf}}%
    \put(0.81197772,0.74406138){\makebox(0,0)[t]{\lineheight{1.25}\smash{\begin{tabular}[t]{c}Max-pooling layer\end{tabular}}}}%
    \put(0,0){\includegraphics[width=\unitlength,page=4]{stride-conv-architecture.pdf}}%
  \end{picture}%
\endgroup%

%% file: img/perf_constraining_.pdf_tex
%% Accompanies image file 'perf_constraining_.pdf' (pdf, eps, ps)
\begingroup%
  \makeatletter%
  \providecommand\color[2][]{%
    \errmessage{(Inkscape) Color is used for the text in Inkscape, but the package 'color.sty' is not loaded}%
    \renewcommand\color[2][]{}%
  }%
  \providecommand\transparent[1]{%
    \errmessage{(Inkscape) Transparency is used (non-zero) for the text in Inkscape, but the package 'transparent.sty' is not loaded}%
    \renewcommand\transparent[1]{}%
  }%
  \providecommand\rotatebox[2]{#2}%
  \newcommand*\fsize{\dimexpr\f@size pt\relax}%
  \newcommand*\lineheight[1]{\fontsize{\fsize}{#1\fsize}\selectfont}%
  \ifx\svgwidth\undefined%
    \setlength{\unitlength}{432bp}%
    \ifx\svgscale\undefined%
      \relax%
    \else%
      \setlength{\unitlength}{\unitlength * \real{\svgscale}}%
    \fi%
  \else%
    \setlength{\unitlength}{\svgwidth}%
  \fi%
  \global\let\svgwidth\undefined%
  \global\let\svgscale\undefined%
  \makeatother%
  \begin{picture}(1,0.66666667)%
    \lineheight{1}%
    \setlength\tabcolsep{0pt}%
    \put(0,0){\includegraphics[width=\unitlength,page=1]{perf_constraining_.pdf}}%
    \put(0.16022728,0.04954063){\makebox(0,0)[t]{\lineheight{1.25}\smash{\begin{tabular}[t]{c}0.0\end{tabular}}}}%
    \put(0,0){\includegraphics[width=\unitlength,page=2]{perf_constraining_.pdf}}%
    \put(0.31174243,0.04954063){\makebox(0,0)[t]{\lineheight{1.25}\smash{\begin{tabular}[t]{c}0.2\end{tabular}}}}%
    \put(0,0){\includegraphics[width=\unitlength,page=3]{perf_constraining_.pdf}}%
    \put(0.46325758,0.04954063){\makebox(0,0)[t]{\lineheight{1.25}\smash{\begin{tabular}[t]{c}0.4\end{tabular}}}}%
    \put(0,0){\includegraphics[width=\unitlength,page=4]{perf_constraining_.pdf}}%
    \put(0.61477273,0.04954063){\makebox(0,0)[t]{\lineheight{1.25}\smash{\begin{tabular}[t]{c}0.6\end{tabular}}}}%
    \put(0,0){\includegraphics[width=\unitlength,page=5]{perf_constraining_.pdf}}%
    \put(0.76628791,0.04954063){\makebox(0,0)[t]{\lineheight{1.25}\smash{\begin{tabular}[t]{c}0.8\end{tabular}}}}%
    \put(0.48587963,0.01329572){\makebox(0,0)[lt]{\lineheight{1.25}\smash{\begin{tabular}[t]{l}$\tau_{imp}$\end{tabular}}}}%
    \put(0,0){\includegraphics[width=\unitlength,page=6]{perf_constraining_.pdf}}%
    \put(0.1087963,0.1102793){\makebox(0,0)[rt]{\lineheight{1.25}\smash{\begin{tabular}[t]{r}30\end{tabular}}}}%
    \put(0,0){\includegraphics[width=\unitlength,page=7]{perf_constraining_.pdf}}%
    \put(0.1087963,0.18719398){\makebox(0,0)[rt]{\lineheight{1.25}\smash{\begin{tabular}[t]{r}40\end{tabular}}}}%
    \put(0,0){\includegraphics[width=\unitlength,page=8]{perf_constraining_.pdf}}%
    \put(0.1087963,0.26410866){\makebox(0,0)[rt]{\lineheight{1.25}\smash{\begin{tabular}[t]{r}50\end{tabular}}}}%
    \put(0,0){\includegraphics[width=\unitlength,page=9]{perf_constraining_.pdf}}%
    \put(0.1087963,0.3410233){\makebox(0,0)[rt]{\lineheight{1.25}\smash{\begin{tabular}[t]{r}60\end{tabular}}}}%
    \put(0,0){\includegraphics[width=\unitlength,page=10]{perf_constraining_.pdf}}%
    \put(0.1087963,0.41793799){\makebox(0,0)[rt]{\lineheight{1.25}\smash{\begin{tabular}[t]{r}70\end{tabular}}}}%
    \put(0,0){\includegraphics[width=\unitlength,page=11]{perf_constraining_.pdf}}%
    \put(0.1087963,0.49485265){\makebox(0,0)[rt]{\lineheight{1.25}\smash{\begin{tabular}[t]{r}80\end{tabular}}}}%
    \put(0,0){\includegraphics[width=\unitlength,page=12]{perf_constraining_.pdf}}%
    \put(0.1087963,0.57176732){\makebox(0,0)[rt]{\lineheight{1.25}\smash{\begin{tabular}[t]{r}90\end{tabular}}}}%
    \put(0.06526692,0.335){\color[rgb]{0,0,0}\rotatebox{90}{\makebox(0,0)[t]{\lineheight{1.25}\smash{\begin{tabular}[t]{c}f1 score\end{tabular}}}}}%
    \put(0,0){\includegraphics[width=\unitlength,page=13]{perf_constraining_.pdf}}%
    \put(0.91620368,0.09741762){\makebox(0,0)[lt]{\lineheight{1.25}\smash{\begin{tabular}[t]{l}0\end{tabular}}}}%
    \put(0,0){\includegraphics[width=\unitlength,page=14]{perf_constraining_.pdf}}%
    \put(0.91620368,0.18893277){\makebox(0,0)[lt]{\lineheight{1.25}\smash{\begin{tabular}[t]{l}5\end{tabular}}}}%
    \put(0,0){\includegraphics[width=\unitlength,page=15]{perf_constraining_.pdf}}%
    \put(0.91620368,0.28044792){\makebox(0,0)[lt]{\lineheight{1.25}\smash{\begin{tabular}[t]{l}10\end{tabular}}}}%
    \put(0,0){\includegraphics[width=\unitlength,page=16]{perf_constraining_.pdf}}%
    \put(0.91620368,0.37196309){\makebox(0,0)[lt]{\lineheight{1.25}\smash{\begin{tabular}[t]{l}15\end{tabular}}}}%
    \put(0,0){\includegraphics[width=\unitlength,page=17]{perf_constraining_.pdf}}%
    \put(0.91620368,0.46347823){\makebox(0,0)[lt]{\lineheight{1.25}\smash{\begin{tabular}[t]{l}20\end{tabular}}}}%
    \put(0,0){\includegraphics[width=\unitlength,page=18]{perf_constraining_.pdf}}%
    \put(0.91620368,0.55499339){\makebox(0,0)[lt]{\lineheight{1.25}\smash{\begin{tabular}[t]{l}25\end{tabular}}}}%
    \put(0.97274306,0.23777778){\color[rgb]{0,0,0}\rotatebox{90}{\makebox(0,0)[lt]{\lineheight{1.25}\smash{\begin{tabular}[t]{l}\# of data sources\end{tabular}}}}}%
    \put(0,0){\includegraphics[width=\unitlength,page=19]{perf_constraining_.pdf}}%
    \put(0.22505405,0.15102728){\makebox(0,0)[lt]{\lineheight{1.25}\smash{\begin{tabular}[t]{l}Performance\end{tabular}}}}%
    \put(0,0){\includegraphics[width=\unitlength,page=20]{perf_constraining_.pdf}}%
    \put(0.22505405,0.11705013){\makebox(0,0)[lt]{\lineheight{1.25}\smash{\begin{tabular}[t]{l}$|\mathcal{S}_y|$ on avg.\end{tabular}}}}%
  \end{picture}%
\endgroup%

%% file: img/kr-shl-real-scenario.pdf_tex
%% Accompanies image file 'kr-shl-real-scenario.pdf' (pdf, eps, ps)
\begingroup%
  \makeatletter%
  \providecommand\color[2][]{%
    \errmessage{(Inkscape) Color is used for the text in Inkscape, but the package 'color.sty' is not loaded}%
    \renewcommand\color[2][]{}%
  }%
  \providecommand\transparent[1]{%
    \errmessage{(Inkscape) Transparency is used (non-zero) for the text in Inkscape, but the package 'transparent.sty' is not loaded}%
    \renewcommand\transparent[1]{}%
  }%
  \providecommand\rotatebox[2]{#2}%
  \newcommand*\fsize{\dimexpr\f@size pt\relax}%
  \newcommand*\lineheight[1]{\fontsize{\fsize}{#1\fsize}\selectfont}%
  \ifx\svgwidth\undefined%
    \setlength{\unitlength}{1105.71426783bp}%
    \ifx\svgscale\undefined%
      \relax%
    \else%
      \setlength{\unitlength}{\unitlength * \real{\svgscale}}%
    \fi%
  \else%
    \setlength{\unitlength}{\svgwidth}%
  \fi%
  \global\let\svgwidth\undefined%
  \global\let\svgscale\undefined%
  \makeatother%
  \begin{picture}(1,0.77601312)%
    \lineheight{1}%
    \setlength\tabcolsep{0pt}%
    \put(0,0){\includegraphics[width=\unitlength,page=1]{kr-shl-real-scenario.pdf}}%
    \put(0.13772868,0.0149396){\makebox(0,0)[t]{\lineheight{1.25}\smash{\begin{tabular}[t]{c}30000\end{tabular}}}}%
    \put(0,0){\includegraphics[width=\unitlength,page=2]{kr-shl-real-scenario.pdf}}%
    \put(0.18879772,0.0149396){\makebox(0,0)[t]{\lineheight{1.25}\smash{\begin{tabular}[t]{c}32000\end{tabular}}}}%
    \put(0,0){\includegraphics[width=\unitlength,page=3]{kr-shl-real-scenario.pdf}}%
    \put(0.23986674,0.0149396){\makebox(0,0)[t]{\lineheight{1.25}\smash{\begin{tabular}[t]{c}34000\end{tabular}}}}%
    \put(0.20028825,0.00256923){\makebox(0,0)[t]{\lineheight{1.25}\smash{\begin{tabular}[t]{c}Step\end{tabular}}}}%
    \put(0.16381224,0.1288323){\makebox(0,0)[t]{\lineheight{1.25}\smash{\begin{tabular}[t]{c}$\tau_{confidence}$\end{tabular}}}}%
    \put(0.65844278,0.12803064){\makebox(0,0)[t]{\lineheight{1.25}\smash{\begin{tabular}[t]{c}$\tau_{confidence}$\end{tabular}}}}%
    \put(0,0){\includegraphics[width=\unitlength,page=4]{kr-shl-real-scenario.pdf}}%
    \put(0.1283068,0.02154097){\makebox(0,0)[rt]{\lineheight{1.25}\smash{\begin{tabular}[t]{r}0.0\end{tabular}}}}%
    \put(0,0){\includegraphics[width=\unitlength,page=5]{kr-shl-real-scenario.pdf}}%
    \put(0.1283068,0.06087121){\makebox(0,0)[rt]{\lineheight{1.25}\smash{\begin{tabular}[t]{r}0.2\end{tabular}}}}%
    \put(0,0){\includegraphics[width=\unitlength,page=6]{kr-shl-real-scenario.pdf}}%
    \put(0.1283068,0.10020145){\makebox(0,0)[rt]{\lineheight{1.25}\smash{\begin{tabular}[t]{r}0.4\end{tabular}}}}%
    \put(0,0){\includegraphics[width=\unitlength,page=7]{kr-shl-real-scenario.pdf}}%
    \put(0.1283068,0.13953167){\makebox(0,0)[rt]{\lineheight{1.25}\smash{\begin{tabular}[t]{r}0.6\end{tabular}}}}%
    \put(0,0){\includegraphics[width=\unitlength,page=8]{kr-shl-real-scenario.pdf}}%
    \put(0.1283068,0.1788619){\makebox(0,0)[rt]{\lineheight{1.25}\smash{\begin{tabular}[t]{r}0.8\end{tabular}}}}%
    \put(0,0){\includegraphics[width=\unitlength,page=9]{kr-shl-real-scenario.pdf}}%
    \put(0.1283068,0.21819214){\makebox(0,0)[rt]{\lineheight{1.25}\smash{\begin{tabular}[t]{r}1.0\end{tabular}}}}%
    \put(0.1084257,0.12330254){\rotatebox{90}{\makebox(0,0)[t]{\lineheight{1.25}\smash{\begin{tabular}[t]{c}Entropy of predictions\end{tabular}}}}}%
    \put(0,0){\includegraphics[width=\unitlength,page=10]{kr-shl-real-scenario.pdf}}%
    \put(0.47254748,0.37297047){\makebox(0,0)[t]{\lineheight{1.25}\smash{\begin{tabular}[t]{c}...\end{tabular}}}}%
    \put(0,0){\includegraphics[width=\unitlength,page=11]{kr-shl-real-scenario.pdf}}%
    \put(0.57609595,0.35151601){\makebox(0,0)[t]{\lineheight{1.25}\smash{\begin{tabular}[t]{c}Walk\end{tabular}}}}%
    \put(0,0){\includegraphics[width=\unitlength,page=12]{kr-shl-real-scenario.pdf}}%
    \put(0.08636725,0.35151601){\makebox(0,0)[t]{\lineheight{1.25}\smash{\begin{tabular}[t]{c}Still\end{tabular}}}}%
    \put(0,0){\includegraphics[width=\unitlength,page=13]{kr-shl-real-scenario.pdf}}%
    \put(0.06511628,0.72819335){\makebox(0,0)[lt]{\lineheight{1.25}\smash{\begin{tabular}[t]{l}$\mathcal{X} = \{$\end{tabular}}}}%
    \put(0.06511628,0.71869723){\makebox(0,0)[lt]{\lineheight{1.25}\smash{\begin{tabular}[t]{l}   gyr\_\_torso,\end{tabular}}}}%
    \put(0.06511628,0.7092011){\makebox(0,0)[lt]{\lineheight{1.25}\smash{\begin{tabular}[t]{l}   acc\_\_hips,\end{tabular}}}}%
    \put(0.06511628,0.69970498){\makebox(0,0)[lt]{\lineheight{1.25}\smash{\begin{tabular}[t]{l}   mag\_\_hips,\end{tabular}}}}%
    \put(0.06511628,0.69020885){\makebox(0,0)[lt]{\lineheight{1.25}\smash{\begin{tabular}[t]{l}   lacc\_\_hips,\end{tabular}}}}%
    \put(0.06511628,0.68071273){\makebox(0,0)[lt]{\lineheight{1.25}\smash{\begin{tabular}[t]{l}   ori\_\_bag,\end{tabular}}}}%
    \put(0.06511628,0.6712166){\makebox(0,0)[lt]{\lineheight{1.25}\smash{\begin{tabular}[t]{l}   acc\_\_bag,\end{tabular}}}}%
    \put(0.06511628,0.66172048){\makebox(0,0)[lt]{\lineheight{1.25}\smash{\begin{tabular}[t]{l}   gra\_\_hips\end{tabular}}}}%
    \put(0.06511628,0.65222436){\makebox(0,0)[lt]{\lineheight{1.25}\smash{\begin{tabular}[t]{l}$\}$\end{tabular}}}}%
    \put(0,0){\includegraphics[width=\unitlength,page=14]{kr-shl-real-scenario.pdf}}%
    \put(0.47096029,0.35151601){\makebox(0,0)[t]{\lineheight{1.25}\smash{\begin{tabular}[t]{c}...\end{tabular}}}}%
    \put(0,0){\includegraphics[width=\unitlength,page=15]{kr-shl-real-scenario.pdf}}%
    \put(0.97052421,0.35151601){\makebox(0,0)[t]{\lineheight{1.25}\smash{\begin{tabular}[t]{c}...\end{tabular}}}}%
    \put(0,0){\includegraphics[width=\unitlength,page=16]{kr-shl-real-scenario.pdf}}%
    \put(0.23875969,0.63560614){\makebox(0,0)[lt]{\lineheight{1.25}\smash{\begin{tabular}[t]{l}$\mathcal{X} = \{$\end{tabular}}}}%
    \put(0.23875969,0.62611001){\makebox(0,0)[lt]{\lineheight{1.25}\smash{\begin{tabular}[t]{l}   gyr\_\_torso,\end{tabular}}}}%
    \put(0.23875969,0.61661389){\makebox(0,0)[lt]{\lineheight{1.25}\smash{\begin{tabular}[t]{l}   acc\_\_hips,\end{tabular}}}}%
    \put(0.23875969,0.60711777){\makebox(0,0)[lt]{\lineheight{1.25}\smash{\begin{tabular}[t]{l}   gra\_\_bag,\end{tabular}}}}%
    \put(0.23875969,0.59762164){\makebox(0,0)[lt]{\lineheight{1.25}\smash{\begin{tabular}[t]{l}   lacc\_\_hips,\end{tabular}}}}%
    \put(0.23875969,0.58812552){\makebox(0,0)[lt]{\lineheight{1.25}\smash{\begin{tabular}[t]{l}   pre\_\_torso,\end{tabular}}}}%
    \put(0.23875969,0.57862939){\makebox(0,0)[lt]{\lineheight{1.25}\smash{\begin{tabular}[t]{l}   acc\_\_torso,\end{tabular}}}}%
    \put(0.23875969,0.56913327){\makebox(0,0)[lt]{\lineheight{1.25}\smash{\begin{tabular}[t]{l}   acc\_\_bag\end{tabular}}}}%
    \put(0.23875969,0.55963715){\makebox(0,0)[lt]{\lineheight{1.25}\smash{\begin{tabular}[t]{l}$\}$\end{tabular}}}}%
    \put(0,0){\includegraphics[width=\unitlength,page=17]{kr-shl-real-scenario.pdf}}%
    \put(0.41138567,0.54573211){\makebox(0,0)[lt]{\lineheight{1.25}\smash{\begin{tabular}[t]{l}$\mathcal{X} = \{$\end{tabular}}}}%
    \put(0.41138567,0.53623598){\makebox(0,0)[lt]{\lineheight{1.25}\smash{\begin{tabular}[t]{l}   gyr\_\_torso,\end{tabular}}}}%
    \put(0.41138567,0.52673986){\makebox(0,0)[lt]{\lineheight{1.25}\smash{\begin{tabular}[t]{l}   acc\_\_hips,\end{tabular}}}}%
    \put(0.41138567,0.51724373){\makebox(0,0)[lt]{\lineheight{1.25}\smash{\begin{tabular}[t]{l}   ori\_\_hips,\end{tabular}}}}%
    \put(0.41138567,0.50774761){\makebox(0,0)[lt]{\lineheight{1.25}\smash{\begin{tabular}[t]{l}   lacc\_\_hips,\end{tabular}}}}%
    \put(0.41138567,0.49825149){\makebox(0,0)[lt]{\lineheight{1.25}\smash{\begin{tabular}[t]{l}   pre\_\_torso,\end{tabular}}}}%
    \put(0.41138567,0.48875536){\makebox(0,0)[lt]{\lineheight{1.25}\smash{\begin{tabular}[t]{l}   acc\_\_torso,\end{tabular}}}}%
    \put(0.41138567,0.47925924){\makebox(0,0)[lt]{\lineheight{1.25}\smash{\begin{tabular}[t]{l}   ori\_\_bag\end{tabular}}}}%
    \put(0.41138567,0.46976311){\makebox(0,0)[lt]{\lineheight{1.25}\smash{\begin{tabular}[t]{l}$\}$\end{tabular}}}}%
    \put(0,0){\includegraphics[width=\unitlength,page=18]{kr-shl-real-scenario.pdf}}%
    \put(0.73052327,0.62848405){\makebox(0,0)[lt]{\lineheight{1.25}\smash{\begin{tabular}[t]{l}$\mathcal{X} = \{$\end{tabular}}}}%
    \put(0.73052327,0.61898792){\makebox(0,0)[lt]{\lineheight{1.25}\smash{\begin{tabular}[t]{l}   gra\_\_hips,\end{tabular}}}}%
    \put(0.73052327,0.6094918){\makebox(0,0)[lt]{\lineheight{1.25}\smash{\begin{tabular}[t]{l}   acc\_\_hips,\end{tabular}}}}%
    \put(0.73052327,0.59999567){\makebox(0,0)[lt]{\lineheight{1.25}\smash{\begin{tabular}[t]{l}   ori\_\_hips,\end{tabular}}}}%
    \put(0.73052327,0.59049955){\makebox(0,0)[lt]{\lineheight{1.25}\smash{\begin{tabular}[t]{l}   lacc\_\_hips,\end{tabular}}}}%
    \put(0.73052327,0.58100342){\makebox(0,0)[lt]{\lineheight{1.25}\smash{\begin{tabular}[t]{l}   pre\_\_torso,\end{tabular}}}}%
    \put(0.73052327,0.5715073){\makebox(0,0)[lt]{\lineheight{1.25}\smash{\begin{tabular}[t]{l}   acc\_\_hand,\end{tabular}}}}%
    \put(0.73052327,0.56201118){\makebox(0,0)[lt]{\lineheight{1.25}\smash{\begin{tabular}[t]{l}   ori\_\_bag\end{tabular}}}}%
    \put(0.73052327,0.55251505){\makebox(0,0)[lt]{\lineheight{1.25}\smash{\begin{tabular}[t]{l}$\}$\end{tabular}}}}%
    \put(0,0){\includegraphics[width=\unitlength,page=19]{kr-shl-real-scenario.pdf}}%
    \put(0.64809705,0.01425158){\makebox(0,0)[t]{\lineheight{1.25}\smash{\begin{tabular}[t]{c}88000\end{tabular}}}}%
    \put(0,0){\includegraphics[width=\unitlength,page=20]{kr-shl-real-scenario.pdf}}%
    \put(0.69916607,0.01425158){\makebox(0,0)[t]{\lineheight{1.25}\smash{\begin{tabular}[t]{c}90000\end{tabular}}}}%
    \put(0,0){\includegraphics[width=\unitlength,page=21]{kr-shl-real-scenario.pdf}}%
    \put(0.75023511,0.01425158){\makebox(0,0)[t]{\lineheight{1.25}\smash{\begin{tabular}[t]{c}92000\end{tabular}}}}%
    \put(0.69753185,0.00188121){\makebox(0,0)[t]{\lineheight{1.25}\smash{\begin{tabular}[t]{c}Step\end{tabular}}}}%
    \put(0,0){\includegraphics[width=\unitlength,page=22]{kr-shl-real-scenario.pdf}}%
    \put(0.62555042,0.02085295){\makebox(0,0)[rt]{\lineheight{1.25}\smash{\begin{tabular}[t]{r}0.0\end{tabular}}}}%
    \put(0,0){\includegraphics[width=\unitlength,page=23]{kr-shl-real-scenario.pdf}}%
    \put(0.62555042,0.06018319){\makebox(0,0)[rt]{\lineheight{1.25}\smash{\begin{tabular}[t]{r}0.2\end{tabular}}}}%
    \put(0,0){\includegraphics[width=\unitlength,page=24]{kr-shl-real-scenario.pdf}}%
    \put(0.62555042,0.09951343){\makebox(0,0)[rt]{\lineheight{1.25}\smash{\begin{tabular}[t]{r}0.4\end{tabular}}}}%
    \put(0,0){\includegraphics[width=\unitlength,page=25]{kr-shl-real-scenario.pdf}}%
    \put(0.62555042,0.13884365){\makebox(0,0)[rt]{\lineheight{1.25}\smash{\begin{tabular}[t]{r}0.6\end{tabular}}}}%
    \put(0,0){\includegraphics[width=\unitlength,page=26]{kr-shl-real-scenario.pdf}}%
    \put(0.62555042,0.17817388){\makebox(0,0)[rt]{\lineheight{1.25}\smash{\begin{tabular}[t]{r}0.8\end{tabular}}}}%
    \put(0,0){\includegraphics[width=\unitlength,page=27]{kr-shl-real-scenario.pdf}}%
    \put(0.62555042,0.21750412){\makebox(0,0)[rt]{\lineheight{1.25}\smash{\begin{tabular}[t]{r}1.0\end{tabular}}}}%
    \put(0.60566932,0.12261453){\rotatebox{90}{\makebox(0,0)[t]{\lineheight{1.25}\smash{\begin{tabular}[t]{c}Entropy of predictions\end{tabular}}}}}%
    \put(0,0){\includegraphics[width=\unitlength,page=28]{kr-shl-real-scenario.pdf}}%
  \end{picture}%
\endgroup%